\documentclass[journal]{IEEEtran}

\usepackage[utf8]{inputenc} %
\usepackage[T1]{fontenc}    %
\usepackage{hyperref}       %
\usepackage{url}            %
\usepackage{booktabs}       %
\usepackage{amsfonts}       %
\usepackage{nicefrac}       %
\usepackage{microtype}      %

\usepackage{graphicx}
\usepackage{comment}
\usepackage{amsmath,amssymb} %
\usepackage{color}

\usepackage{epsfig}
\usepackage{algorithm}
\usepackage{algorithmic}
\usepackage{subfigure}
\usepackage{epstopdf}
\usepackage{booktabs}
\usepackage{array}
\usepackage{multirow}
\usepackage{hhline}
\usepackage{makecell}
\usepackage{framed}
\usepackage{enumitem}

\newcommand{\revise}[1]{{\color{black}#1}}

\makeatletter
\newcommand{\printfnsymbol}[1]{%
  \textsuperscript{\@fnsymbol{#1}}%
}
\makeatother

\begin{document}

\title{\revise{Single image reflection removal via learning with multi-image constraints}}

\author{Yingda~Yin\printfnsymbol{1}, Qingnan~Fan\printfnsymbol{1}, Dongdong~Chen, Yujie~Wang, Angelica~I.~Aviles-Rivero, Ruoteng~Li, Carola-Bibiane~Schönlieb, Baoquan Chen.
\thanks{\printfnsymbol{1}: Equal contribution.}
\thanks{Yingda Yin and Baoquan Chen are with the Center on Frontiers of Computing Studies, Peking University, Beijing 100871, China. E-mail: yingda.yin@gmail.com, baoquan.chen@gmail.com.}
\thanks{Qingnan Fan is with the Visual Computing Center of Tencent AI Lab, Tencent, Shenzhen 518000, China. Email: fqnchina@gmail.com.}
\thanks{Yujie Wang is with the Computer Science and Technology Department, Shandong University, Qingdao, Shandong 266237, China. E-mail: yujiew.cn@gmail.com.}
\thanks{Dongdong Chen is with Microsoft Cloud AI, Redmond, Washington 98052, USA. E-mail: cddlyf@gmail.com.}
\thanks{Angelica I. Aviles-Rivero and Carola-Bibiane~Schönlieb are with the DAMTP and DPMMS, University of Cambridge. Email: ai323@cam.ac.uk, cbs31@cam.ac.uk.}
\thanks{Ruoteng Li is with the Electrical and Computer Engineering Department, National University of Singapore. E-mail: liruoteng@u.nus.edu.}}

\markboth{}{}

\graphicspath{{images/toy_example/}{images/blurry_real/}{images/features/}{images/sir2/}{images/ablation/}{images/sir2_crop/}{images/blur_iccv/}{images/real_train/}{images/i11i22i12i21/}{images_jpg/toy_example/}{images_jpg/blurry_real/}{images_jpg/features/}{images_jpg/sir2/}{images_jpg/ablation/}{images_jpg/sir2_crop/}{images_jpg/blur_iccv/}{images_jpg/real_train/}{images_jpg/i11i22i12i21/}{images_jpg/refl_layer}}

\maketitle

\begin{abstract}
	Reflections are very common phenomena in our daily photography, which distract people's attention from the scene behind the glass. The problem of removing reflection artifacts is important but challenging due to its ill-posed nature. The traditional approaches solve an optimization problem over the constraints induced from multiple images, at the expense of large computation costs. Recent learning-based approaches have demonstrated a significant improvement in both performance and running time for single image reflection removal, but are limited as they require a large number of synthetic reflection/clean image pairs for direct supervision to approximate the ground truth, at the risk of overfitting in the synthetic image domain and degrading in the real image domain.	In this paper, we propose a novel learning-based solution that combines the advantages of the aforementioned approaches and overcomes their drawbacks. Our algorithm works by learning a deep neural network to optimize the target with joint constraints enhanced among multiple input images during the training phase, but is able to eliminate reflections only from a single input for evaluation. Our algorithm runs in real-time and achieves state-of-the-art reflection removal performance on real images. We further propose a strong network backbone that disentangles the background and reflection information into separate latent codes, which are embedded into a shared one-branch deep neural network for both background and reflection predictions. The proposed backbone experimentally performs better than the other common network implementations, and provides insightful knowledge to understand the reflection removal task.
\end{abstract}


\IEEEpeerreviewmaketitle

\section{Introduction}

Reflection is commonly observed when taking photos through a piece of glass due to the interference between reflected light and the light coming from the background scene. These reflection artifacts significantly degrade the visibility of the target scene and distract users from focusing on it. Therefore, removing reflections is a problem of great interest for many computer vision and graphics applications.

To deal with this problem, the traditional optimization-based approaches either leverage the relation from multiple input images \cite{xue2015computational,agrawal2005removing}, or rely on strong priors applied on the background or reflection layer from a single input \cite{li2014single,yang2019fast}. Due to the use of hand-crafted features, these methods tend to fail in some challenging cases where prior assumptions are violated, and they usually suffer from large computation costs. Inspired by the tremendous success of deep learning in many image processing tasks, a variety of recent works \cite{fan2017generic,zhang2018single,wan2018crrn,yang2018seeing,wei2019single,wen2019single,chang2019single} have focused on learning-based solutions for reflection removal by making use of the synthetic reflections as training signals. These techniques have been proven to boost the overall performance significantly and achieve state-of-the-art results. Despite their success, by directly supervising the synthetic labels from a single input, the network may learn limited transferable knowledge and generalizes weakly in the real test data as observed in many previous works \cite{chen2019learning,sun2019not,li2019single,tremblay2018training}.

\begin{figure*}[t!]
	\centering
	\includegraphics[width=1\linewidth]{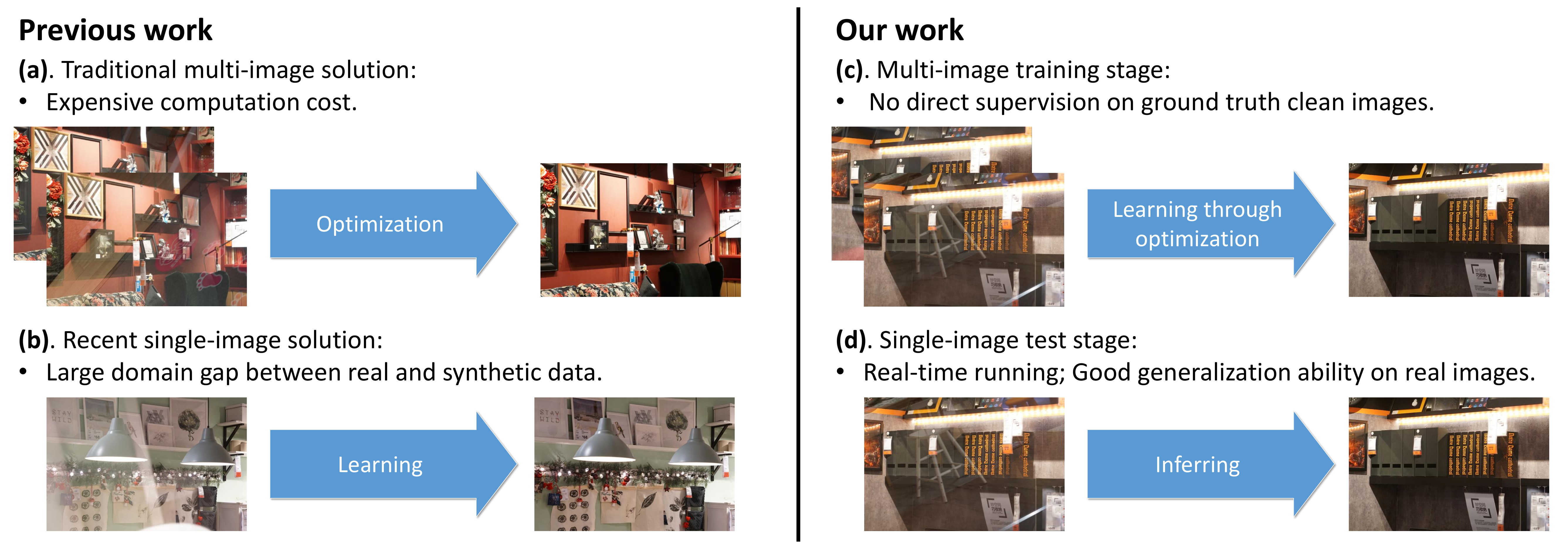}
	\vspace{-3mm}
	\caption{Advantages of our algorithms. \textbf{(a)} The traditional approaches reason the background more theoretically by optimizing an objective function constrained by either the 
	\revise{multi-image constraints} %
	or hand-crafted prior from single-image input, but suffers from expensive computation cost. \textbf{(b)} The recent learning methods work by approximating the reflection data distribution from a single image with direct ground truth supervision, but enlarge the domain gap between synthetic training and real testing images. Our algorithm overcomes their difficulties, and combines their advantages. \textbf{(c)} In the training stage, it learns a deep network to reason the background through the optimization from multiple images without direct ground truth supervision, and \textbf{(d)} is able to infer the background only from a single image in real-time during evaluation. Benefited from the optimization process, it learns better generalization ability on real images.}
	\label{figure:motivation}
\end{figure*}

In this paper, we address the reflection removal problem from a single input image, and propose a novel algorithm to overcome the aforementioned difficulties in both learning and non-learning algorithms. To be specific, (1). In order to avoid the expensive computation cost in non-learning algorithms, we employ a deep neural network to achieve the reflection removal process; (2). In order to minimize the domain gap between the synthetic training and real testing data in learning algorithms, the deep neural network is not directly supervised with ground truth clean image, instead it is indirectly supervised with an optimization-based objective function without requirements of ground truth clean images; (3). In order to relax the hand-crafted gradient priors enforced on a single image in the objective function of non-learning methods, we incorporate the \revise{multi-image constraints}
during training, but are able to reason about the clean background from a single input image during evaluation. Therefore, our algorithm is not only computationally efficient due to the deep learning implementation, but also achieves good generalization ability and state-of-the-art performance on real images thanks to the multi-image optimization-based loss function. Our framework follows a learning through optimization strategy to achieve single image reflection removal, and it builds on an implicit connection between multi-image and single-image methods, illustrated in Figure \ref{figure:motivation}.

For the purpose of implementing the multi-image solution during the training phase, our deep neural network needs to predict both background and reflection layers from an input image. To this aim, we propose a novel and strong network backbone for reflection removal. It couples the background and reflection predictions with the same output channels in a single-branch fully convolution neural network, instead of leveraging a common two-branch implementation. To enable layer separation, two latent codes are independently learned to represent the background and reflection layer. In this manner, the layer representation is embedded into a much smaller weight space constrained by the latent code size compared to the two-branch implementation. Experimentally, we observe better generalization ability of our proposed network backbone benefited from the smaller exploration space. We further demonstrate superior quantitative and qualitative performance over the state-of-the-art reflection removal approaches on various real reflection cases.

All in all, our contributions in this paper can be summarized as follows:

\begin{itemize}
	\item We propose a novel learning through optimization algorithm to address the single image reflection removal problem. It	combines the advantages of both learning and non-learning methods, and does not require the ground truth supervision signal. Our algorithm benefits from multiple inputs during training, but only needs a single input for evaluation.
	\item We develop a strong network backbone for reflection removal. It embeds the background and reflection layer representation into small latent codes, which constrain the solution space for layer separation and enables better generalization ability on the real image domain.
	\item Our proposed algorithm outperforms state-of-the-art supervised approaches on various kinds of challenging real-world reflection images.
\end{itemize}
\section{Related Work}

Reflection removal has been widely explored in the community. The body of literature can be roughly classified into two categories: the non-learning approaches and learning-based approaches.

\textbf{Non-learning approaches.} Since reflection removal is a highly underdetermined problem, approaches leverage different types of priors to deal with this problem. A set of approaches are oriented to use multiple input images \cite{farid1999separating,szeliski2000layer,sarel2004separating,gai2012blind,li2013exploiting,sinha2012image,guo2014robust,xue2015computational,yang2016robust,nandoriya2017video,yang2019fast}. These techniques require that the scene being captured from different viewpoints. They exploit the motion cue, which assumes the background and foreground motion is usually different due to visual parallax \cite{guo2014robust,gai2012blind,szeliski2000layer,xue2015computational}. Some other approaches take a sequence of images using special conditions, such as shooting with polarizer \cite{kong2014physically,schechner2000polarization,sarel2004separating}, flash/non-flash image pair \cite{agrawal2005removing}, focus/defocus image pair \cite{schechner2000separation}, \emph{etc}. Due to the additional information obtained from multiple images, this problem becomes less ill-posed. However, the special data requirement limits these methods from more practical application scenarios.

Another set of techniques rely on the use of single input image \cite{levin2003learning,levin2004separating,wan2018region,arvanitopoulos2017single,yang2019fast}. These approaches leverage the gradient sparsity prior for layer decomposition \cite{springer2017reflection,levin2007user}. As the reflection and background layers are usually at different depths, the reflection plane is very likely to remain outside the depth of field (DOF) and hence becomes blurry. The approaches of \cite{li2014single,wan2016depth} incorporate priors explicitly into the objective function to be optimized. A more realistic physics model takes the thickness of glass into consideration \cite{shih2015reflection}.

\begin{figure*}[t!]
	\centering
	\includegraphics[width=0.94\linewidth]{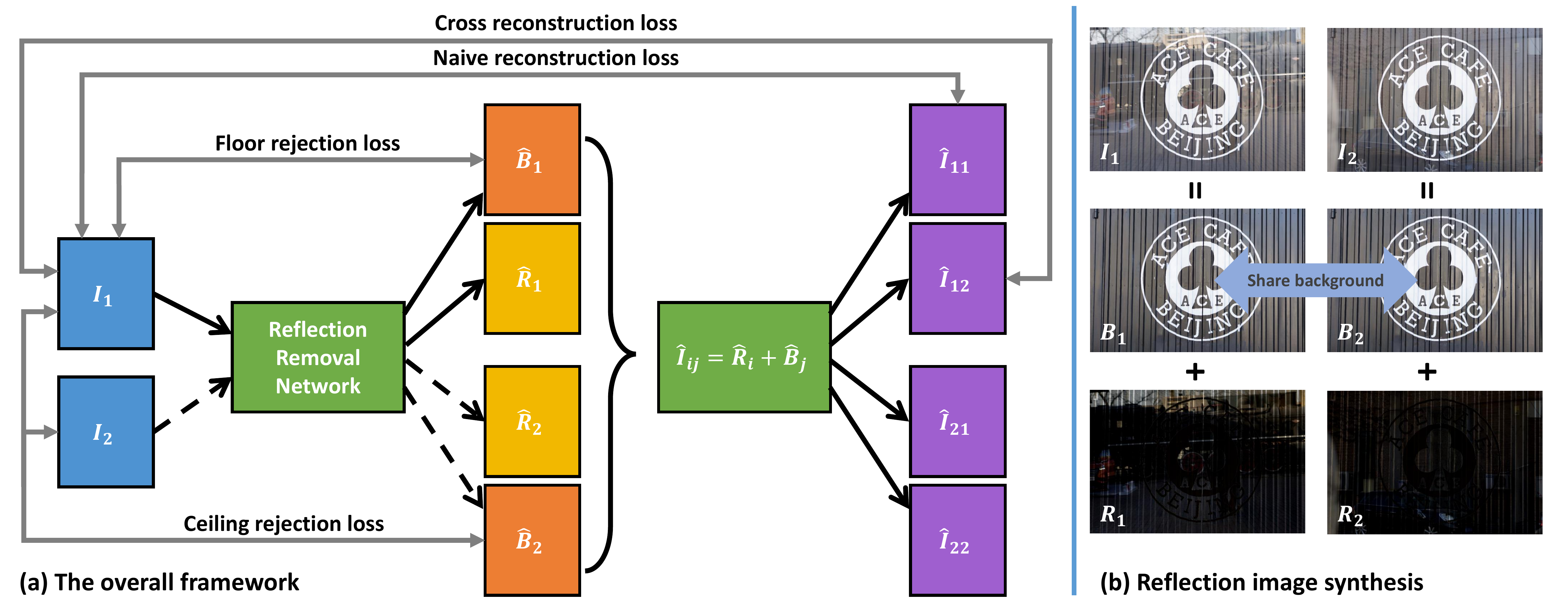}
	\vspace{-1mm}
	\caption{Our overall framework for reflection removal. \textbf{(a)} Illustration of the proposed algorithm. The deep network is fed with a pair of reflection images and decomposes them into corresponding background and reflection layers, which are further composed into new reflection images. We append a joint objective of four loss functions on all the predictions for optimization in absence of ground truth. Since the input images are forwarded into the deep network individually, during evaluation, our framework is able to take a single input to generate the clean image. Note the framework is symmetric and for clarity, we omit similar lines. \textbf{(b)} The corresponding reflection image synthesis pipeline. The two inputs share the same background layer.}
	\label{figure:framework}
\end{figure*}

\textbf{Learning approaches.} \cite{fan2017generic} proposed the first deep learning approach for reflection removal. It relied on the common assumption of blurry reflection to develop a novel reflection image synthesis method. \cite{zhang2018single,wen2019single} followed their data generation approach, and incorporates the adversarial loss to mitigate the image degradation issue. \cite{wan2018crrn, wan2019corrn} acquired a reflection image dataset (RID) which demonstrates mostly sharp and weak reflections, and proposed to unify gradient inference and image reconstruction concurrently into the network design. \cite{wei2019single,ma2019learning} collected the misaligned or unpaired reflection data to enhance the performance on real reflections. 
\revise{\cite{li2020single} proposes a cascaded network that iteratively refines the estimates of transmission and reflection layers. \cite{punnappurath2019reflection} focuses on reflection removal using a dual-pixel sensor by the ``defocus-disparity'' cues.}
Instead of using a single training image, recently \cite{wieschollek2018separating} proposed to learn from multiple polarized images, along with a novel synthetic data generation approach, which accurately simulates reflections in polarized images. \cite{liu2020learning} incorporates a dense motion estimation module and online optimization to remove obstructions from multiple frames.
\revise{Other than removing reflection artifacts, \cite{wan2020reflection} aims at recovering reflection scenes from the mixture image, especially useful for surveillance or criminal investigations.}

Unlike the previous non-learning or learning approaches, our algorithm takes advantage of methods of both types, that is: we leverage multiple input images in the training stage to ease the ill-posedness through optimization, while learning a general parametric solution in the absence of ground truth as direct supervision signal, which requires less hand-crafted features and only a single input during evaluation. It runs in real-time and achieves superior performance on real images.

\section{Approach} \label{sec:objective}

The overall framework is shown in Figure \ref{figure:framework}. 
\revise{Denote reflection image $\mathbf{I}$ as the input image with reflection artifacts, background layer $\mathbf{B}$ and reflection layer $\mathbf{R}$ as the desired contents of the background scene and the undesired reflection artifacts blended in the input images, respectively. }
In the training phase, our algorithm takes two reflection images as input ($I_1, I_2$), which are forwarded into the same reflection removal network independently to predict their corresponding background and reflection layers ($\{\hat{B}_1,\hat{R}_1\}, \{\hat{B}_2,\hat{R}_2\}$). The reflection layers ($\hat{R}_{1}, \hat{R}_{2}$) are composed with every other background layer ($\hat{B}_{1}, \hat{B}_{2}$) to generate a set of reflection images denoted as  ($\hat{I}_{11},\hat{I}_{12},\hat{I}_{21},\hat{I}_{22}$). In the evaluation phase, our algorithm takes only a single reflection image as input, and estimates its background and reflection layer.
Note our algorithm can take unlimited input images for training, here we only demonstrate the two-image case for sake of simplicity.

\subsection{Reflection Image Synthesis} \label{sec:image_synthesis}

Our algorithm is inspired by the traditional multi-image optimization-based algorithms, which take advantage of the motion difference between background and reflection layer from multiple images to design the objective function. However, due to the difficulty of collecting a sufficient amount of reflection image pairs for training deep networks, we choose to synthesize the training data. In order to differentiate the background and reflection layer of each synthetic image pair, we freeze the background layer, and only change the reflection layer. Note this synthetic setting doesn't conform to the realistic scenario since it's difficult to maintain a steady background due to the common camera shake. However, the multi-image prior is utilized in an optimization manner to encourage the neural network to remove reflections from a single input image, hence the network is less sensitive to the synthesis process of reflection image pairs.

Mathematically, the reflection image synthesis process is formulated as follows. Given a background ($B$) and a reflection ($R$) layer, the corresponding reflection image ($I$) is a linear combination as,
\begin{equation} \label{eq:base}
I_i = B_i + R_i
\end{equation}
where $i$ indicates the image index. Therefore to prepare for multi-image reflection removal, we build an inner connection between the different input reflection images, which reads:
\begin{equation}
B_1 = B_2
\end{equation}
It means that all the input reflection images share the same background, and differentiate only in the reflection layer.

\setlength{\tabcolsep}{0.5pt}
\renewcommand{\arraystretch}{1}
\begin{figure*}[t]
	\begin{center}
		
		\begin{tabular}{cccc cccc}
	
			\includegraphics[width=2.2cm]{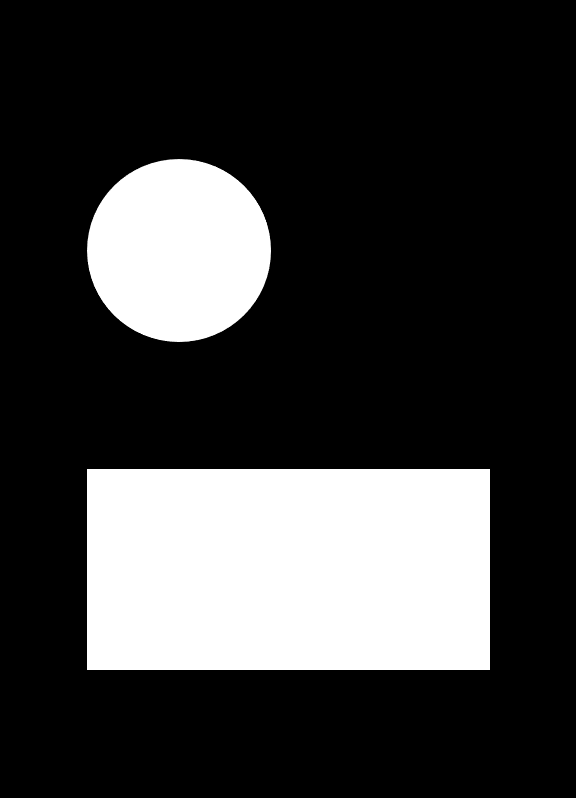}
			&\includegraphics[width=2.2cm]{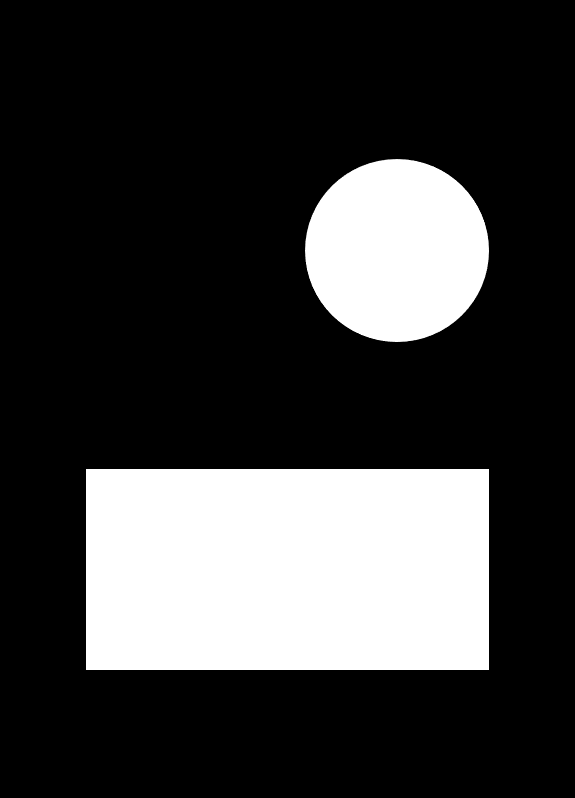}
			&\includegraphics[width=2.2cm]{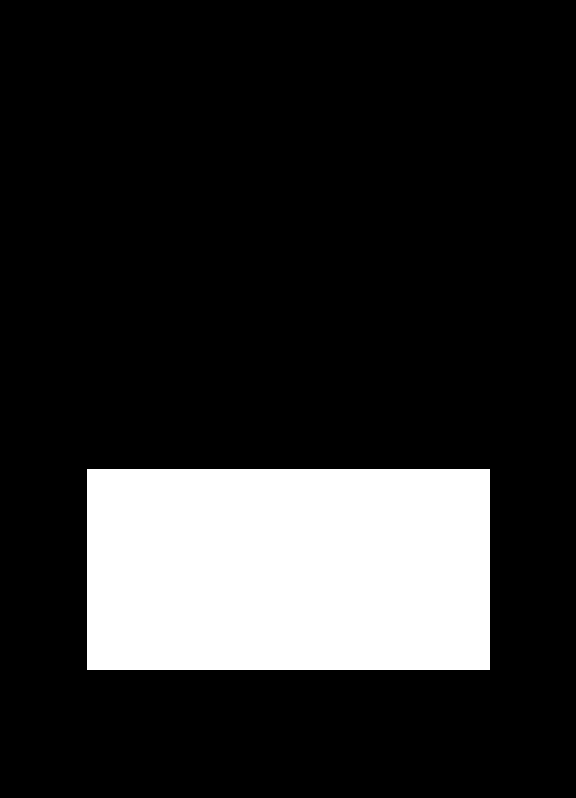}
			&\includegraphics[width=2.2cm]{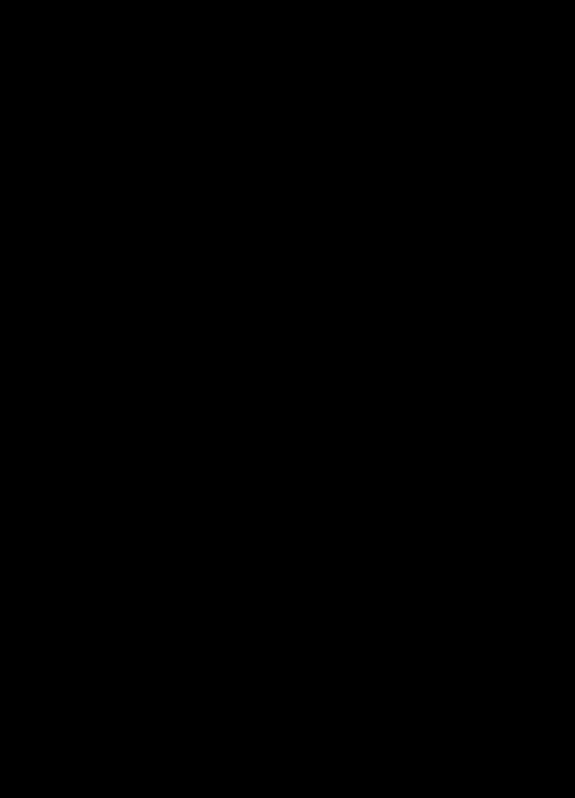}
			&\includegraphics[width=2.2cm]{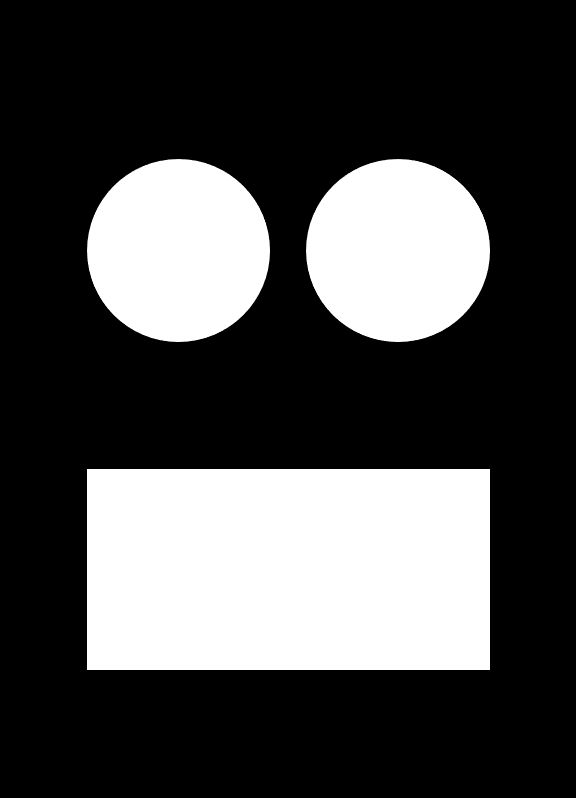}
			&\includegraphics[width=2.2cm]{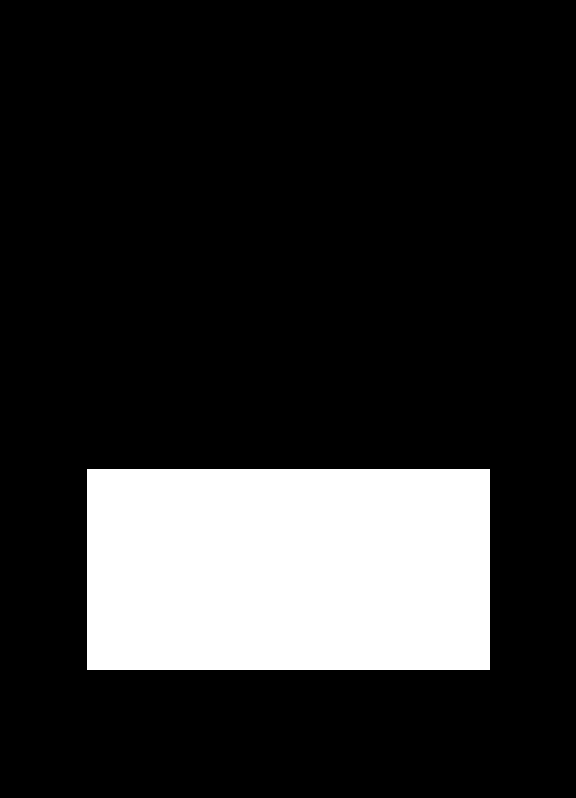}
			&\includegraphics[width=2.2cm]{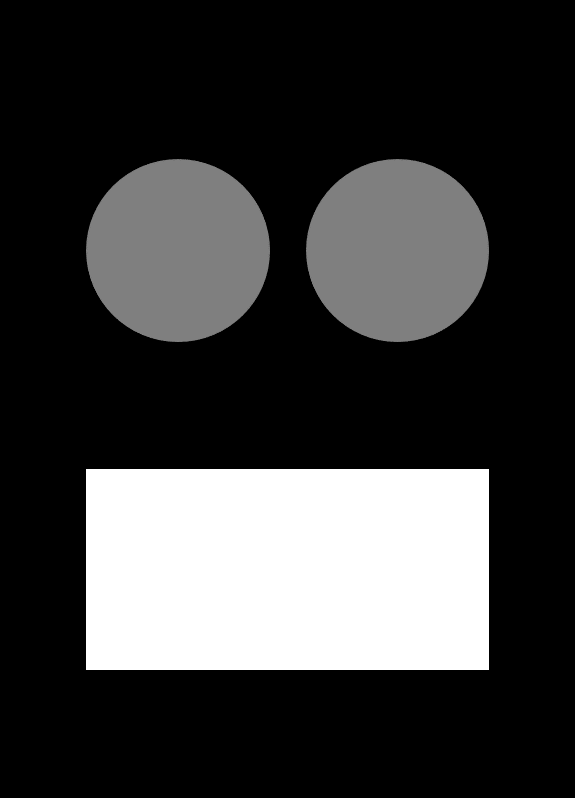}
			&\includegraphics[width=2.2cm]{loss12_v2.png}									
			\\
			
			\small{$I_1$} & \small{$I_2$} & \small{${B}_1 \& {B}_2$} & \footnotesize{$\hat{B}_1 \& \hat{B}_2(\mathcal{L}_{r.})$} & \scriptsize{$\hat{B}_1 \& \hat{B}_2(\mathcal{L}_{r.}$+$\mathcal{L}_{f.})$} & \scriptsize{$\hat{B}_1 \& \hat{B}_2(\mathcal{L}_{r.}$+$\mathcal{L}_{f.})$} & \scriptsize{$\hat{B}_1 \& \hat{B}_2(\mathcal{L}_{r.}$+$\mathcal{L}_{f.})$} & \scriptsize{$\hat{B}_1 \& \hat{B}_2$(all losses)}
			\\
		\end{tabular}
	
		\begin{tabular}{cccc cccc}
			\vspace{-0.6mm}			
			\includegraphics[width=2.20cm]{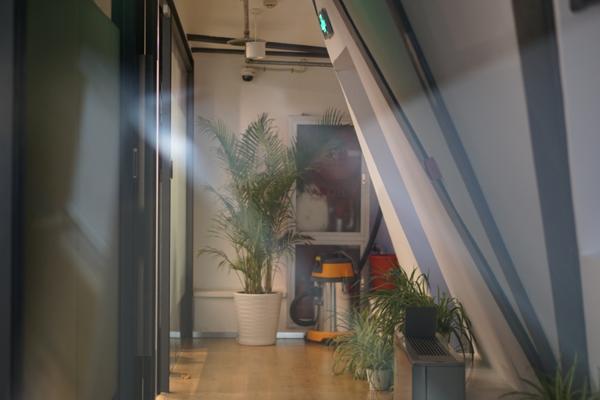}
			&\includegraphics[width=2.20cm]{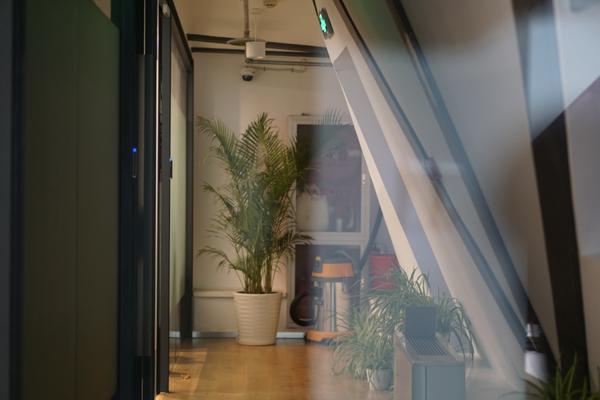}
			&\includegraphics[width=2.20cm]{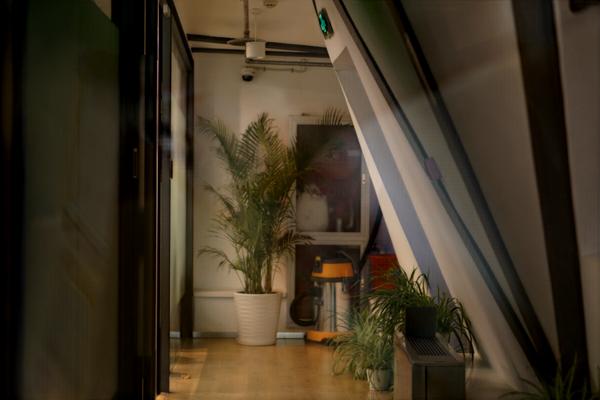}			
			&\includegraphics[width=2.20cm]{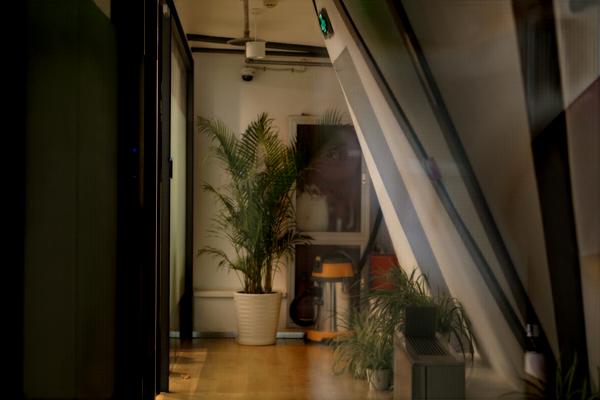}
			&\includegraphics[width=2.20cm]{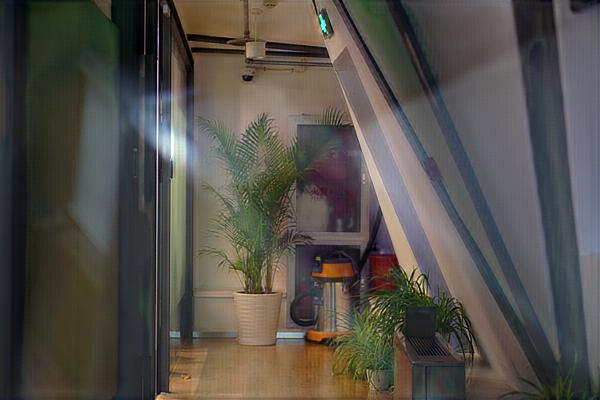}
			&\includegraphics[width=2.20cm]{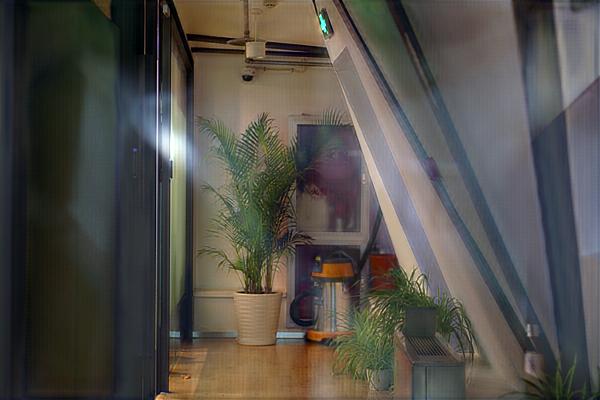}
			&\includegraphics[width=2.20cm]{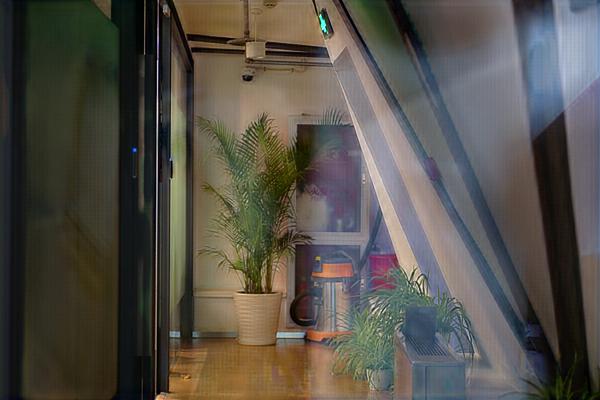}
			&\includegraphics[width=2.20cm]{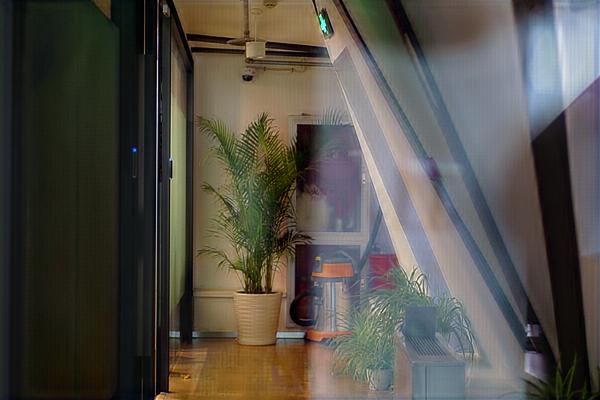}									
			\\
			\vspace{-0.6mm}			
			\includegraphics[width=2.20cm]{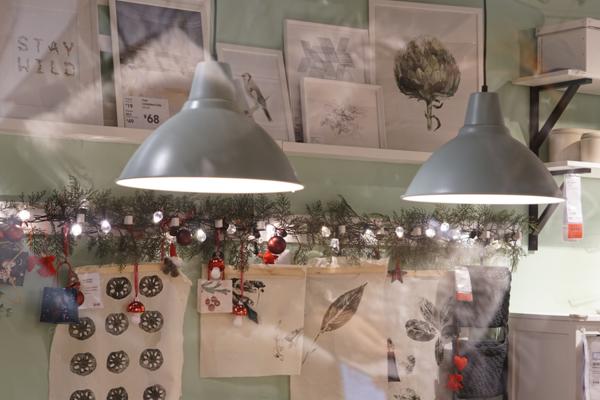}
			&\includegraphics[width=2.20cm]{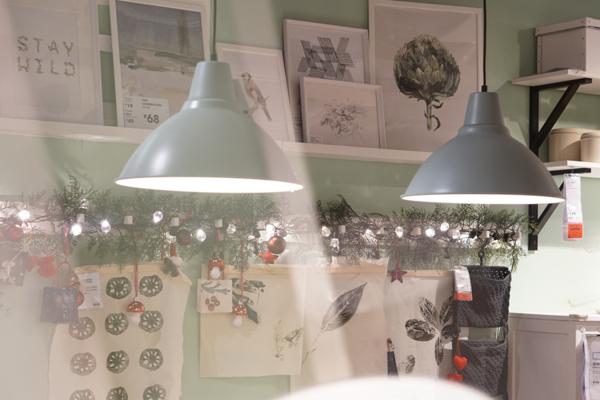}
			&\includegraphics[width=2.20cm]{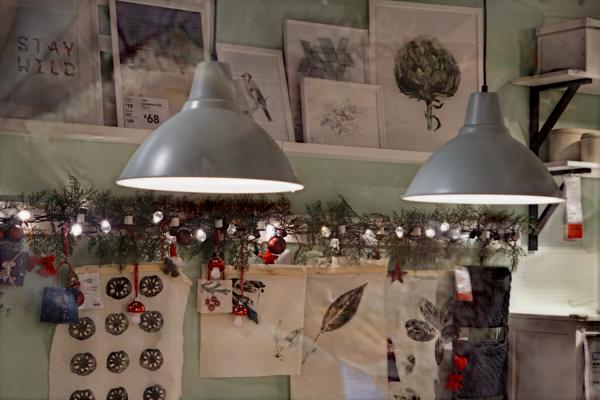}			
			&\includegraphics[width=2.20cm]{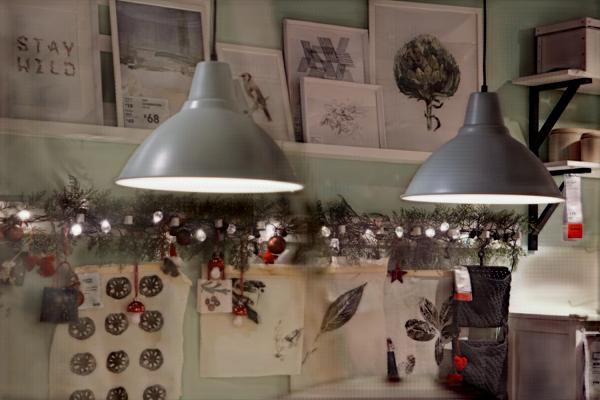}
			&\includegraphics[width=2.20cm]{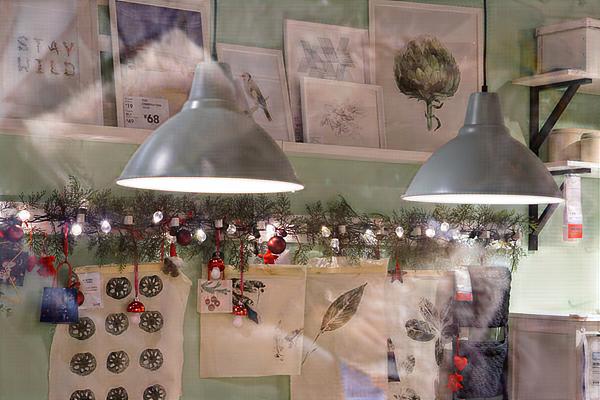}
			&\includegraphics[width=2.20cm]{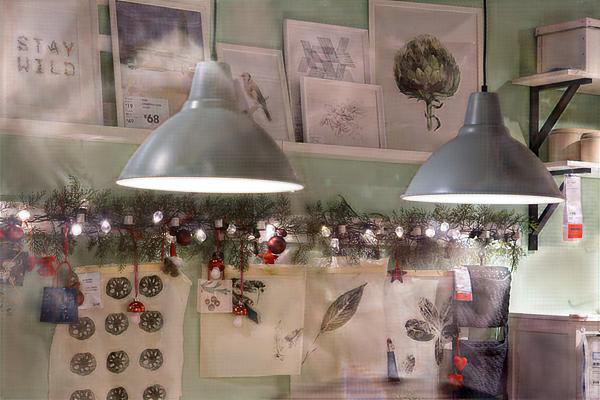}
			&\includegraphics[width=2.20cm]{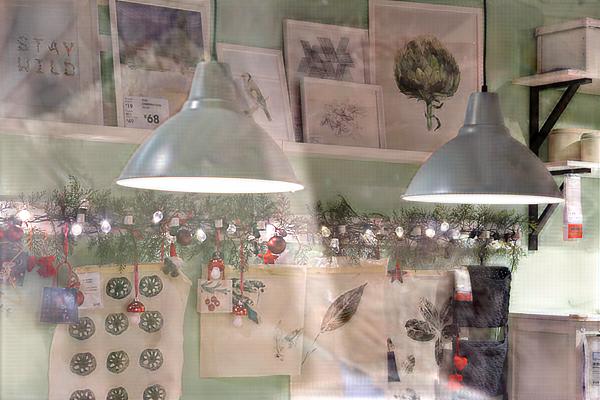}
			&\includegraphics[width=2.20cm]{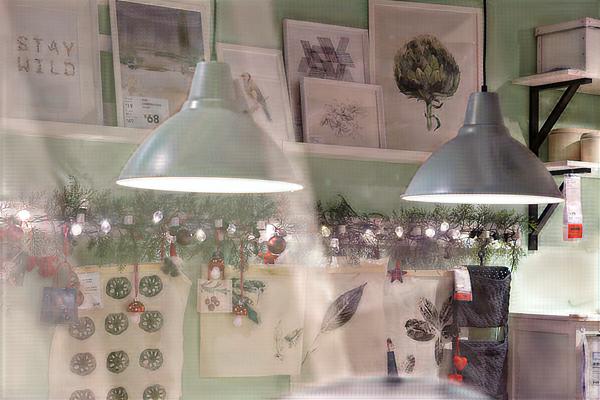}									
			\\
			\vspace{-0.6mm}			
			\small{$I_1$} & \small{$I_2$} &\small{$\hat{B}_{1}$} &\small{$\hat{B}_{2}$} &\small{$\hat{I}_{11}$} &\small{$\hat{I}_{12}$} &\small{$\hat{I}_{21}$} &\small{$\hat{I}_{22}$} 
			\\
		\end{tabular}

	\end{center}
	\vspace{-2mm}
	\caption{Analysis of the proposed objective function applied on a toy example (top) and optimized on two real reflection image pairs (bottom two). $\mathcal{L}_{r.}$ and $\mathcal{L}_{f.}$ are short for $\mathcal{L}_{recons}$ and $\mathcal{L}_{floor}$ separately. The toy example demonstrates the theoretical analysis of how each loss function affects the final background predictions, while the real reflection image pairs show the results optimized via a simple gradient descent approach over the overall object function.}
	\label{figure:illustration}
\end{figure*}

\subsection{Objective Function}

Taking advantage of the constraints applied on the input reflection image pair, we design an objective function composed of four loss functions, where the background and reflection layer 
\revise{($\hat{B}_1$,$\hat{B}_2$,$\hat{R}_1$,$\hat{R}_2$)}
are the optimization targets. This objective function is implemented as a differentiable layer in the deep learning framework to supervise the neural network. Therefore, the common gradient descent optimizer is used to optimize the objective function, and the gradient is backpropagated from the background and reflection predictions to the network. We demonstrate the four loss functions as below.

\subsubsection{Naive Reconstruction Loss}
Given the decomposed background and reflection layer, the most straightforward supervision can be applied by minimizing the per-pixel difference between the recomposed reflection image and the original input samples as,
\begin{equation}\label{equation:loss_naive}
\mathcal{L}_{naive} = ||\hat{I}_{11}-I_1||_2^2 + ||\hat{I}_{22}-I_2||_2^2
\end{equation}
where $\hat{I}_{ij}$ is synthesized via $\hat{I}_{ij} = \hat{R}_i + \hat{B}_j$. Despite this objective already reduces the solution space greatly, it is still unclear which one of the two outputs should be background ($\hat{B}_j$), or likewise reflection ($\hat{R}_j$).

\subsubsection{Cross Reconstruction Loss}
Following the definition of our reflection synthesis pipeline, given the predicted $\hat{B}_2$ and $\hat{R}_1$, we should ideally be able to reconstruct the reflection input $I_1$ due to shared background among multiple inputs. Therefore to enhance such a constraint, we further reconstruct the cross compositions of background and reflection layers as
\begin{equation}\label{equation:loss_cross}
\mathcal{L}_{cross} = ||\hat{I}_{12}-I_1||_2^2 + ||\hat{I}_{21}-I_2||_2^2
\end{equation}

Then the network is forced to disentangle the background from reflection through the reconstruction process by specifying each output component. The naive reconstruction loss and cross reconstruction loss can be combined as
\begin{equation}\label{equation:loss_reconstruct}
\mathcal{L}_{recons} = \mathcal{L}_{naive} + \mathcal{L}_{cross}
\end{equation}

We demonstrate how the loss functions benefit the reflection removal for a toy example in Figure \ref{figure:illustration}. The two input images ($I_1, I_2$) contain a rectangle that belongs to the background, and a circle that belongs to the reflection. By simply optimizing the naive and cross reconstruction losses, a naive yet optimal solution of zero energy can be easily explored by pushing all the information from the input images into the reflection layer, while having the network output a black background layer of all zero values, denoted as $\hat{B}_1 \& \hat{B}_2(\mathcal{L}_{r.})$ in Figure \ref{figure:illustration}. Therefore, in order to enforce the correct layer separation, we propose the following two loss functions as further constraints on the background and reflection predictions.

\begin{figure*}[t]
	\centering
	\includegraphics[width=0.94\linewidth]{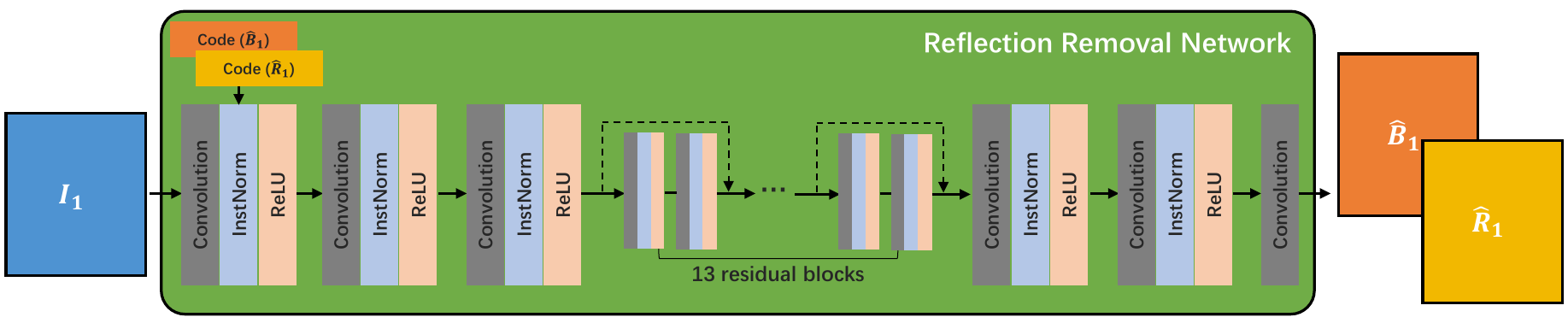}
	\vspace{-1mm}
	\caption{Our detailed reflection removal network backbone. This is a one-branch fully convolution neural network. To enable both background and reflection prediction within such a network, we introduce the latent code defined by the learnable parameters in the first instance normalization layer to encode the output image information. To be specific, two latent codes that represent background and reflection are learned separately, and fed to the network independently to generate the corresponding output.}
	\label{figure:network}
\end{figure*}

\subsubsection{Floor Rejection Loss}

To eliminate the background degeneration artifacts as shown in $\hat{B}_1 \& \hat{B}_2(\mathcal{L}_{r.})$ in Figure \ref{figure:illustration}, we further encourage the output background layer to approach the input reflection image by appending a floor rejection loss. Mathematically, we define: 
\begin{equation}\label{equation:loss_floor}
\mathcal{L}_{floor} = ||\hat{B}_1-I_1||_1 + ||\hat{B}_2-I_2||_1
\end{equation}

By applying the aforementioned three loss functions, we demonstrate three possible output backgrounds, shown in the toy example in Figure \ref{figure:illustration}, denoted as $\hat{B}_1 \& \hat{B}_2(\mathcal{L}_{r.}$+$\mathcal{L}_{f.})$. Numerically, these alternative solutions are all able to achieve zero energy for the reconstruction losses and the same energy for the floor rejection loss, while visually we observe the white rectangle background manage to appear at the risk of the emergence of reflections in the background layer.

\subsubsection{Ceiling Rejection Loss}
The combination of the above loss functions yields a better disentanglement, which is however still insufficient and ambiguous to generate a purely clean background as shown in the toy example. As the two input reflection images are highly correlated, we take better advantage of their relationship by appending a spatially-variant ceiling rejection loss which prevents the pixel-wise background intensity from exceeding each counterpart of the two input images, reads as,
\begin{equation}\label{equation:loss_ceiling_pixel}
f(\hat{B}_i, I_j, m) =
\begin{cases}
||\hat{B}_{i,m} - I_{j,m}||_1 & \hat{B}_{i,m} > I_{j,m},\\
0 &\mbox{otherwise};
\end{cases}
\end{equation}
\begin{equation}\label{equation:loss_ceiling}
\begin{aligned}
\mathcal{L}_{ceiling} = \sum_{m}(f(\hat{B}_1, I_1, m)&+f(\hat{B}_1, I_2, m)+ \\[-8pt]
f(\hat{B}_2, I_1, m)&+f(\hat{B}_2, I_2, m))
\end{aligned}
\end{equation}
where $m$ indicates each image pixel. The ceiling rejection loss is a summation of $f$ among the whole image region. It sets an upper bound for each background prediction, and punishes any background pixel whose intensity value exceeds any input. As shown in Figure \ref{figure:illustration}, via applying the four loss functions, the ceiling rejection loss helps alleviate the ambiguity of multiple background solutions, and leaves only the true answer, denoted as $\hat{B}_1 \& \hat{B}_2$(all losses).

The overall objective function is a weighted combination of the above four loss terms, defined as
\begin{equation}\label{equation:loss_all}
\mathcal{L}_{all} = \lambda_1 \mathcal{L}_{recons} + \lambda_2 \mathcal{L}_{floor} + \mathcal{L}_{ceiling}
\end{equation}
where the specific value for $\lambda_1$ and $\lambda_2$ are addressed in the experiment section. Note the network training procedure is essentially a continuous coordination and competition process of all the loss terms. The reconstruction loss sets the lower bound of the background, and the floor and ceiling rejection losses set the upper bound of the background.

\subsection{Optimization} \label{sec:optimization}

We apply the straightforward gradient descent approach for optimization of the objective function in order to justify the effectiveness of the multi-image reflection removal solution. We adopt the stochastic optimizer Adam \cite{kingma2014adam} which is also very popular as a deep learning training strategy. The background and reflection layer for two input images are initialized as random noise sampled from the unit Gaussian distribution, and the initial learning rate for Adam is set as 0.1.

To better demonstrate the robustness of the objective function, instead of optimizing over the synthetic data, we capture two real reflection image pairs that follow the principle in the reflection image synthesis pipeline in Section \ref{sec:image_synthesis}. We hold the camera steady by attaching it on a tripod to maintain the background unchanged, and only vary the reflections using a floating piece of glass for different input reflection images.

We show the captured reflection images and the corresponding optimized results in the bottom two rows in Figure \ref{figure:illustration}. We observe very clean background results ($\hat{B}_{1},\hat{B}_{2}$) by leveraging our proposed objective functions. Furthermore, the recomposed input images ($\hat{I}_{11},\hat{I}_{12},\hat{I}_{21},\hat{I}_{22}$) are also visually close to the captured reflection images. The results verify the effectiveness and robustness of the objective function.

\setlength{\tabcolsep}{1pt}
\renewcommand{\arraystretch}{1}
\begin{figure*}[t]
	\begin{center}
		
		\begin{tabular}{cccc cccc}
			
			\vspace{-0.6mm}
			\includegraphics[width=2.18cm]{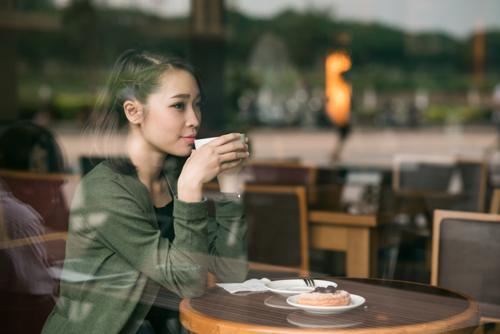}
			&\includegraphics[width=2.18cm]{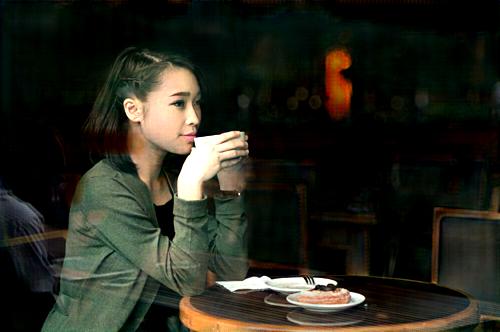}
			&\includegraphics[width=2.18cm]{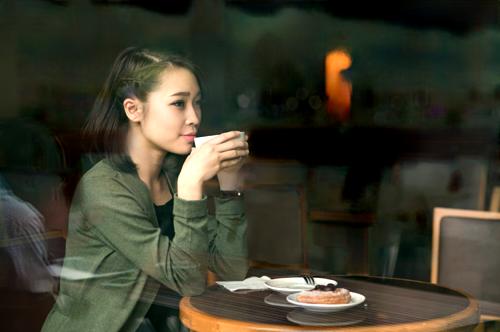}
			&\includegraphics[width=2.18cm]{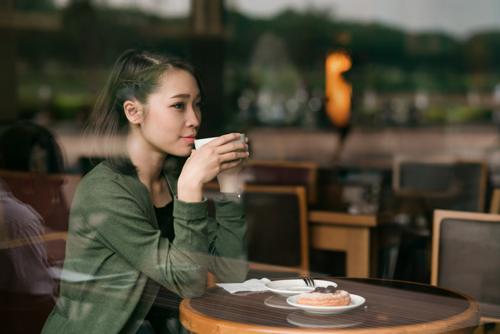}
			&\includegraphics[width=2.18cm]{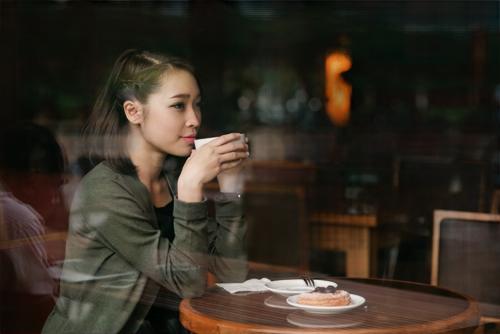}
			&\includegraphics[width=2.18cm]{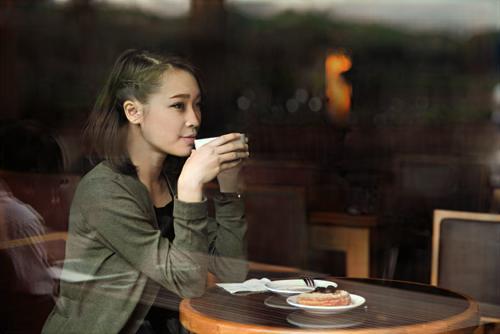}
			&\includegraphics[width=2.18cm]{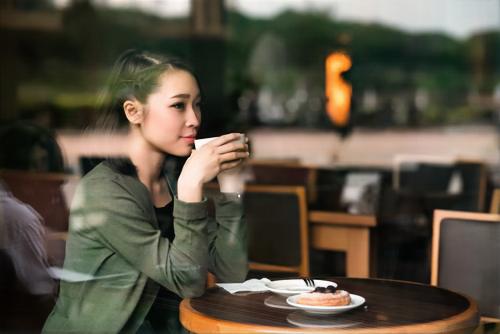}
			&\includegraphics[width=2.18cm]{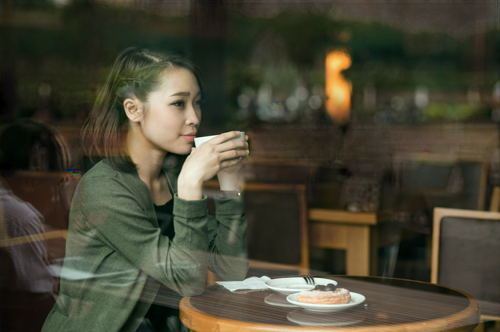}
			\\
			&\includegraphics[width=2.18cm]{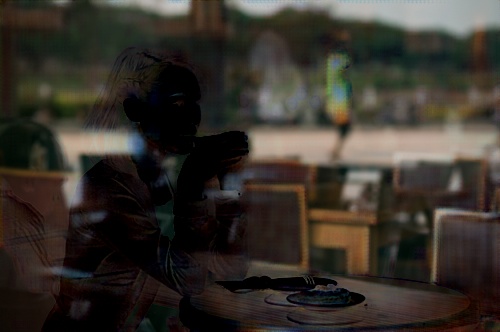}
			&\includegraphics[width=2.18cm]{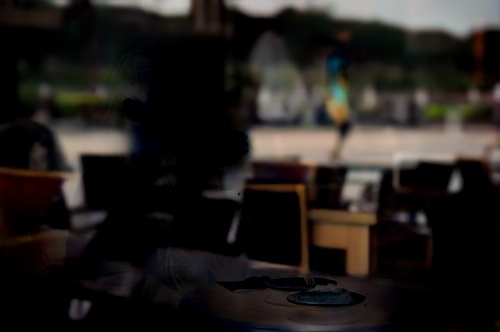}
			&\includegraphics[width=2.18cm]{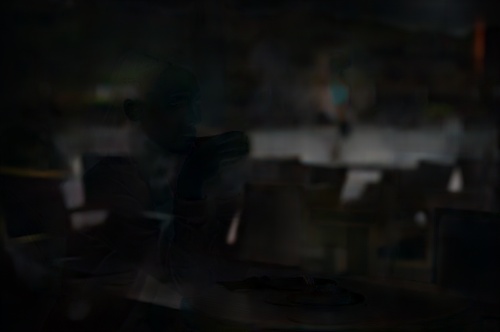}
			&\includegraphics[width=2.18cm]{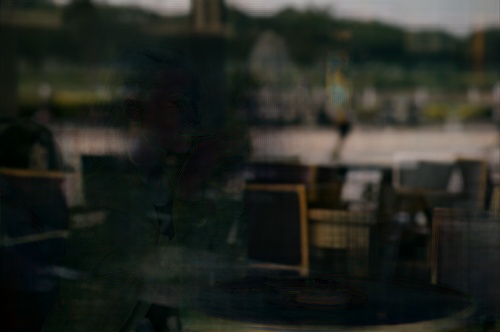}
			&\includegraphics[width=2.18cm]{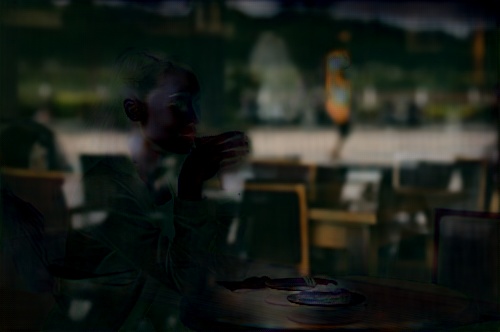}
			&\includegraphics[width=2.18cm]{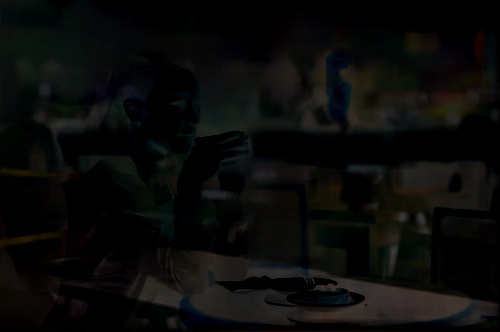}
			&\includegraphics[width=2.18cm]{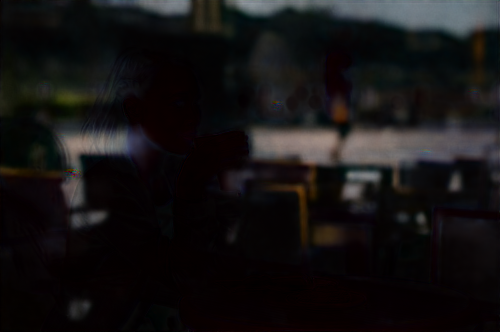}
            \\
            
			\vspace{-0.6mm}
			\includegraphics[width=2.18cm]{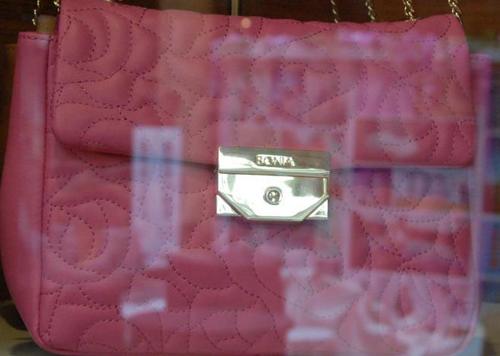}
			&\includegraphics[width=2.18cm]{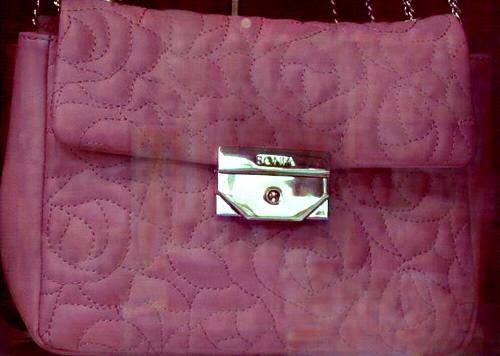}
			&\includegraphics[width=2.18cm]{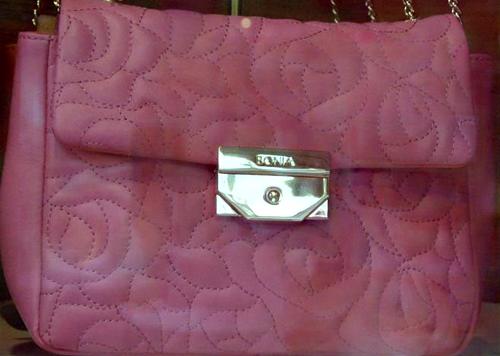}
			&\includegraphics[width=2.18cm]{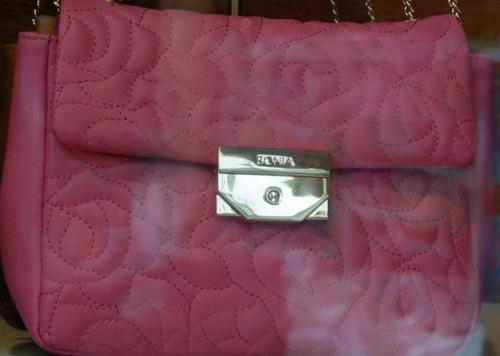}
			&\includegraphics[width=2.18cm]{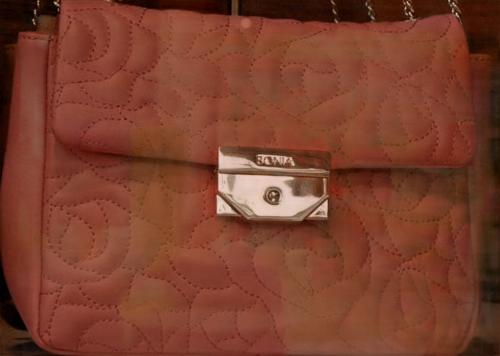}
			&\includegraphics[width=2.18cm]{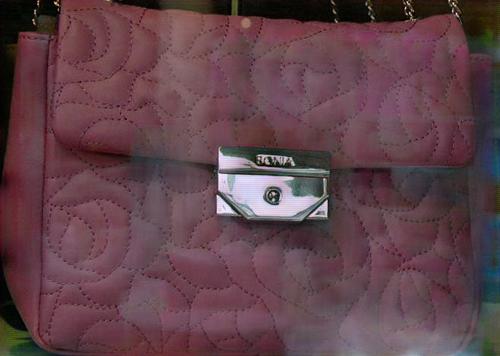}
			&\includegraphics[width=2.18cm]{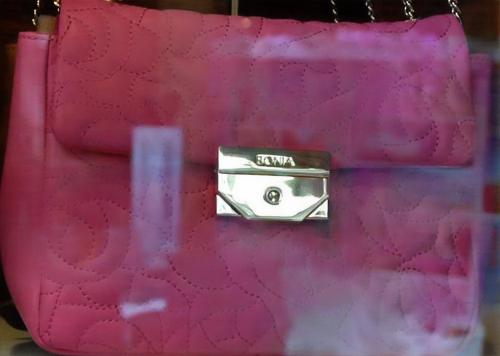}
			&\includegraphics[width=2.18cm]{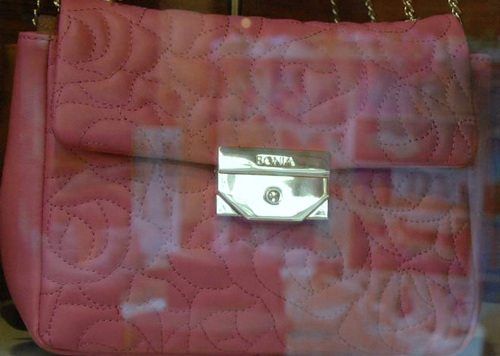}
			\\
			&\includegraphics[width=2.18cm]{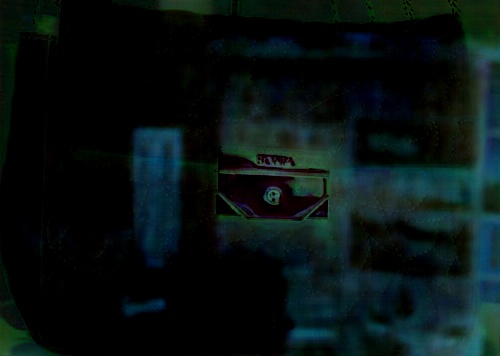}
			&\includegraphics[width=2.18cm]{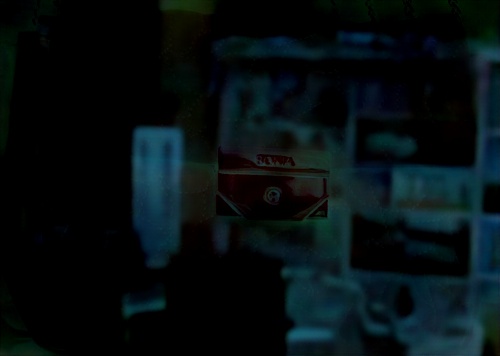}
			&\includegraphics[width=2.18cm]{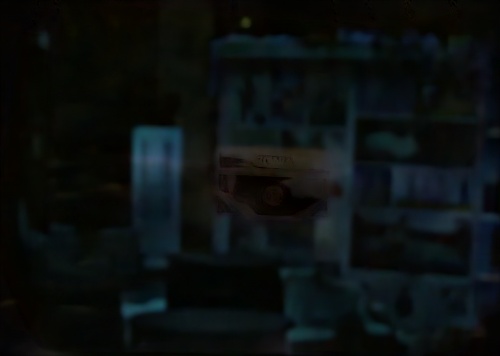}
			&\includegraphics[width=2.18cm]{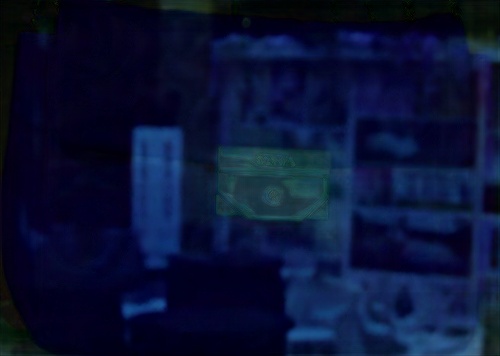}
			&\includegraphics[width=2.18cm]{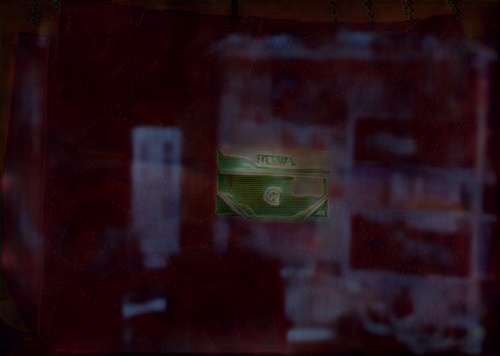}
			&\includegraphics[width=2.18cm]{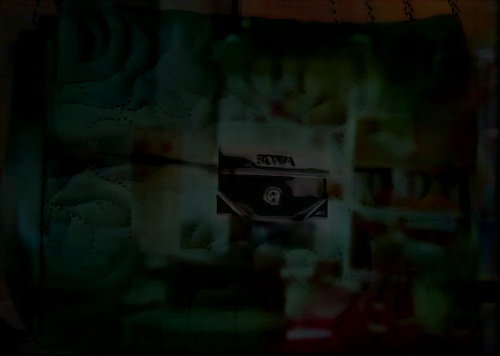}
			&\includegraphics[width=2.18cm]{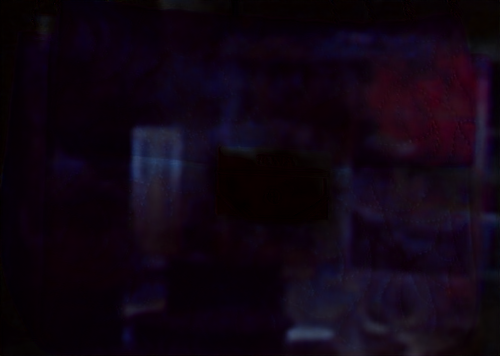}	
			\\

			\vspace{-0.6mm}
			\includegraphics[width=2.18cm]{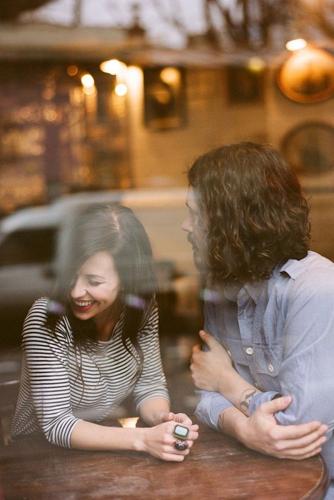}
			&\includegraphics[width=2.18cm]{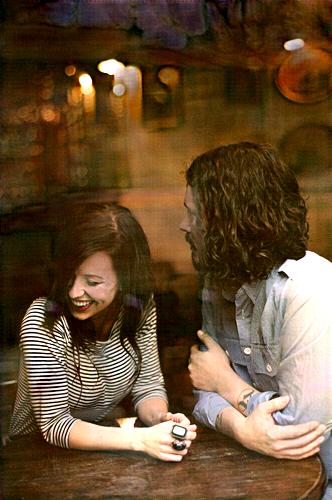}
			&\includegraphics[width=2.18cm]{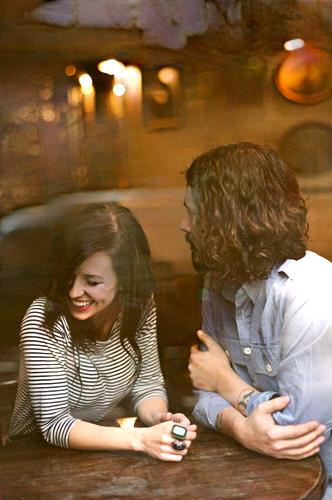}
			&\includegraphics[width=2.18cm]{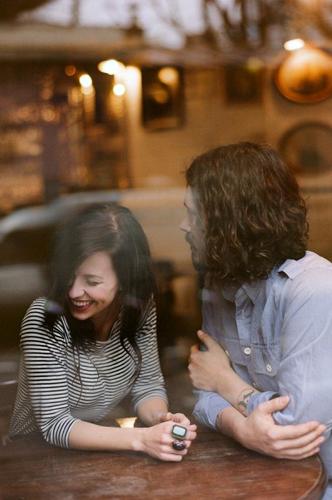}
			&\includegraphics[width=2.18cm]{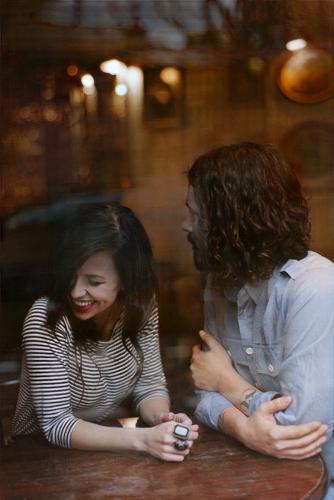}
			&\includegraphics[width=2.18cm]{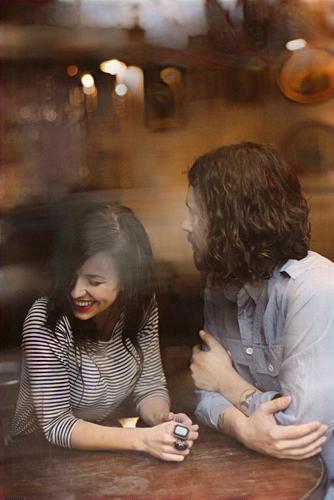}
			&\includegraphics[width=2.18cm]{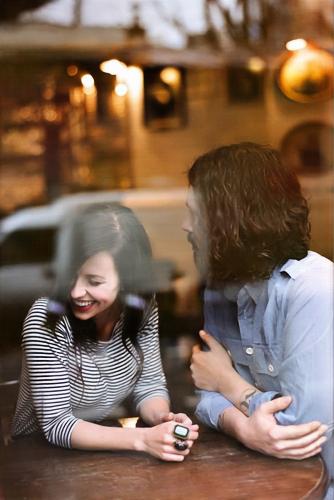}
			&\includegraphics[width=2.18cm]{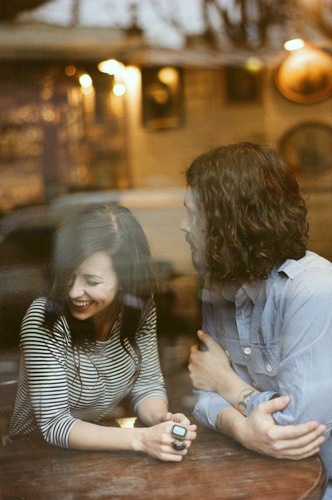}
			\\
			
			&\includegraphics[width=2.18cm]{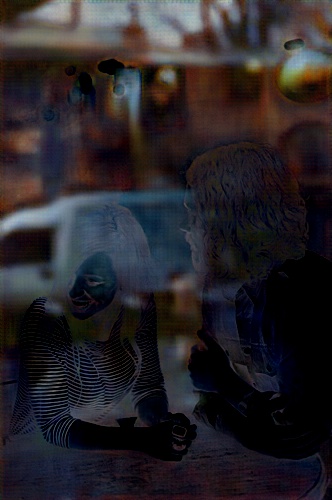}
			&\includegraphics[width=2.18cm]{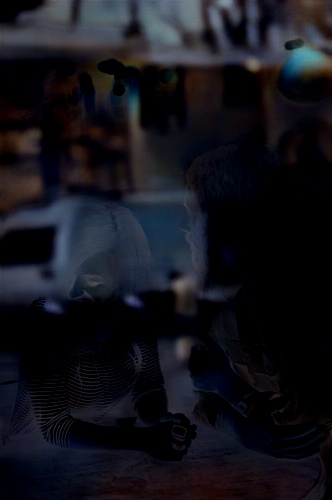}
			&\includegraphics[width=2.18cm]{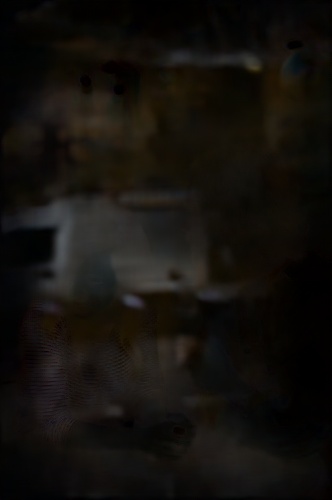}
			&\includegraphics[width=2.18cm]{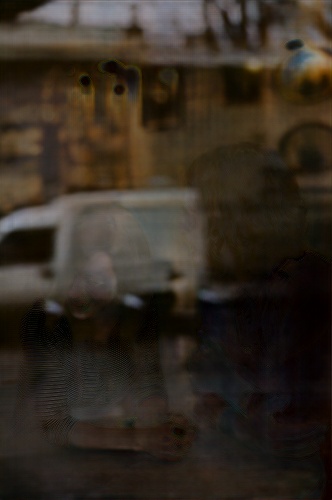}
			&\includegraphics[width=2.18cm]{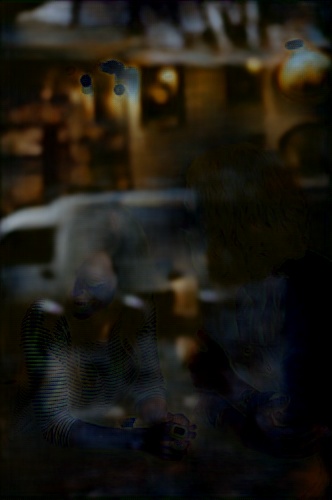}
			&\includegraphics[width=2.18cm]{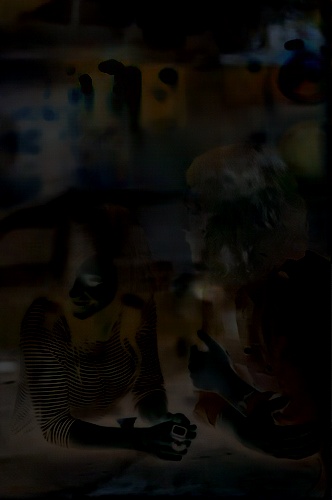}
			&\includegraphics[width=2.18cm]{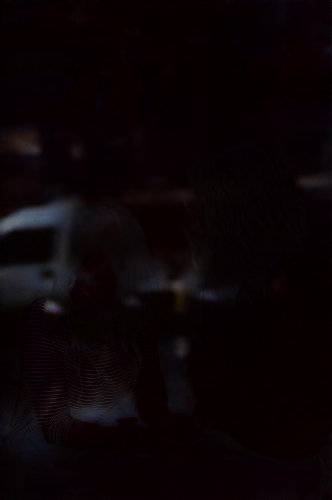}
			
			\\

			\small{Input} & \small{Ours} & \small{Ours*} & \small{\cite{wei2019single}} & \small{\cite{zhang2018single}} & \small{\cite{fan2017generic}} & \small{\cite{yang2018seeing}}  & \small{\cite{li2020single}}
			\\
		\end{tabular}
		
	\end{center}
	\vspace{-3mm}
	\caption{Visual comparison \revise{of the predicted background (top) and reflection (bottom) layers} on the strong real-world reflection images from \cite{fan2017generic}. These images are featured by subtraction and clipping image model. Ours* refers to the version with ground truth supervision.}
	\label{figure:fanICCV}
\end{figure*}

\subsection{Learning Through Optimization}

The aforementioned objective function and optimization approach is designed for multi-image reflection removal, but suffers from (1) expensive computation time due to the iterative optimization process, (2) various input images with the same background layer. Our goal in this paper is to relax these constraints to achieve single image reflection removal in real-time.

Towards this goal, we propose to leverage a deep neural network to generate the background and reflection layer from a single input image. Therefore, in the training stage where multiple images are taken as input, the deep network processes them sequentially to generate all the outputs. Then the proposed objective function is optimized by the stochastic gradient descent approach as shown in Section \ref{sec:optimization}, and the gradients are propagated back from all the outputs to the same deep neural network to update its parameters. In this manner, we are able to achieve single image reflection removal for evaluation while taking advantage of the multi-image prior during the training phase. Since the network is powered with GPU implementation, it is very time-efficient for running.

\noindent\textbf{Reflection removal network backbone.} To achieve the goal of having both background and reflection predictions, one network of multiple output branches or even two independent neural networks are usually required. In this paper, we develop a novel approach to couple the background and reflection within a single network. Following the basic network structure in \cite{fan2017generic}, our neural network is a fully convolution neural network (FCN), which contains 32 convolution layers, where the middle 26 ones are organized into 13 residual blocks to accelerate convergence. All the convolution layers use 3$\times$3 kernels with 64 output channels, and are followed by instance normalization \cite{ulyanov2016instance} and ReLU layer except for the last one. The input and output of the network are both a three-channel color image.

However, training such a one-branch fully convolution neural network is challenging, as it does not support two concurrent outputs for a given single input image. To transform this network more suitably, we introduce the latent code that explicitly represents the background or reflection information. This latent code is implemented as the learnable parameters (scale and shift) in the first instance normalization layer, which is decoupled from the network as an independent parameterized vector. During the training process, two latent codes that represent the background and reflection are independently learned. Given a single input image, the output of the reflection removal network is determined by switching the two controllable latent codes. As the instance normalization layer takes 64 feature maps as input, the latent code is only defined in a tiny vector of size 128. Benefited from such a design, the specific background or reflection information are all encoded into these 128 learnable parameters, while all the other network weights are shared for them. Experimentally, we observe the best performance by selecting the first instance normalization layer, since in this manner, the most subsequent layers can leverage the latent code to better differentiate the background and reflection.

Experimentally, we observe better generalization ability on real images of our proposed reflection removal network backbone compared to the implementation of two independent networks. Our solution benefits from the limited exploration space for background and reflection information defined by the size of the latent code, and hence tends to overfit less on the synthetic images.

\setlength{\tabcolsep}{5pt}
\begin{table*}[t]
	\begin{center}
		\small
		\begin{tabular}{lc ccc cc cc}
			\toprule
			Data source& Collected Real & \multicolumn{3}{c}{SIR$^2$} & \multicolumn{2}{c}{\revise{Zhang}} & \multicolumn{2}{c}{\revise{Nature}} \\
			\midrule
			Error metric & \% & PSNR & SSIM & SSIM$_r$ & PSNR & SSIM & PSNR & SSIM \\
			\midrule
			Input & - & 26.19 & 0.916 & 0.822 & 18.80 & 0.765 & 22.82 & 0.786 \\
			\midrule
			\cite{wan2018crrn} & - & 22.16 & 0.828 & 0.821 & 18.46 & 0.710 & 20.93 & 0.776 \\
			\cite{li2020single} & - & 22.96 & \textcolor{red}{0.900} & 0.816 & 18.92 & \textcolor{blue}{0.770} & 22.30 & \textcolor{blue}{0.787} \\
			\cite{wen2019single} & 0.28 & 21.23 & 0.854 & 0.779 & 16.81 & 0.664 & 19.57 & 0.702 \\	
			\cite{yang2018seeing} & 9.07 & 22.25 & 0.853 & \textcolor{blue}{0.832} & 17.84 & 0.738 & 20.39 & 0.762 \\
			\cite{li2014single} & 2.97 & 18.87 & 0.782 & 0.761 & 17.30 & 0.727 & 18.58 & 0.725 \\
			\cite{fan2017generic} & 8.50 & 20.97 & 0.839 & 0.775 & 18.43 & 0.760 & 20.78 & 0.740 \\
			\cite{zhang2018single} & \textcolor{blue}{19.83} & 21.20 & 0.862 & 0.807 & 18.32 & 0.750 & 21.82 & 0.768 \\
			\cite{wei2019single} & 6.52 & \textcolor{red}{24.14} & \textcolor{blue}{0.879} & 0.819 & \textcolor{red}{19.01} & 0.767 & \textcolor{blue}{22.71} & 0.786\\
			\midrule
			Ours* & 18.13 & 22.45 & 0.778 & 0.765 & 18.54 & 0.753 & 21.85 & 0.752 \\
			Ours & \textcolor{red}{34.70} & \textcolor{blue}{23.24} & {0.870} & \textcolor{red}{0.833} & \textcolor{blue}{19.00} & \textcolor{red}{0.771} & \textcolor{red}{22.91} & \textcolor{red}{0.788} \\
			\bottomrule
		\end{tabular}
	\end{center}
	\caption{Quantitative comparison on the user study of strong real reflections with our subtraction model (left), and image quality evaluation of the wild scene images of SIR$^2$ dataset with our linear addition model (right). \revise{{Collected Real} refers to our collected 30 real-world reflection images with strong reflection artifacts. The {Input} row refers to the results that are computed by comparing input reflection images with the ground truth background with given error metrics.} Ours* refers to the version with ground truth supervision. The numbers in red and blue are the best and second best results respectively.}
	\label{table:experiment}
\end{table*}

\revise{
\subsection{Extension to Supervised Learning}
Since our network structure is a simple FCN, it can be naturally transformed for the single image supervised settings with available ground truth labels. We extend our algorithm to a supervised learning version, denoted as Ours*.
To be specific, the supervised objective follows \cite{wang2018pix2pixHD} and is a combination of three loss functions.
\begin{equation*} 
\mathcal{L}_{super} = \lambda_1 \mathcal{L}_{adv} + \lambda_2 \mathcal{L}_{feat\_D} + \mathcal{L}_{feat\_VGG}
\end{equation*}

Considering our network as a generator $G$, we incorporate a discriminator $D$, which is trained in an alternating manner along with $G$ to solve the adversarial min-max problem:
\begin{equation*}\label{equation:minimax}
\min_{G} \max_{D} (\log D(I,B) + \log (1-D(I,G(I)))),
\end{equation*}
then the adversarial loss reads
\begin{equation*} \label{equation:adversarial}
\mathcal{L}_{adv} = \log (D(I,G(I))).
\end{equation*}

A feature matching loss is appended based on the learned discriminator, 
\begin{equation*}
\mathcal{L}_{feat\_D} = \sum_{i=1}^M ||D^{i}(I,B)-D^{i}(I,G(I))||_1
\end{equation*}
where $M$ is the total number of layers in $D$.
Similarly, we add another VGG feature matching loss as in \cite{wang2018pix2pixHD},
\begin{equation*}
\mathcal{L}_{feat\_VGG} = \sum_{i=1}^N ||F^{i}(B)-F^{i}(G(I))||_1
\end{equation*}
where $F^{i}$ denotes the ``conv$i$\_2'' layer in the pretrained VGG network, and $N = 5$.
}

\section{Experiments}

Due to the lack of large-scale high-quality reflection image datasets, the learning-based approaches mostly synthesize reflection images for training. Currently, there are two mainstream data synthesis methods, which mainly differ in the logic of computing the reflection layer: (1) One of the methods is developed by \cite{fan2017generic} and followed by \cite{zhang2018single,wei2019single}. Their reflection layer is generated via subtraction and clipping operations. It reflects some physical properties observed in natural scenes. The generated images tend to contain strong and blurry reflections. (2) The other method is the common linear additive image mixing model used by \cite{wan2018crrn,yang2018seeing}. Their reflection layer is only scaled by a small factor for direct composition. The resultant reflections are weaker yet sharper.

In order to conduct a fair comparison with corresponding previous works, and as these approaches have their specialties in ithe mage formation process and bias towards target images, we split our experiments by different training data generated from the above two synthesis methods for our algorithm\footnote{Note \cite{wen2019single} learns the alpha mask in the linear composition model for reflection synthesis, which we also experiment with, but doesn't work better than the image model in our problem setting.}.

\begin{figure}[tbp]
\centering
\small
	\begin{framed}
		\vspace{-3pt}
		\begin{enumerate}
			\itemsep+0.1em
			\item $\mathbf{R} \leftarrow \text{gauss\_blur}_{\sigma}(\mathbf{R^\prime})$ with  $\sigma \sim \mathcal{U}(1,5)$
			\item $\mathbf{B} \leftarrow \mathbf{B^\prime}$
			\item $\mathbf{I} \leftarrow \mathbf{B}+\mathbf{R}$
			\item $m \leftarrow \operatorname{mean}(\{\mathbf{I}(\mathbf{x}, c) \mid \mathbf{I}(\mathbf{x}, c)>1, \forall \mathbf{x}, \forall c=1,2,3\})$
			\item $\mathbf{R}(\mathbf{x}, c) \leftarrow {\mathbf{R}}(\mathbf{x}, c)-\gamma \cdot(m-1), \forall \mathbf{x}, \forall c$; $\gamma$ set as 1.1
			\item $\mathbf{R} \leftarrow \operatorname{clip}_{[0,1]}(\mathbf{R})$
			\item $\mathbf{I} \leftarrow \operatorname{clip}_{[0,1]}(\mathbf{B}+\mathbf{R})$
		\end{enumerate}
		\vspace{-6pt}
	\end{framed}
	\vspace{-12pt}
	\caption{\revise{The subtraction and clipping image model for reflection images synthesis.}
	\label{fig:sub_model}}
	\vspace{-5pt}
\end{figure}

\begin{figure}[tbp]
\centering
\small
	\begin{framed}
		\vspace{-3pt}
		\begin{enumerate}
			\itemsep+0.1em
			\item $\mathbf{R} \leftarrow \text{gauss\_blur}_{\sigma}(\mathbf{R^\prime})$ with  $\sigma \sim \mathcal{U}(1,3)$
			\item $\mathbf{B} \leftarrow \mathbf{B^\prime}$
			\item $\mathbf{I} \leftarrow \alpha \cdot \mathbf{B}+ (1-\alpha) \cdot \mathbf{R},$ with $\alpha \sim \mathcal{U}(0.6, 0.8)$
			\item $\mathbf{I} \leftarrow \operatorname{clip}_{[0,1]}(\mathbf{I})$
		\end{enumerate}
		\vspace{-6pt}
	\end{framed}
	\vspace{-12pt}
	\caption{\revise{The linear addition image model for reflection images synthesis.}
	\label{fig:linear_model}}
	\vspace{-5pt}
\end{figure}

\subsection{Subtraction and Clipping Image Model} \label{sec:sub}

\noindent\textbf{Implementation Details.}  Following \cite{fan2017generic}, we collect around 17,000 natural images from PASCAL VOC dataset as the data source of the image synthesis approach, while 5\% of them are treated as test data and the left are training data. 
\revise{To synthesize the multiple reflection images for training, we freeze the background layer and only change the reflection layer. To be specific, we randomly choose a natural image $B^\prime$ and $n$ different natural images $R_1^\prime, R_2^\prime, ..., R_n^\prime$ to represent the background and reflection respectively. We feed each background-reflection pair $<B^\prime,R_i^\prime>$ to the image synthesis models to generate the reflection image.
The subtraction and clipping image model is proposed in \cite{fan2017generic} which avoids the brightness overflow issue by subtracting an adaptively computed value followed by the clipping operation. The process is summarized in Figure \ref{fig:sub_model}.}

Our framework is implemented in PyTorch. Our deep network is optimized by Adam with mini-batch size as 8. Some specifics about our training setting: initial learning rate (0.01), training epoch number (60), $\lambda_1$ (15), $\lambda_2$ (20).

\setlength{\tabcolsep}{1pt}
\renewcommand{\arraystretch}{1}
\begin{figure*}[t]
	\begin{center}
		
		\begin{tabular}{cccc cccc}
			\vspace{-0.6mm}
			\includegraphics[width=2.18cm]{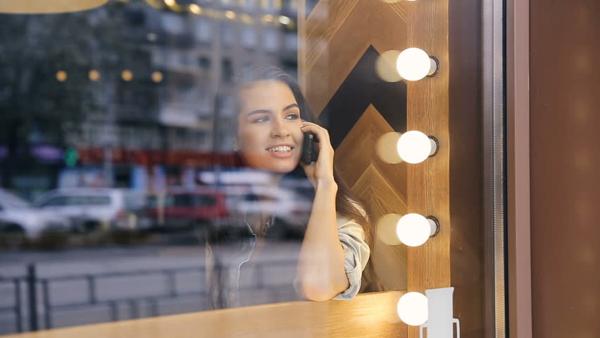}
			&\includegraphics[width=2.18cm]{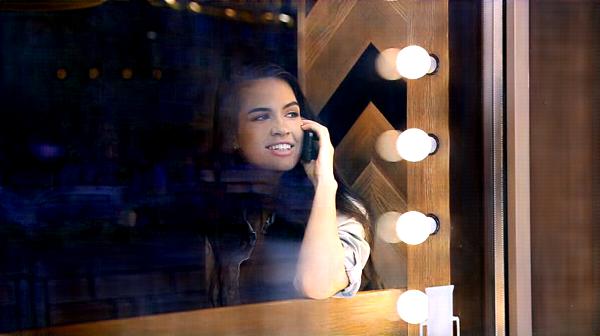}
			&\includegraphics[width=2.18cm]{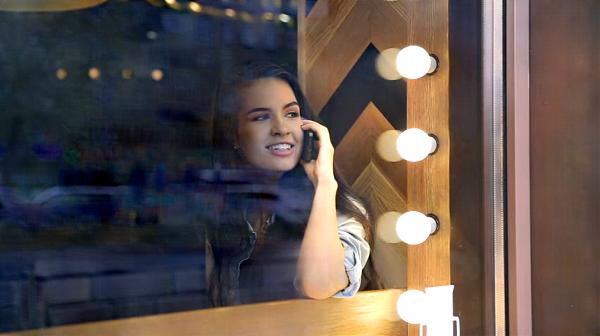}
			&\includegraphics[width=2.18cm]{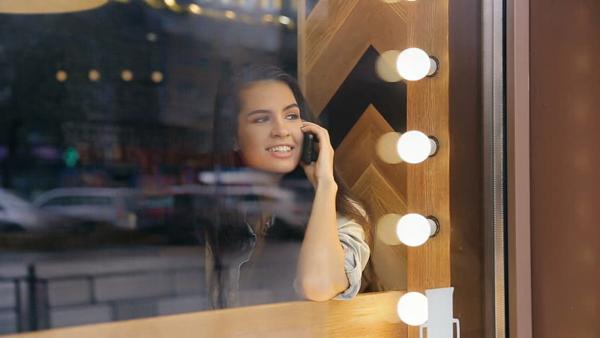}
			&\includegraphics[width=2.18cm]{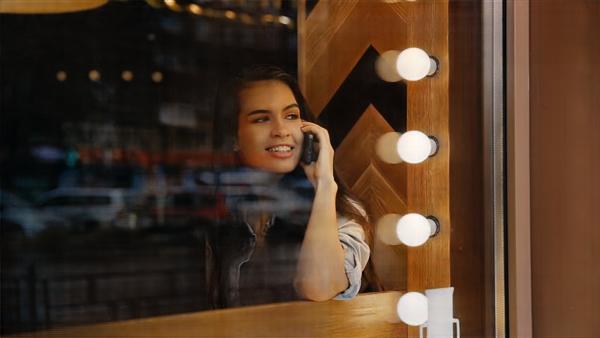}
			&\includegraphics[width=2.18cm]{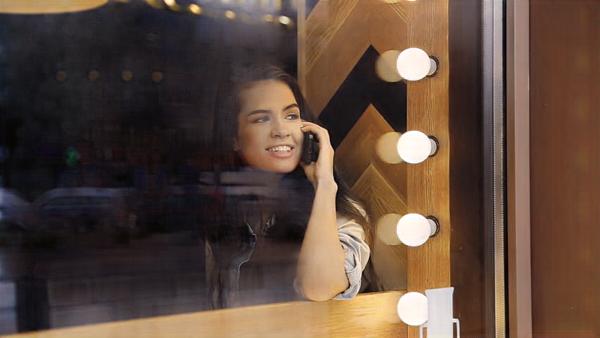}
			&\includegraphics[width=2.18cm]{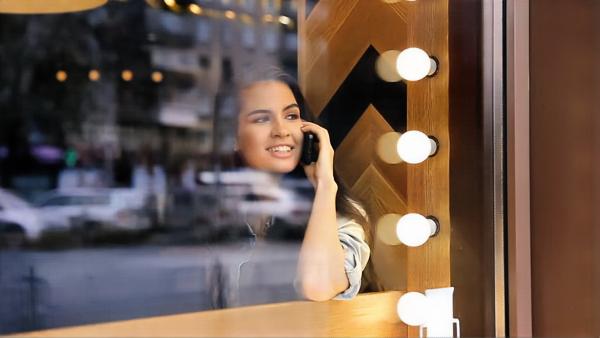}
			&\includegraphics[width=2.18cm]{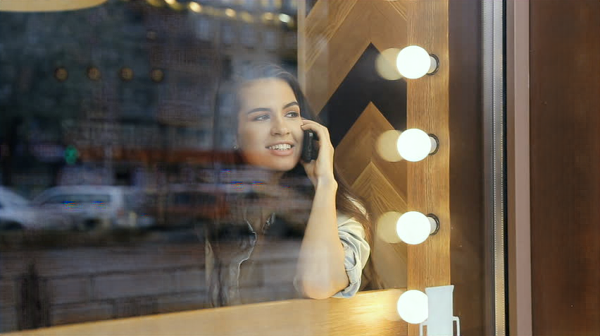}
			\\

			\vspace{-0.6mm}
			\includegraphics[width=2.18cm]{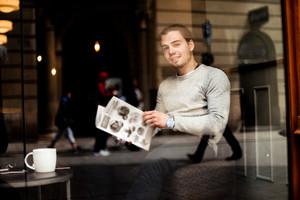}
			&\includegraphics[width=2.18cm]{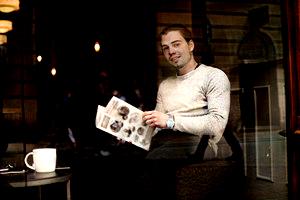}
			&\includegraphics[width=2.18cm]{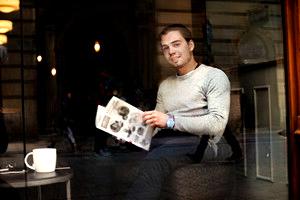}
			&\includegraphics[width=2.18cm]{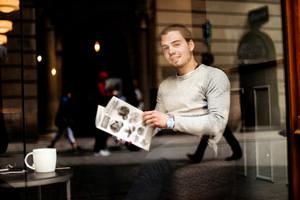}
			&\includegraphics[width=2.18cm]{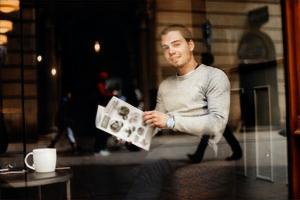}
			&\includegraphics[width=2.18cm]{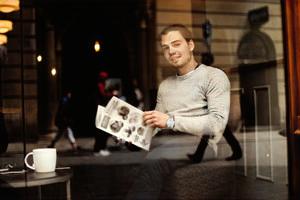}
			&\includegraphics[width=2.18cm]{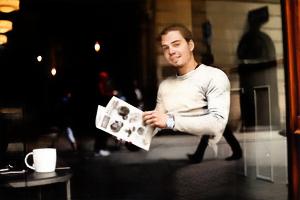}			
			&\includegraphics[width=2.18cm]{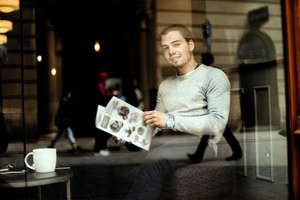}
			\\
			\vspace{-0.6mm}
			\includegraphics[trim={0 0 0 1.2cm},clip,width=2.18cm]{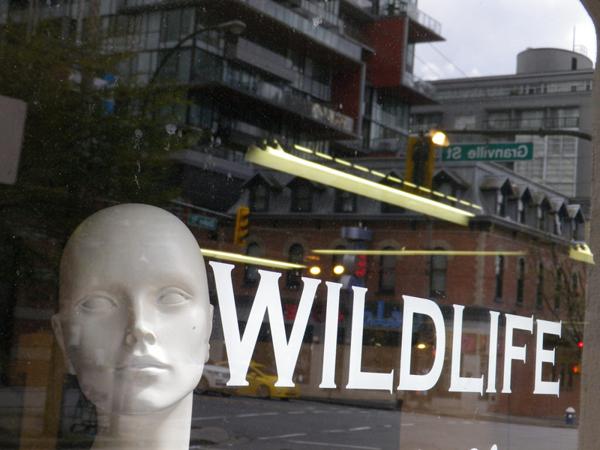}
			&\includegraphics[trim={0 0 0 1.2cm},clip,width=2.18cm]{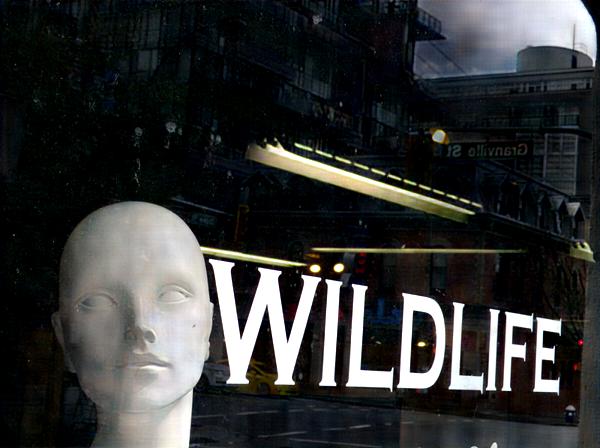}
			&\includegraphics[trim={0 0 0 1.2cm},clip,width=2.18cm]{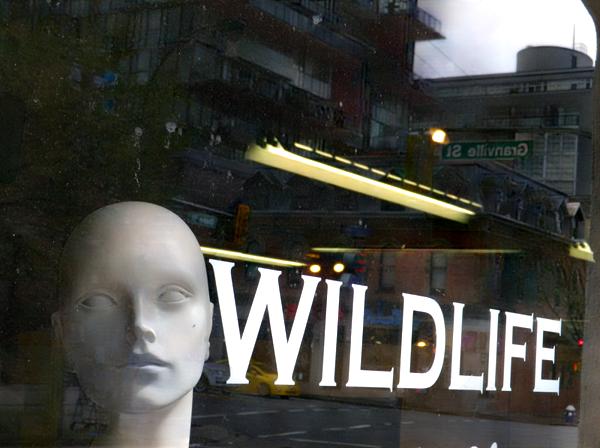}
			&\includegraphics[trim={0 0 0 1.2cm},clip,width=2.18cm]{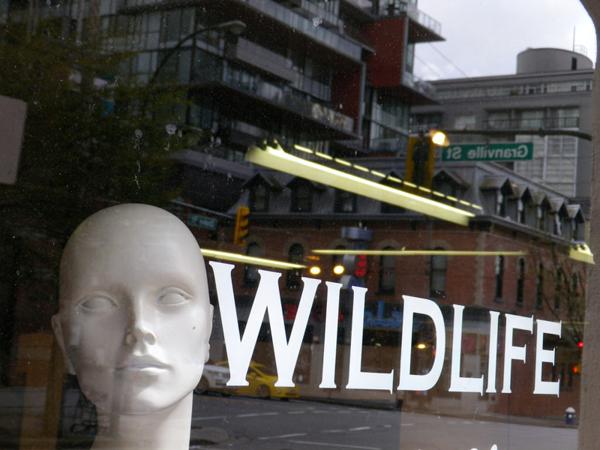}
			&\includegraphics[trim={0 0 0 1.2cm},clip,width=2.18cm]{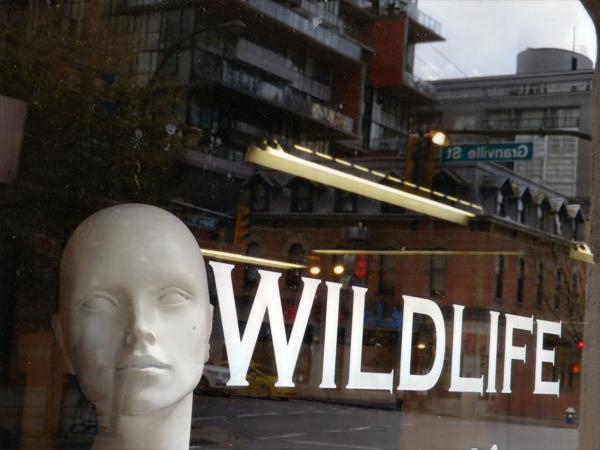}
			&\includegraphics[trim={0 0 0 1.2cm},clip,width=2.18cm]{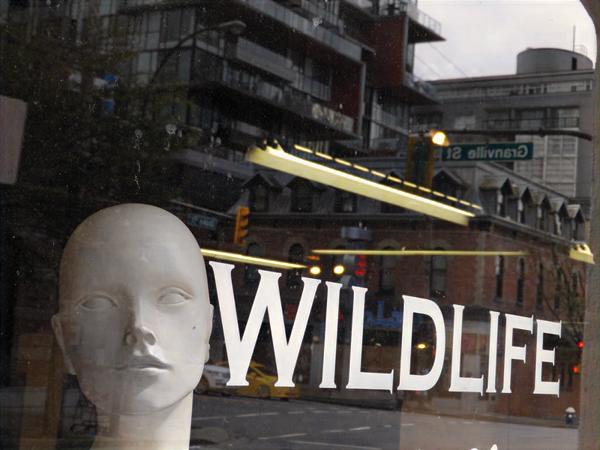}
			&\includegraphics[trim={0 0 0 1.2cm},clip,width=2.18cm]{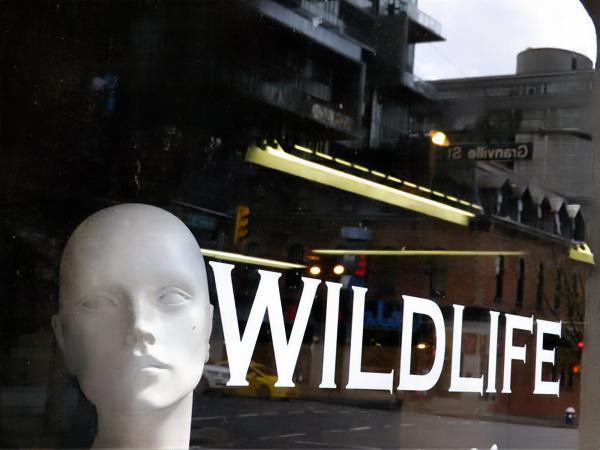}
			&\includegraphics[trim={0 0 0 1.2cm},clip,width=2.18cm]{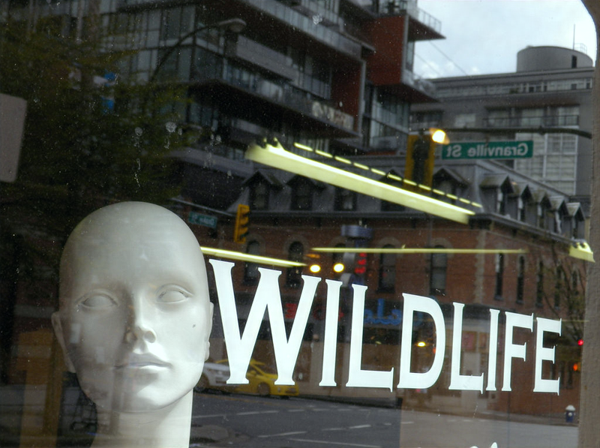}
			\\	
			\vspace{-0.6mm}
			\includegraphics[width=2.18cm]{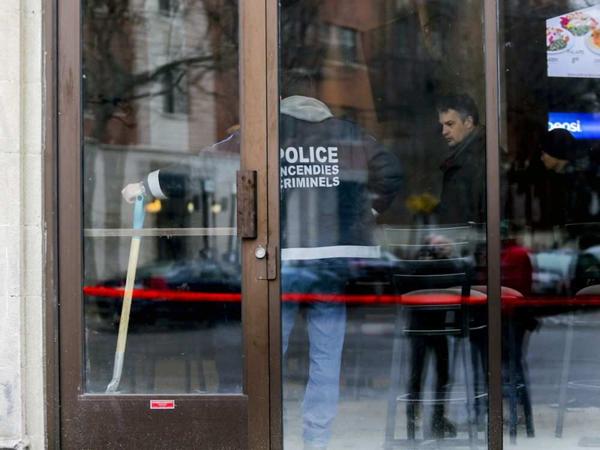}
			&\includegraphics[width=2.18cm]{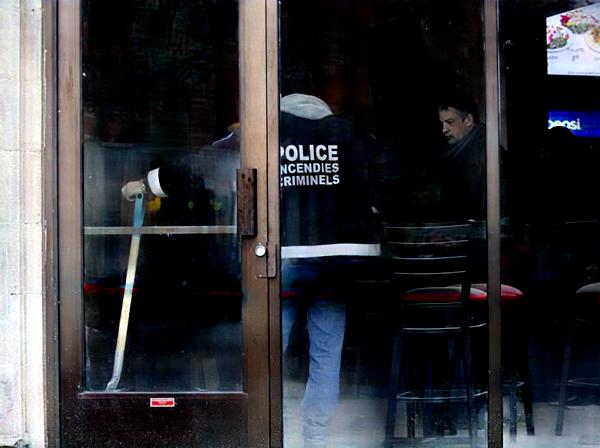}
			&\includegraphics[width=2.18cm]{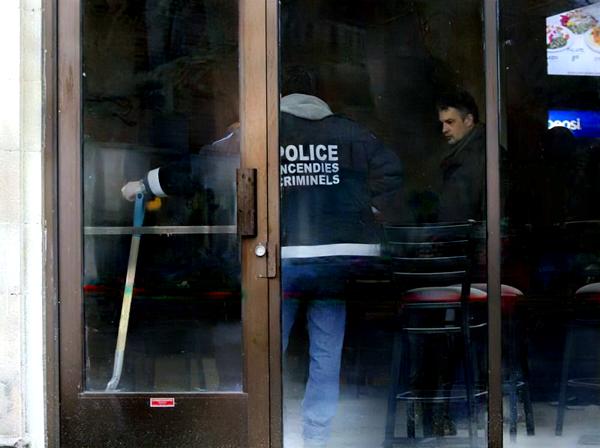}
			&\includegraphics[width=2.18cm]{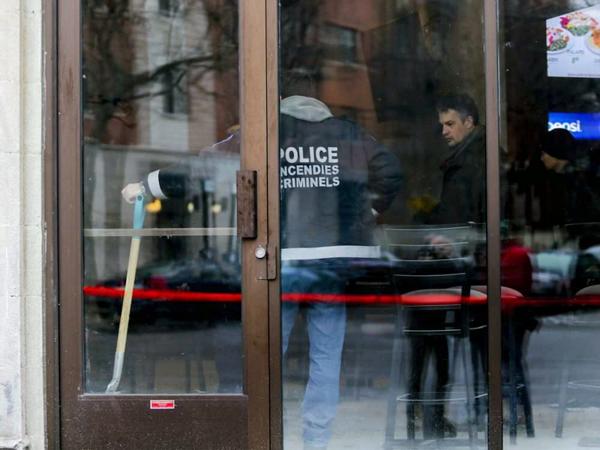}
			&\includegraphics[width=2.18cm]{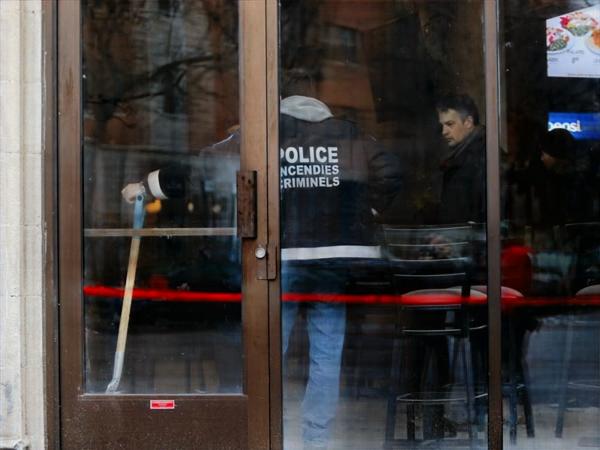}
			&\includegraphics[width=2.18cm]{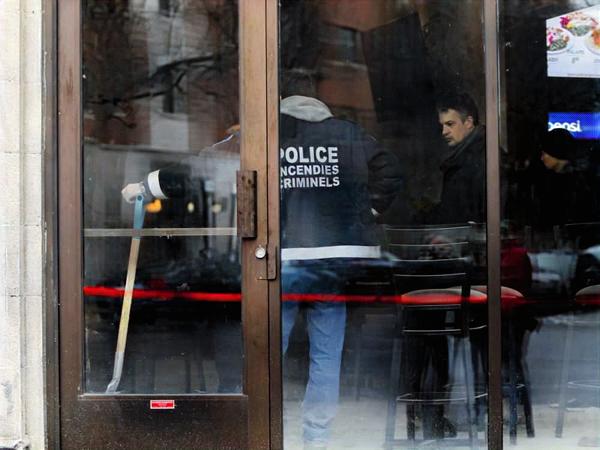}
			&\includegraphics[width=2.18cm]{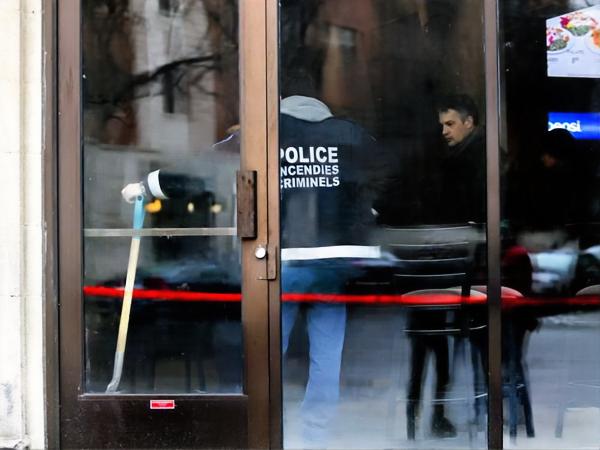}
			&\includegraphics[width=2.18cm]{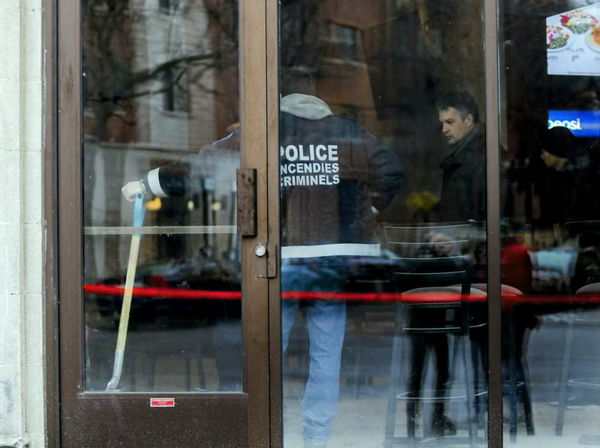}
			\\
			\small{Input} & \small{Ours} & \small{Ours*} & \small{\cite{wei2019single}} & \small{\cite{zhang2018single}} & \small{\cite{fan2017generic}} & \small{\cite{yang2018seeing}} &
			\revise{\small{\cite{li2020single}}}
			\\
		\end{tabular}
		
	\end{center}
	\vspace{-3mm}
	\caption{Visual comparison on strong real-world reflection images collected by ourselves. These images are featured by subtraction and clipping image model. Ours* refers to the version with ground truth supervision.}
	\label{figure:collected}
\end{figure*}

\noindent\textbf{Results.} 
To evaluate our algorithm on the real images, we randomly collect around 30 reflection images that feature strong and blurry reflections, and conduct a user study among 40 users. They are asked to pick the best visual result among all approaches for each example\footnote{Due to inaccessibility to \cite{wan2018crrn}'s codes or trained model, their results are not present for the subtraction and clipping image model.}, and we evaluate the user selections for each method in Table \ref{table:experiment}. \revise{Input reflection images are not included in the user study.}
The corresponding qualitative results are demonstrated in Figure \ref{figure:fanICCV} (\cite{fan2017generic}'s test images) and Figure \ref{figure:collected} (self-collected images).

For both numerical and visual results, our proposed pipeline outperforms the others. Note on the visual side, our algorithm learns to remove more reflections and generate cleaner background, even compared with \cite{zhang2018single,fan2017generic,wei2019single} that share the similar data generation approach as ours. Since the demonstrated reflections are very strong, they are not handled very well by \cite{yang2018seeing} due to different types of training data used for their deep network. \cite{li2014single} is a popular optimization-based algorithm that relies on the small gradient prior on the reflection layer ($R$) to solve this problem, and hence does not work well in these challenging cases of strong reflections. For many more visual results, please refer to the supplemental material.

\setlength{\tabcolsep}{1pt}
\renewcommand{\arraystretch}{1}
\begin{figure*}[t]
	\begin{center}
		
		\begin{tabular}{cccc cccc}
			\vspace{-0.6mm}
			\includegraphics[width=2.18cm]{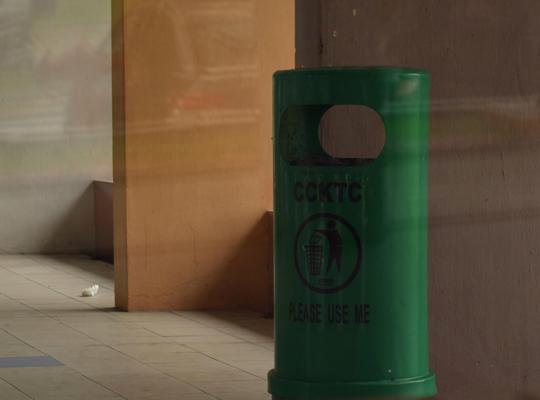}
			&\includegraphics[width=2.18cm]{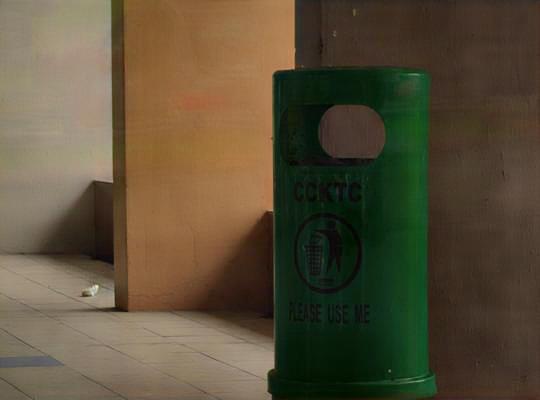}
			&\includegraphics[width=2.18cm]{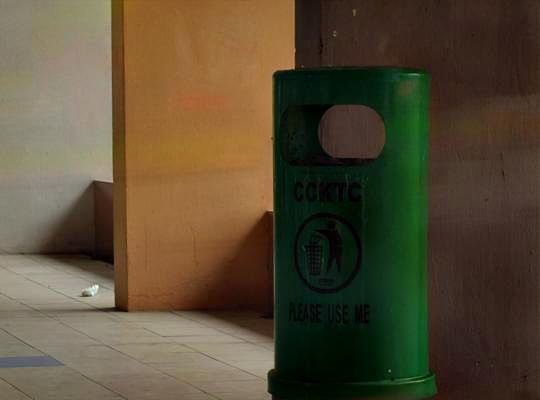}
			&\includegraphics[width=2.18cm]{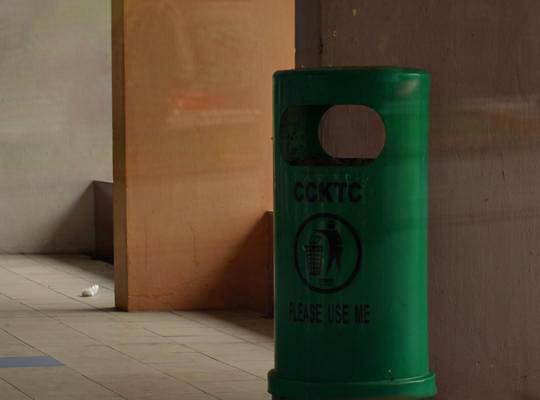}
			&\includegraphics[width=2.18cm]{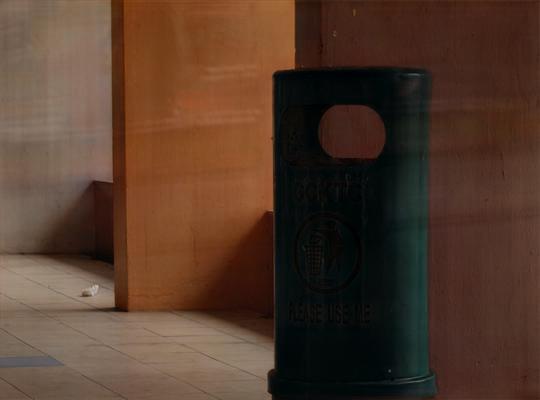}
			&\includegraphics[width=2.18cm]{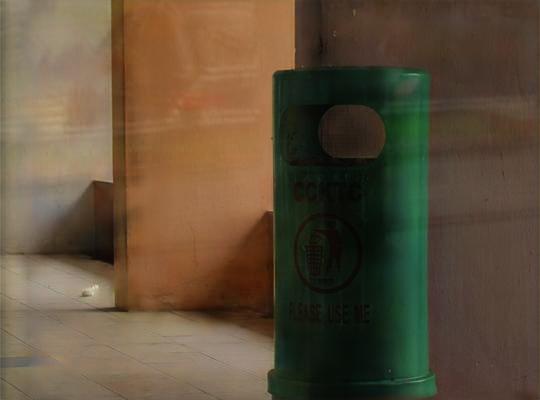}
			&\includegraphics[width=2.18cm]{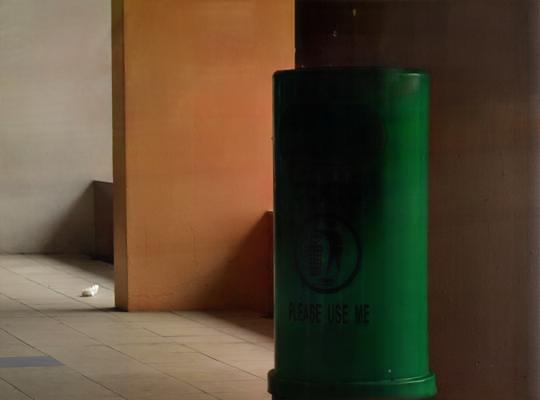}		
			&\includegraphics[width=2.18cm]{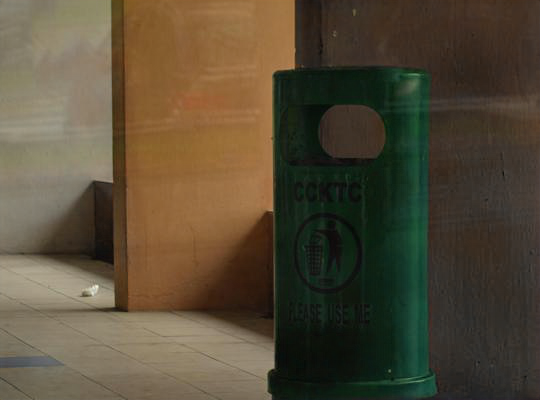}
			\\
			&\includegraphics[width=2.18cm]{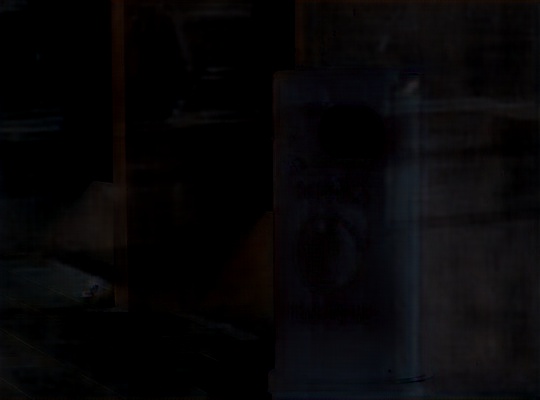}
			&\includegraphics[width=2.18cm]{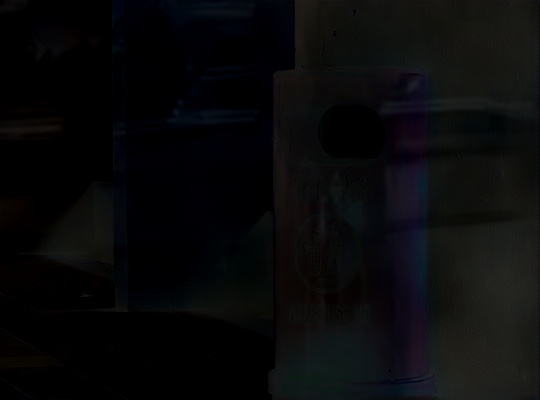}
			&\includegraphics[width=2.18cm]{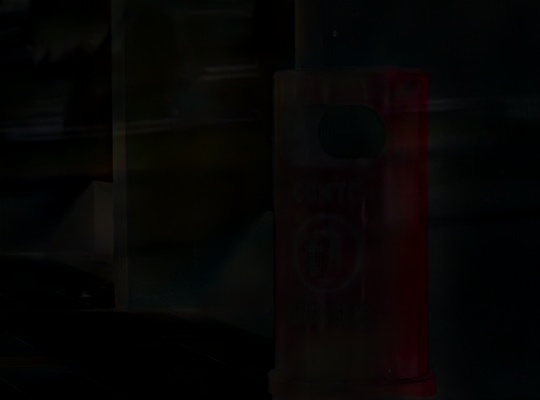}
			&\includegraphics[width=2.18cm]{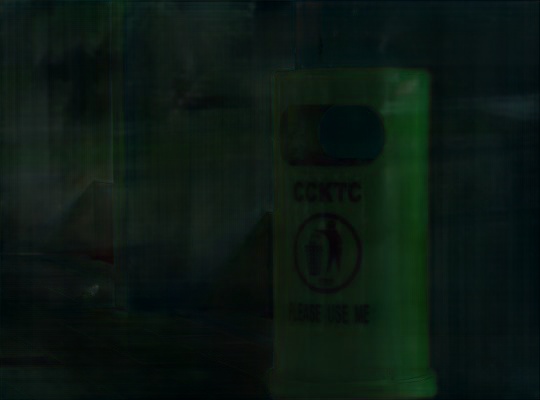}
			&\includegraphics[width=2.18cm]{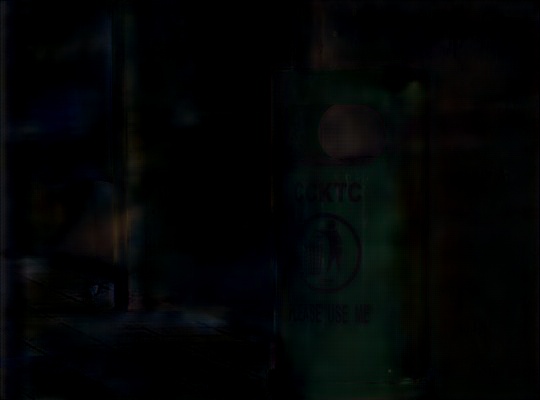}
			&\includegraphics[width=2.18cm]{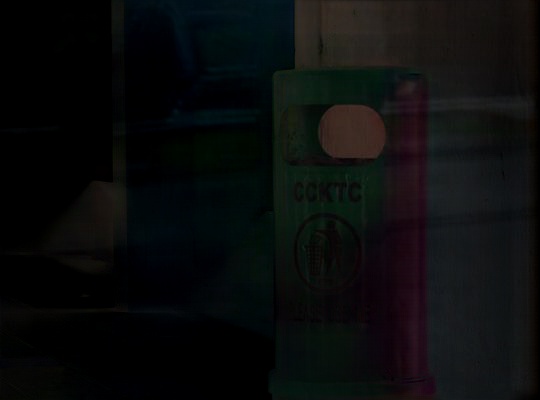}
			&\includegraphics[width=2.18cm]{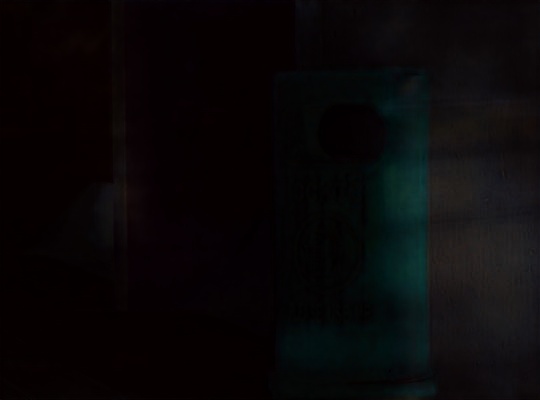}	
			\\

			\vspace{-0.6mm}
			\includegraphics[width=2.18cm]{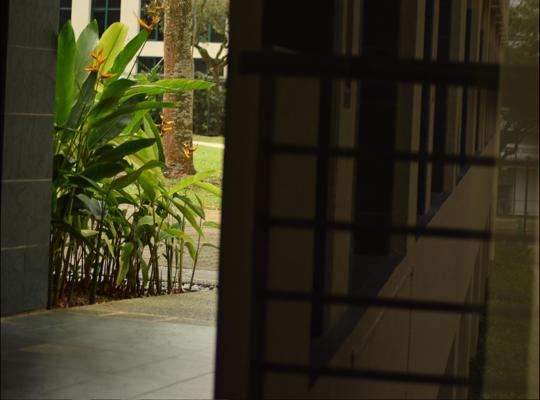}
			&\includegraphics[width=2.18cm]{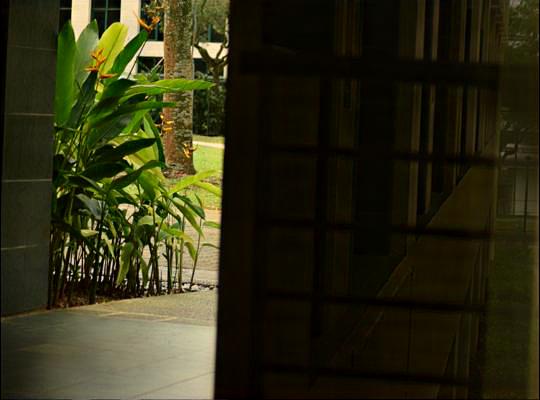}
			&\includegraphics[width=2.18cm]{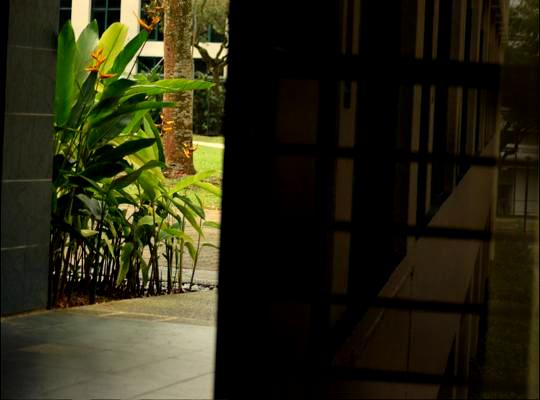}
			&\includegraphics[width=2.18cm]{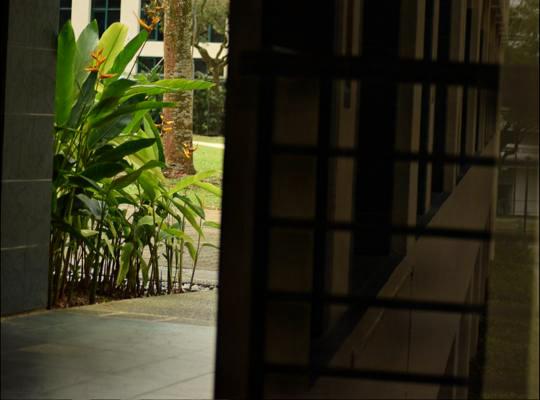}
			&\includegraphics[width=2.18cm]{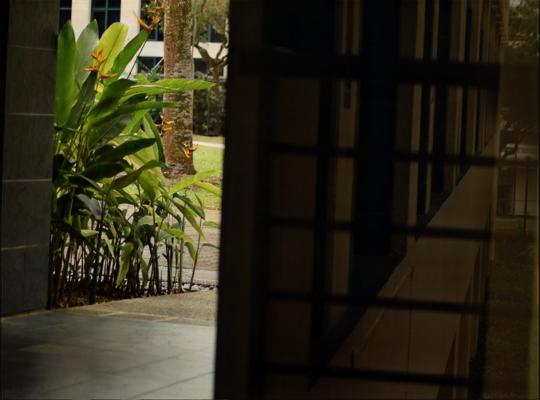}
			&\includegraphics[width=2.18cm]{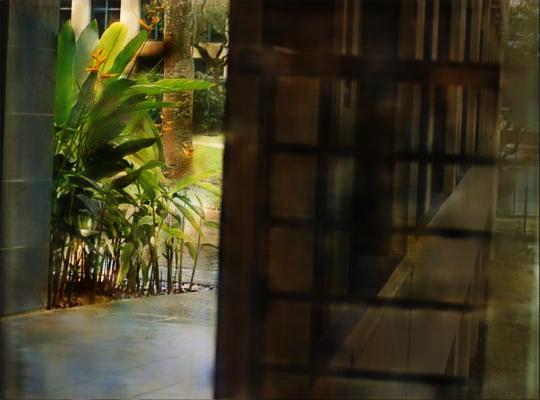}
			&\includegraphics[width=2.18cm]{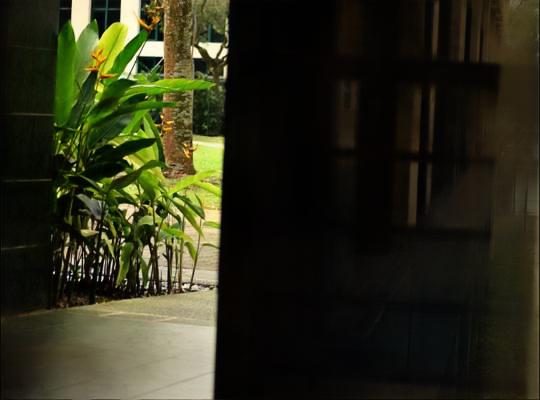}
			&\includegraphics[width=2.18cm]{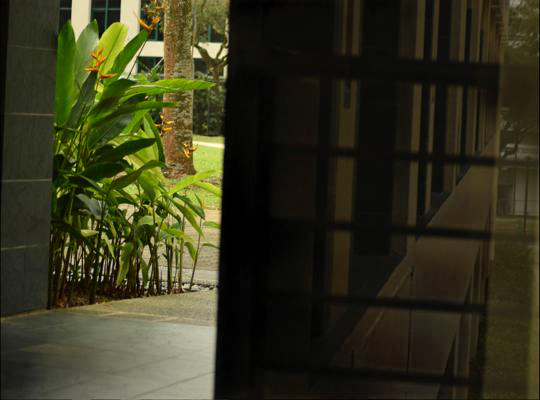}
			\\
			
			&\includegraphics[width=2.18cm]{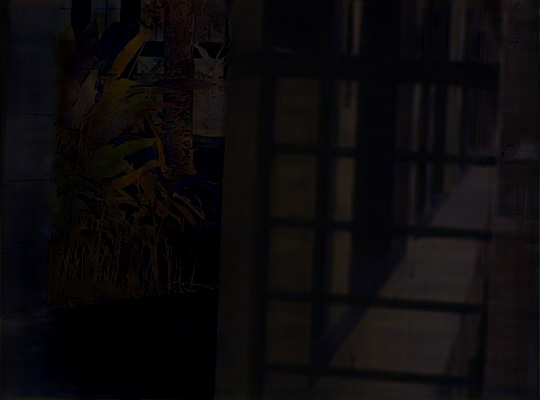}
			&\includegraphics[width=2.18cm]{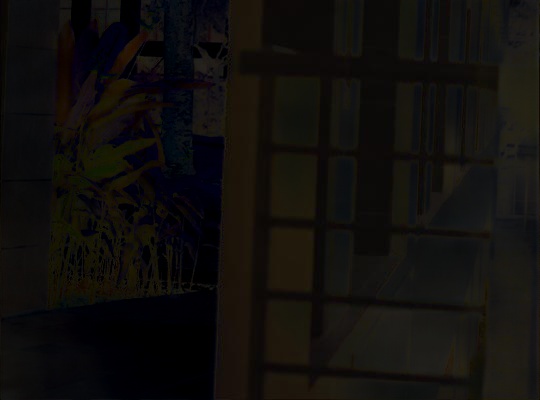}
			&\includegraphics[width=2.18cm]{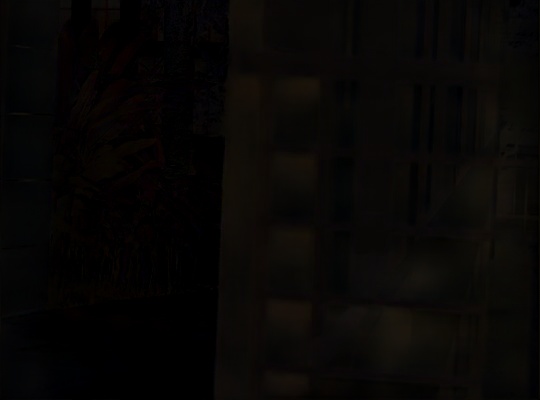}
			&\includegraphics[width=2.18cm]{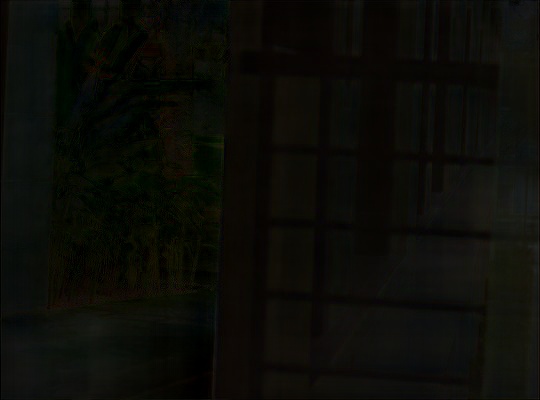}
			&\includegraphics[width=2.18cm]{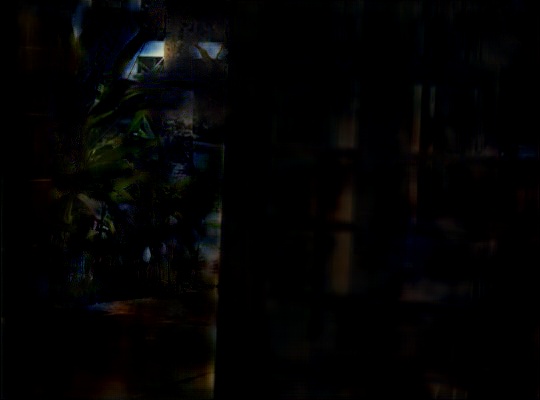}
			&\includegraphics[width=2.18cm]{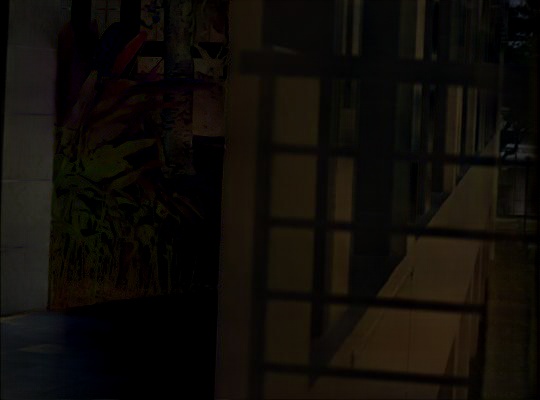}
			&\includegraphics[width=2.18cm]{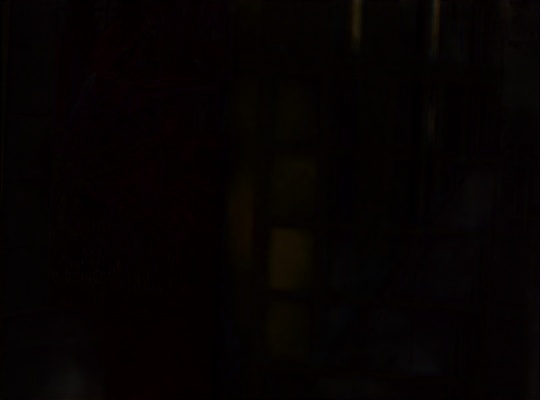}
			\\

			\vspace{-0.6mm}			
			\includegraphics[width=2.18cm]{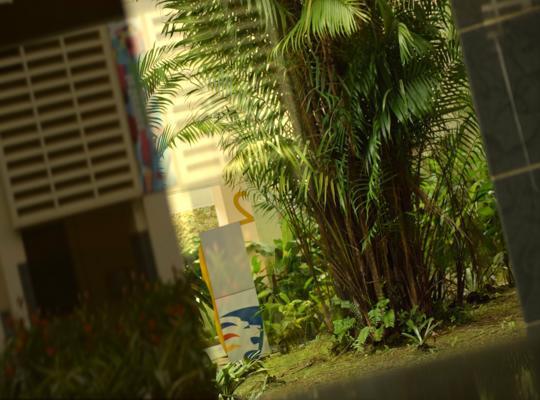}
			&\includegraphics[width=2.18cm]{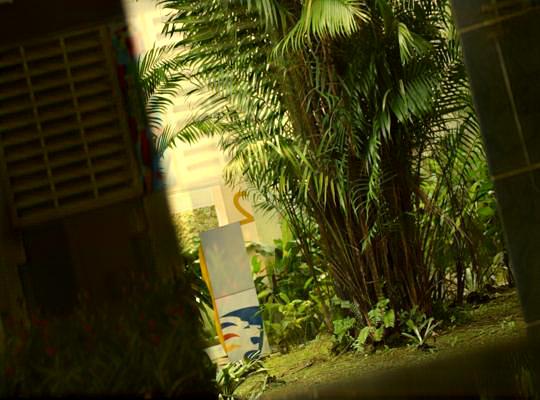}
			&\includegraphics[width=2.18cm]{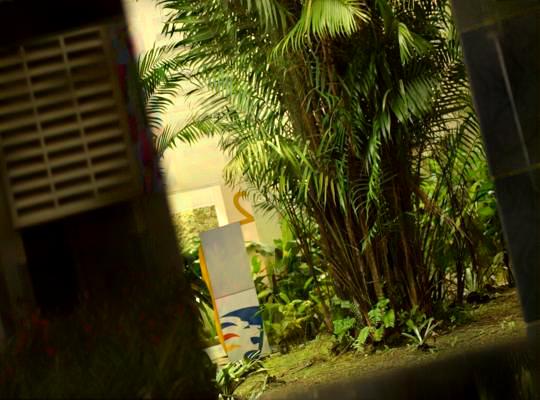}
			&\includegraphics[width=2.18cm]{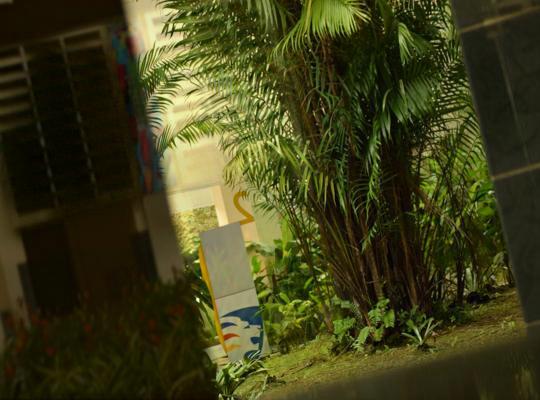}
			&\includegraphics[width=2.18cm]{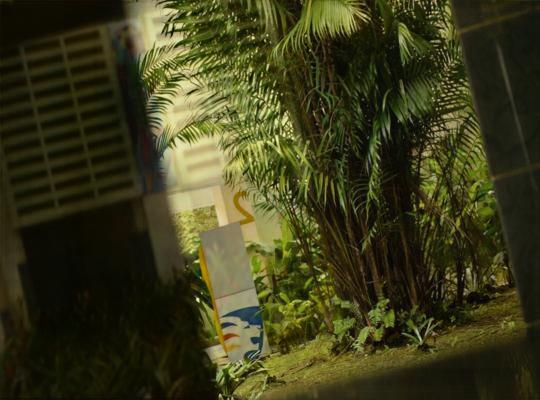}
			&\includegraphics[width=2.18cm]{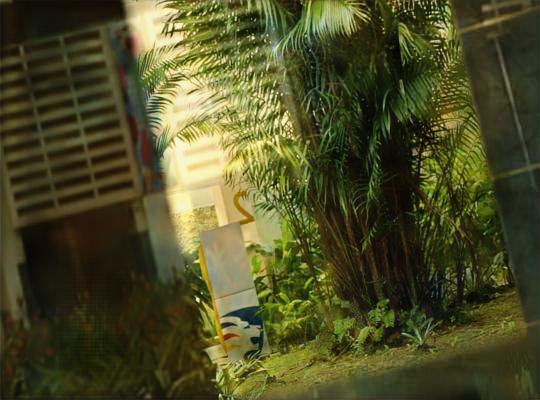}
			&\includegraphics[width=2.18cm]{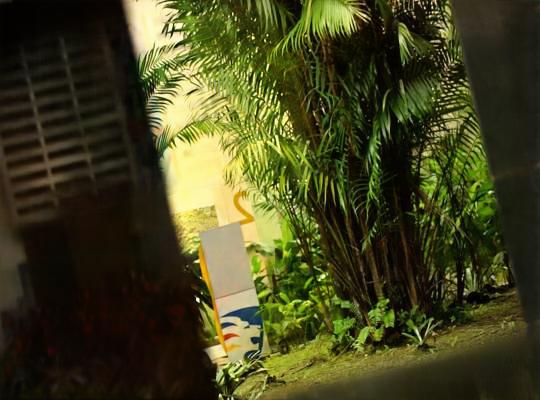}
			&\includegraphics[width=2.18cm]{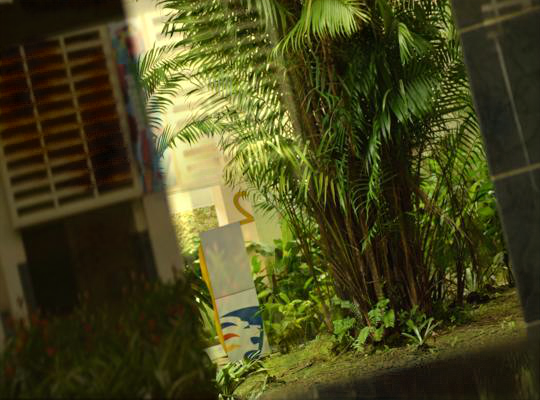}
			\\
			
			&\includegraphics[width=2.18cm]{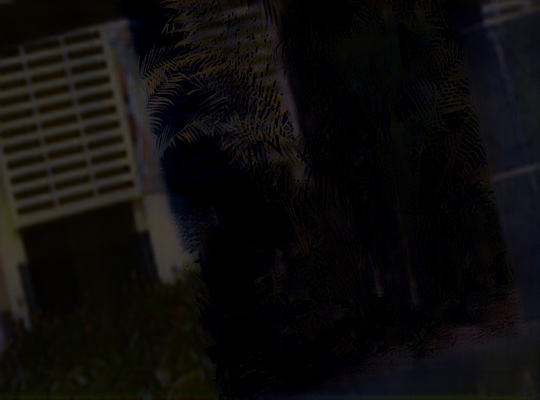}
			&\includegraphics[width=2.18cm]{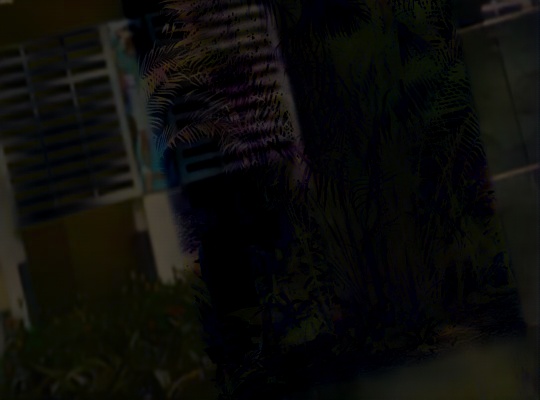}
			&\includegraphics[width=2.18cm]{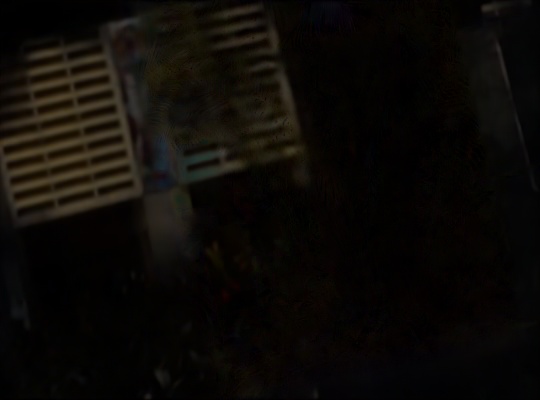}
			&\includegraphics[width=2.18cm]{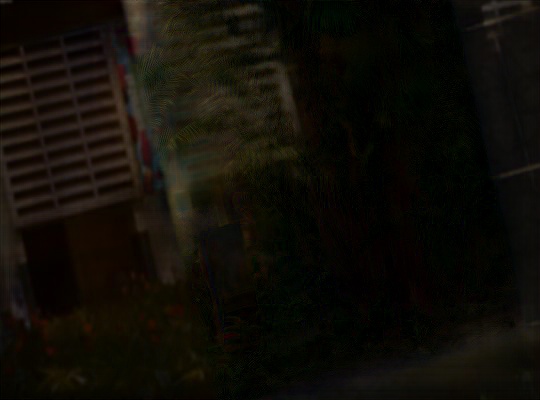}
			&\includegraphics[width=2.18cm]{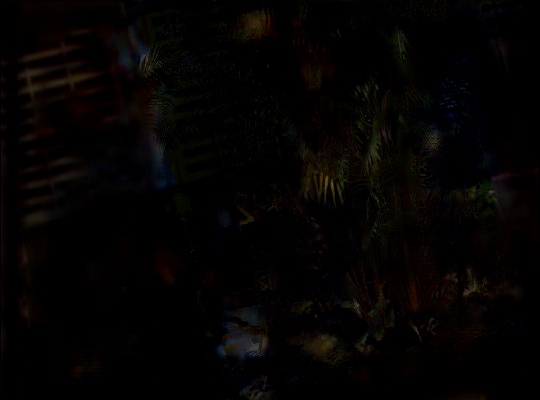}
			&\includegraphics[width=2.18cm]{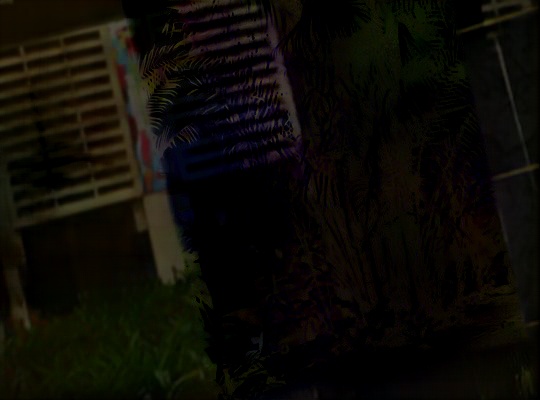}
			&\includegraphics[width=2.18cm]{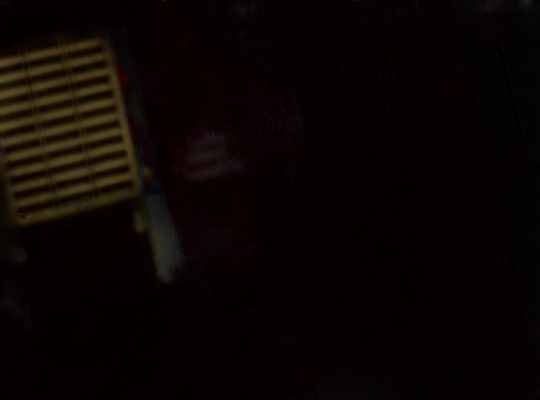}
			\\

			\vspace{-0.6mm}
			\small{Input} & \small{Ours} & \small{Ours*} & \small{\cite{wei2019single}} & \small{\cite{zhang2018single}} & \small{\cite{wen2019single}} & \small{\cite{yang2018seeing}} & \small{\cite{li2020single}}
			\\
		\end{tabular}
		
	\end{center}
	\vspace{-3mm}
	\caption{Visual comparison \revise{of the predicted background (top) and reflection (bottom) layers} on the wild scene images of SIR$^2$ dataset. Ours* refers to the version with ground truth supervision. \textbf{Zoom in to see the details.}}
	\label{figure:wanICCV}
\end{figure*}

\subsection{Linear Addition Image Model}

\noindent\textbf{Implementation Details.} 
Similar to the subtraction and clipping image model, we directly take their train/test split in the PASCAL VOC dataset as our data source. For a fair comparison with the previous work, we follow \cite{yang2018seeing}'s approach to linearly mix two natural images with a constant scale factor to synthesize the reflection image, \revise{summarized in Figure \ref{fig:linear_model}.} We adopt the similar training setting as in the subtraction and clipping image model. The difference lies in: training epoch number (30), $\lambda_1$ (80), $\lambda_2$ (50). The parameter change is caused by the significantly different data generation approach.

\noindent\textbf{Results.} \revise{We conduct experiments on three well-known real-world reflection benchmark datasets, i.e., SIR$^2$ \cite{wan2017benchmarking}, Zhang \cite{zhang2018single} and Nature \cite{li2020single}. We follow \cite{wan2018crrn, li2020single, yang2018seeing} to utilize the linear addition image model to prepare for the training data.
Note that our algorithm requires multiple reflection images with the same background as inputs for training, which is not satisfied by \textit{Zhang} and \textit{Nature} datasets. For a fair comparison, all the methods are trained on the synthetic data in their recommended settings and tested on real data benchmarks.}

We demonstrate the visual results in Figure \ref{figure:wanICCV}. The reflections 
are weaker yet sharper compared to the previous section. However, our approach is still able to generate cleaner background, and also preserves better image structure and color. 
For the first example of the green rubbish can, \cite{yang2018seeing} tends to over-smooth the image, and hence many important details, such as the blank hole and the text on the rubbish can, are inaccurately diminished. 
The second case demonstrates a very challenging scene where there are some reflected blinds, most of which are eliminated by our approach. 

The numerical performances are listed in Table \ref{table:experiment}. 
\revise{As shown in the results, our method performs consistently better among all the benchmark datasets. To be specific, our approach achieves the best result for four columns, and the second best results for two columns. \textit{Ours} (without ground truth supervision) consistently outperforms \textit{Ours*} (with ground truth supervision), further demonstrating the effectiveness of our learning framework and designed objective functions.}

\revise{Interestingly, in SIR$^2$ dataset, we observe somewhat inconsistency between the quantitative and qualitative results in terms of PSNR and SSIM metrics. Also pointed out in \cite{wan2018crrn}, since most existing reflection removal methods tend to touch an image as a whole, more or less, some distortions of the predicted images are observed, including luminance and contrast changes. Due to the regional properties of reflections, the distortions may downgrade the quality of the whole image, although local reflections are removed cleanly.
Therefore, SIR$^2$ dataset provides the manually annotated reflection regions of the input images and proposes SSIM$_r$ to evaluate the SSIM error only within the reflection regions which complements the limitations of the global error metrics. We believe SSIM$_r$ is a more proper measurement to better reflect the performance of reflection removal, and under the evaluation of SSIM$_r$, our approach achieves the best score among all the compared approaches and input images. }

\section{Analysis}

\setlength{\tabcolsep}{0.5pt}
\renewcommand{\arraystretch}{1}
\begin{figure*}[htp]
	\begin{center}
		
		\begin{tabular}{cccc ccc}

			\vspace{-0.6mm}
			\includegraphics[width=2.5cm]{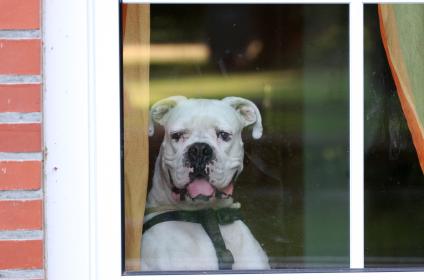}
			&\includegraphics[width=2.5cm]{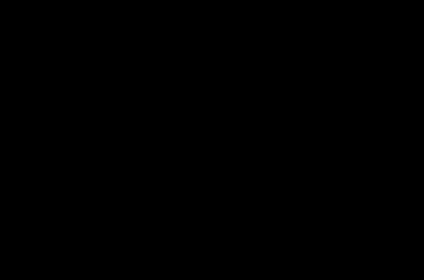}			
			&\includegraphics[width=2.5cm]{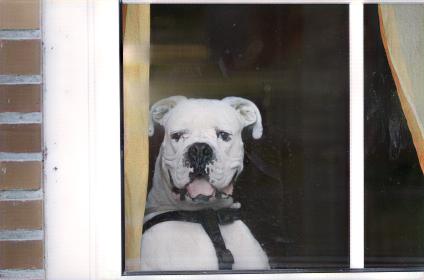}
			&\includegraphics[width=2.5cm]{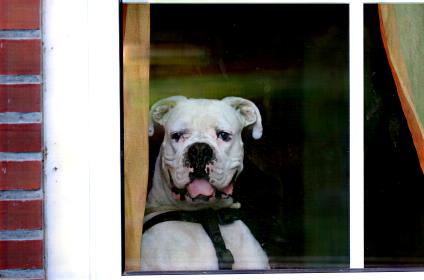}			
			&\includegraphics[width=2.5cm]{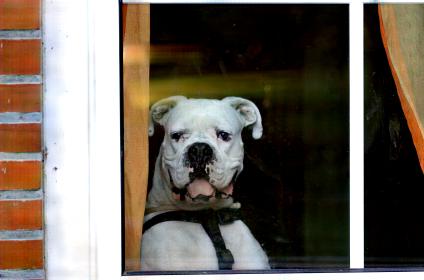}		
			&\includegraphics[width=2.5cm]{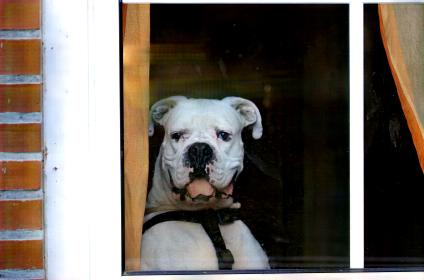}		
			&\includegraphics[width=2.5cm]{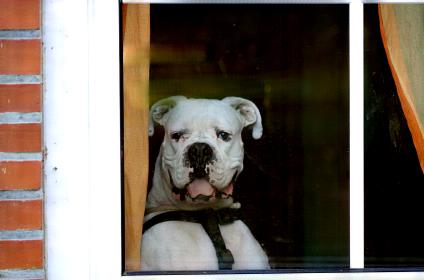}
			\\
			
			\vspace{-0.6mm}
			\includegraphics[width=2.5cm]{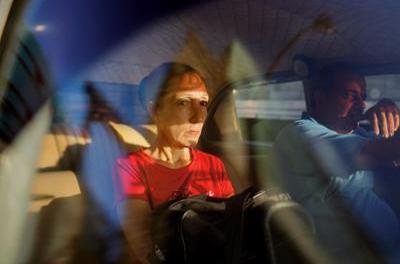}
			&\includegraphics[width=2.5cm]{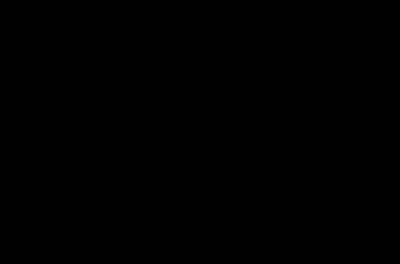}			
			&\includegraphics[width=2.5cm]{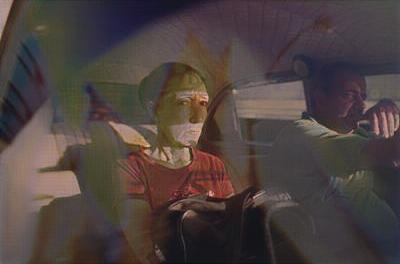}
			&\includegraphics[width=2.5cm]{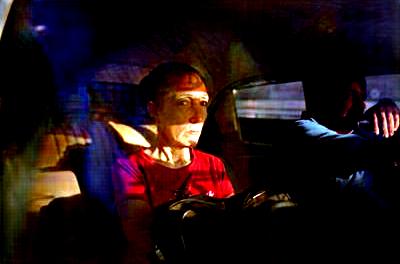}				
			&\includegraphics[width=2.5cm]{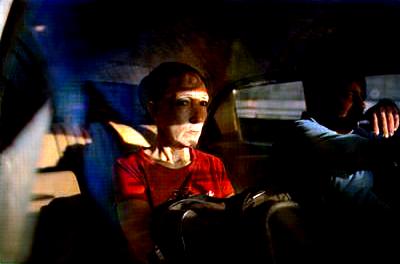}			
			&\includegraphics[width=2.5cm]{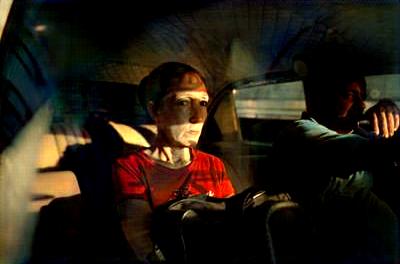}		
			&\includegraphics[width=2.5cm]{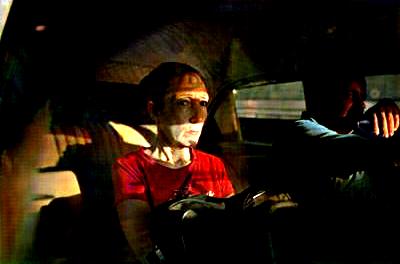}
			\\

			\vspace{-0.6mm}
			\includegraphics[width=2.5cm]{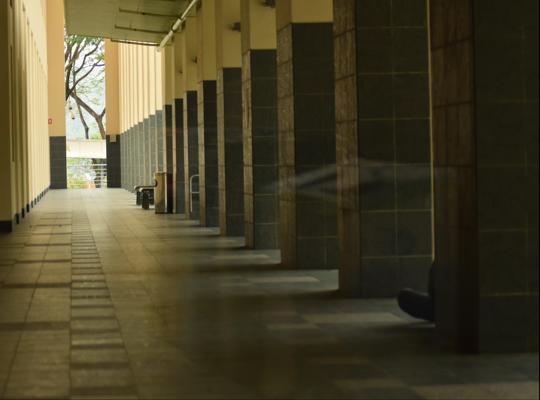}
			&\includegraphics[width=2.5cm]{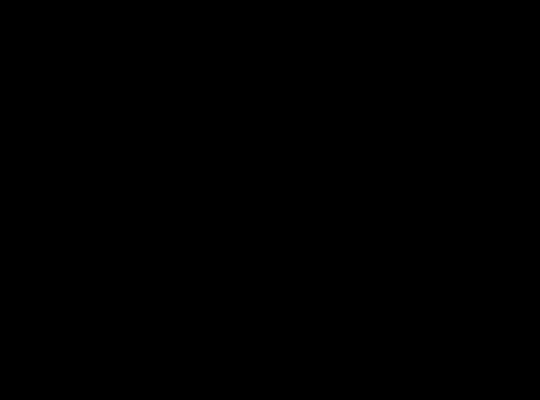}
			&\includegraphics[width=2.5cm]{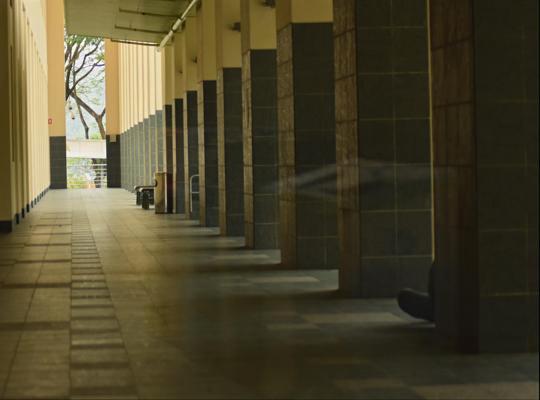}
			&\includegraphics[width=2.5cm]{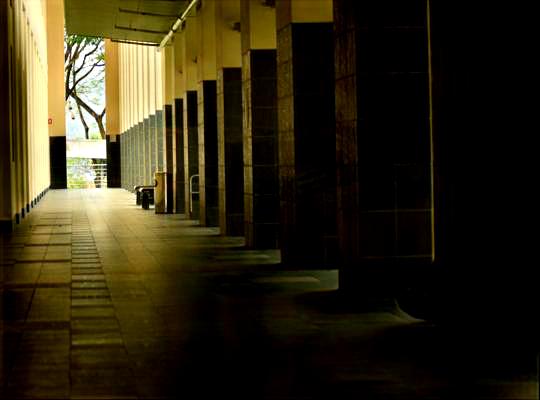}				
			&\includegraphics[width=2.5cm]{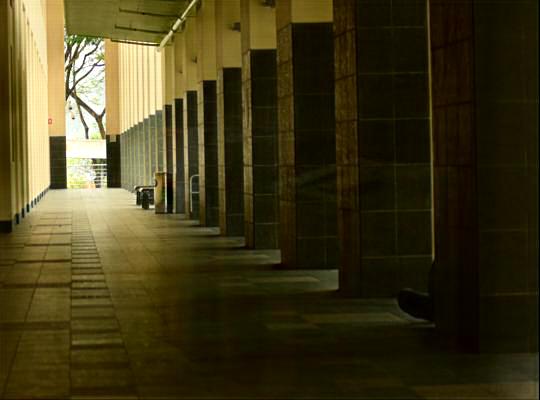}	
			&\includegraphics[width=2.5cm]{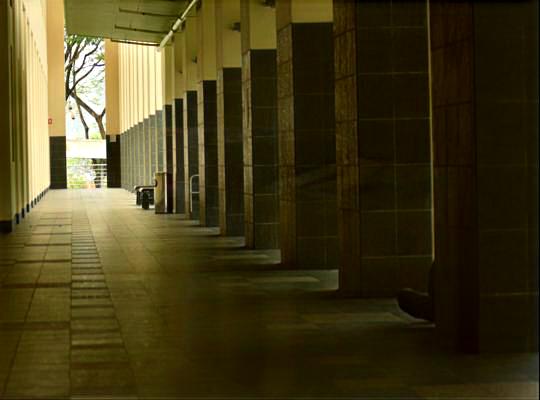}
			&\includegraphics[width=2.5cm]{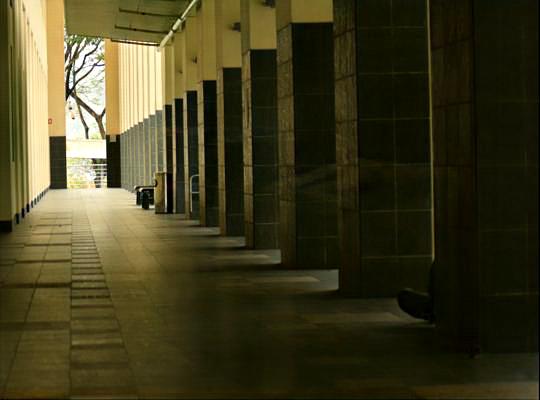}
			\\
			
			\vspace{-0.6mm}
			\includegraphics[width=2.5cm]{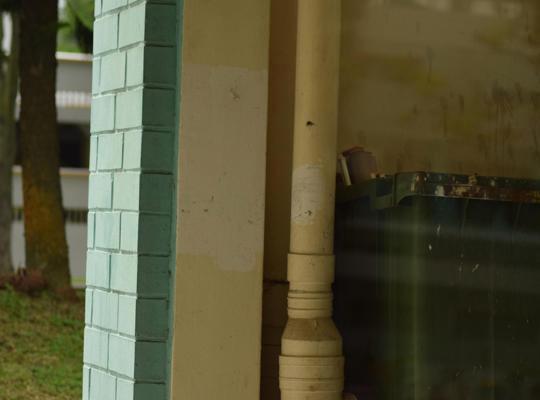}
			&\includegraphics[width=2.5cm]{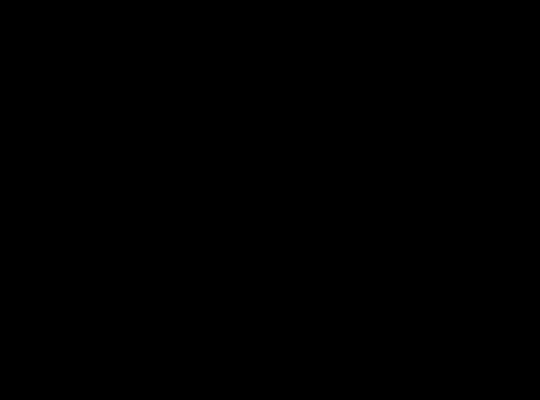}
			&\includegraphics[width=2.5cm]{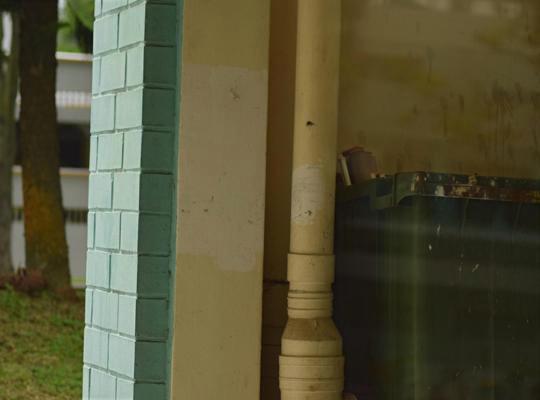}
			&\includegraphics[width=2.5cm]{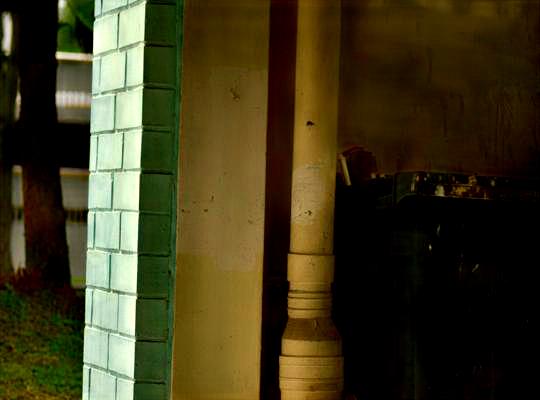}				
			&\includegraphics[width=2.5cm]{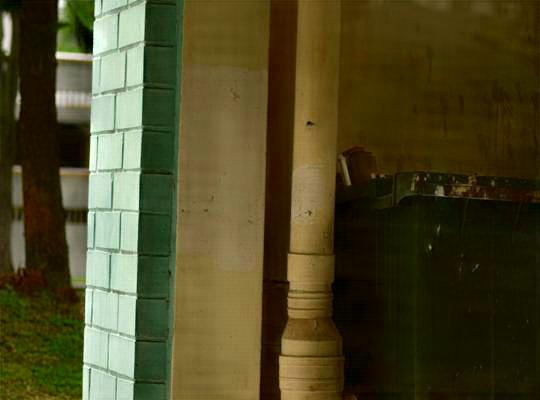}	
			&\includegraphics[width=2.5cm]{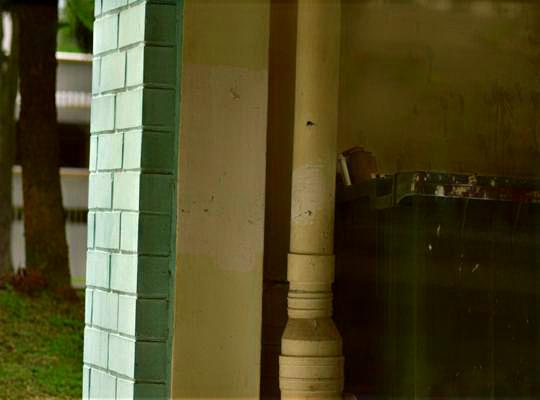}	
			&\includegraphics[width=2.5cm]{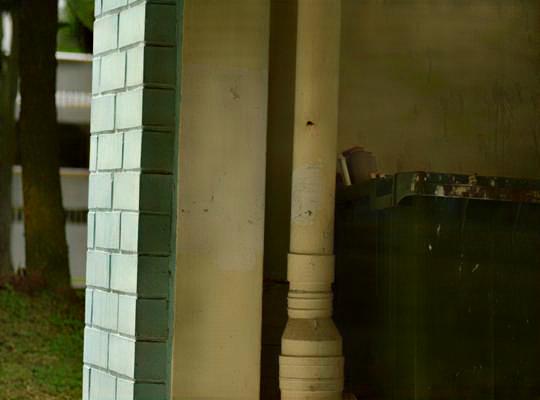}
			\\
			
			\small{Input} & \small{wo. $\mathcal{L}_{f.}$} & \small{wo. $\mathcal{L}_{c.}$}  & \small{wo. $\mathcal{L}_{r.}$} & \small{6-channel} & \small{2-networks} 
            &\small{Ours}
			\\
		\end{tabular}
		
	\end{center}
	\vspace{-3mm}
	\caption{Visual comparison of different alternatives of our algorithm on the strong real-world reflection images from \cite{fan2017generic} (top 2) and wild scene images in the SIR$^2$ dataset (bottom 2). $\mathcal{L}_{f.}$, $\mathcal{L}_{c.}$ and $\mathcal{L}_{r.}$ are short for $\mathcal{L}_{floor}$, $\mathcal{L}_{ceiling}$ and $\mathcal{L}_{recons}$ separately.}
	\label{figure:ablation}
	\vspace{-2mm}
\end{figure*}

\subsection{Ablation Study}
\subsubsection{How important is each loss?} \label{sec:loss_importance}
To analyze the effectiveness of each loss function, we train our network with three alternatives, by removing (1) the floor rejection loss (wo. $\mathcal{L}_{f.}$), (2) the ceiling rejection loss (wo. $\mathcal{L}_{c.}$), and (3) the reconstruction loss (wo. $\mathcal{L}_{r.}$). 

The results are shown in Figure \ref{figure:ablation} and Table \ref{table:ablation}. (1) When training without the floor rejection loss, the network learns a naive solution by simply setting the background to be completely black, and achieves the global minimum of zero energy for the combined objective of reconstruction and ceiling rejection loss. (2) When removing the supervision of the ceiling rejection loss, the learned background tends to approach both input images, and confuses the network with the attempt to maintain reflections without punishment. Hence in the visual results, some obvious reflections remain and the color degradation effect appears. (3) When removing the reconstruction loss, the predicted reflections are left without any supervision, and the two predicted backgrounds have no constraints to be the same. Hence the multi-input constraint is not fully leveraged in this condition. In the according visual results, some reflections are not removed correctly. 
All in all, compared to the three variants of our algorithm, our full pipeline achieves the best numerical and visual results.

\subsubsection{Replacement for latent code?}
Instead of learning latent code within one single network, one alternative solution is to directly learn two networks for background and reflection separately. As shown in Figure \ref{figure:ablation}, this 2-network solution shows very similar visual results, in comparison to our proposed one-network solution. It demonstrates our efficiency by learning a single network to achieve similar performance with two networks.

Another naive replacement is to directly learn a single network with a 6-channel output, where the half are for background and the other half are for reflection. The predicted background from this trained network still contains many obvious reflections, which do not exist in our result. It justifies the effectiveness of our specific network design.

\setlength{\tabcolsep}{2.5pt}
\begin{table*}[ht]
	\begin{center}
        \small
        \vspace{12pt}
		\begin{tabular}{lcccccccc}

			\toprule
			& wo. $\mathcal{L}_{f.}$ & wo. $\mathcal{L}_{c.}$  & wo. $\mathcal{L}_{r.}$  & 6-channel & 2-network & \revise{3-Input image} & \revise{4-Input image}  & Ours \\
			\midrule
			SSIM$_r$ & 0.004 & 0.815 & 0.607 & 0.790 & 0.820 &\revise{\textbf{0.834}} & \revise{0.828} &  {0.833} \\
			\bottomrule
			
		\end{tabular}
	\end{center}
	\caption{Ablation study on the SIR$^2$ dataset. $\mathcal{L}_{f.}$, $\mathcal{L}_{c.}$ and $\mathcal{L}_{r.}$ are short for $\mathcal{L}_{floor}$, $\mathcal{L}_{ceiling}$ and $\mathcal{L}_{recons}$ separately.}
	\label{table:ablation}
\end{table*}

\subsection{Comparison of Computation Cost}

\setlength{\tabcolsep}{5pt}
\begin{table*}[htbp]
  \centering
  \small
    \begin{tabular}{lccccccccc}
    \toprule
    Method &   \cite{wan2018crrn}   & \cite{li2020single}    &   \cite{wen2019single}    &  \cite{yang2018seeing}     & \cite{li2014single}      &  \cite{fan2017generic}     & \cite{zhang2018single}      &  \cite{wei2019single}   &  Ours \\
    \midrule
    Running Time \scriptsize{($\times 10^{-2}$s)} & 1.045 &  3.709  & 1.043 & 0.780 & 250.7 & 1.032 & 8.305 & 1.627 & 0.921 \\
    \bottomrule
    \vspace{1mm}
    \end{tabular}
    \caption{\revise{Running time on images in SIR$^2$ dataset of resolution $540\times400$ on a single Tesla V100 GPU.}}
  \label{tab:running_time}
\end{table*}

\revise{We evaluate the running time for all the methods on images in SIR$^2$ dataset of resolution $540\times400$. For learning-based methods, we compute the average time cost of a single forward pass of the adopted network, and for the optimization-based method \cite{li2014single}, we record the average running time of the optimization process until convergence. All the experiments are conducted on the same device with a Tesla V100 GPU.
As shown in Table \ref{tab:running_time}, benefited from the mechanism of neural networks, learning-based methods significantly run faster than the optimization-based method.
}

Due to the employment of the deep neural network, our algorithm is freed from optimization during evaluation and runs in real time, 
while the traditional optimizer such as the typical gradient descent approach Adam \cite{kingma2014adam} that directly optimizes the target image over thousands of iterations for convergence and takes 84s, which is significantly slower than our approach.

\subsection{Domain Shift Issue}

We further test our algorithm on the synthetic test data generated by both subtraction and clipping image model and linear addition image model, and observe degraded performance compared to our fully supervised alternative. This is due to the fact that our algorithm is essentially powered with the multi-image optimization-based solution, and hence reasons the clean background layer instead of simply approximating the distribution of the synthetic data domain. Therefore, our proposed approach is able to overfit less on the synthetic training data but generalizes better on the real test data.

\revise{
\subsection{Generalization Ability}
We summarize the numerical results of \textit{Ours*} and \textit{Ours} on both the synthetic data generated by our image model and real data from public benchmarks in Table \ref{tab:generalize}. \textit{Ours*} is directly supervised with the ground truth background layer to overfit the synthetic data, hence achieving better performance than \textit{Ours} on the synthetic data. \textit{Ours} learns the methodology to remove the reflections with the multi-image constraints, instead of directly memorizing the synthetic data distribution, hence performing better on the real reflection images. It also demonstrates the fact that \textit{Ours} learns better cross-domain generalizable knowledge than \textit{Ours*}.
}

\begin{table*}[t]
  \centering
  \small
  \setlength{\tabcolsep}{6pt}
    \begin{tabular}{l|cc|ccc|cc|cc}
    \toprule
    \multicolumn{1}{l}{Data source} & \multicolumn{2}{c}{Synthetic} & \multicolumn{3}{c}{SIR$^2$} & \multicolumn{2}{c}{Zhang} & \multicolumn{2}{c}{Nature}  \\
    \midrule
    Error metric & {PSNR} & {SSIM}  & {PSNR} & {SSIM} &{SSIM$_r$} &  {PSNR} & {SSIM} &  {PSNR} & {SSIM}  \\
    \midrule
    Ours*   &  \textbf{22.94}     &  \textbf{0.935}     &  22.45     &  0.778     &   0.765    &   18.54    &  0.753     &  21.85     &   0.752   \\
    Ours    &   21.66    &  0.910     &  \textbf{23.24}     &  \textbf{0.870}     &   \textbf{0.833}    &   \textbf{19.00}    &  \textbf{0.771}     &  \textbf{22.91}     &   \textbf{0.787}   \\
    \bottomrule
    \end{tabular}
    \vspace{2mm}
  \caption{\revise{Comparison with two versions of our algorithms on synthetic data and real data (\textit{SIR$^2$}, \textit{Zhang} and \textit{Nature}). \textit{Ours*} fits the synthetic data well while \textit{Ours} performs better on real images.}}
  \label{tab:generalize}
\end{table*}

\subsection{What Does the Latent Code Learn?}

In our implementation, the latent code encodes the unique information for background or reflection, while the rest of the network weights are shared. It means the layer separation task is conducted by the specific instance normalization layer where our learned latent code is embedded. Then we are interested in the difference between the feature maps generated by this instance normalization layer with the background and reflection latent codes.

Inspired by the sparsity property of convolution features \cite{chen2017stylebank}, we study the active feature channel for background and reflection. To achieve it, we deactivate each channel of the feature map by setting its value to all zero, and computing the difference (MSE) between the manipulated and untouched output image. We conduct such an experiment among 100 real reflection images. As long as the average MSE is larger than 0, this feature channel is considered ``active'' for output generation.

As a result, among all 64 feature maps, interestingly we find that only 35 channels are active for background, and 40 channels for reflection, while eliminating all the other channels do not influence the output generation. Moreover, the background and reflection share only 20 common active channels. However, the active channel is not always effective. When filtering out the ``ineffective'' ones (average MSE $<$ 1), there are only 11 overlapped channels. It means that the learned latent code differentiates background and reflection by allocating different active feature maps.

\begin{figure}[t]
	\begin{center}
		
		\begin{tabular}{ccc}
			
			\includegraphics[width=2.7cm]{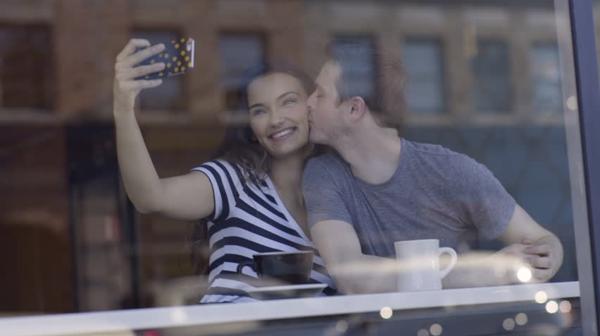}
			& \includegraphics[width=2.7cm]{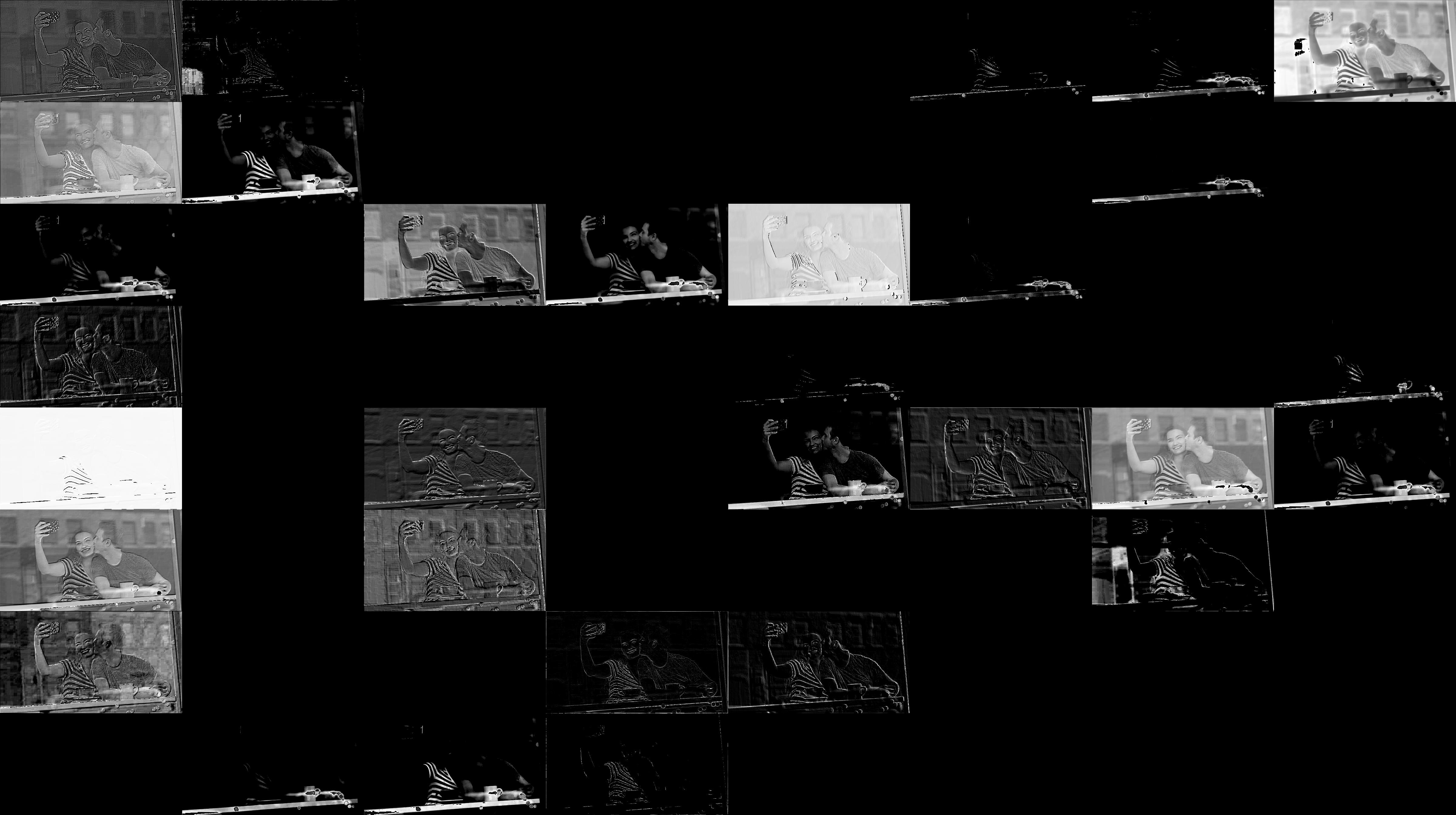}
			& \includegraphics[width=2.7cm]{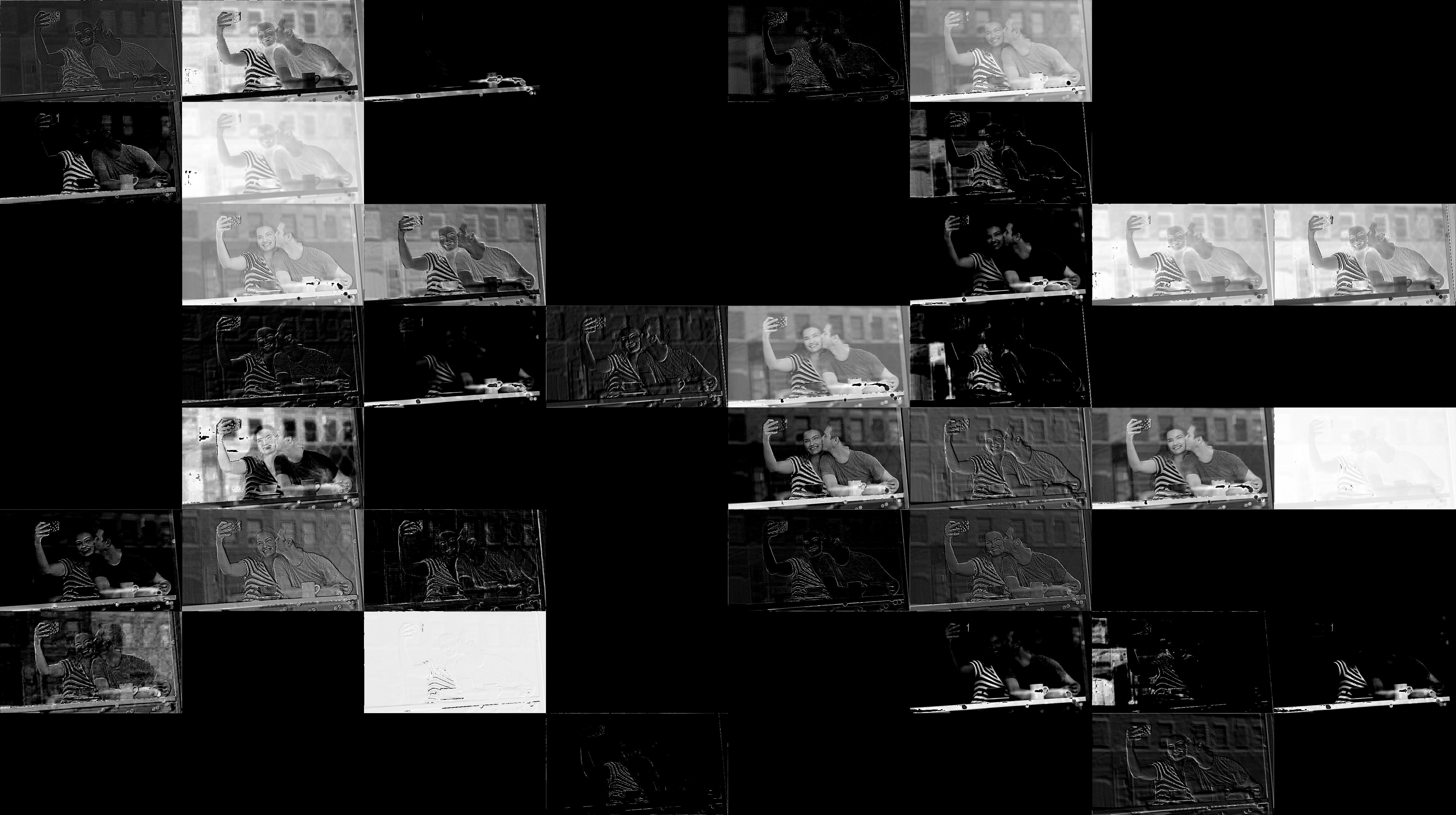}
			\\
			
			\includegraphics[width=2.7cm]{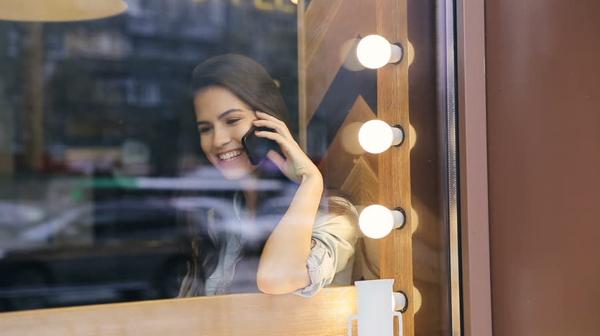}
			& \includegraphics[width=2.7cm]{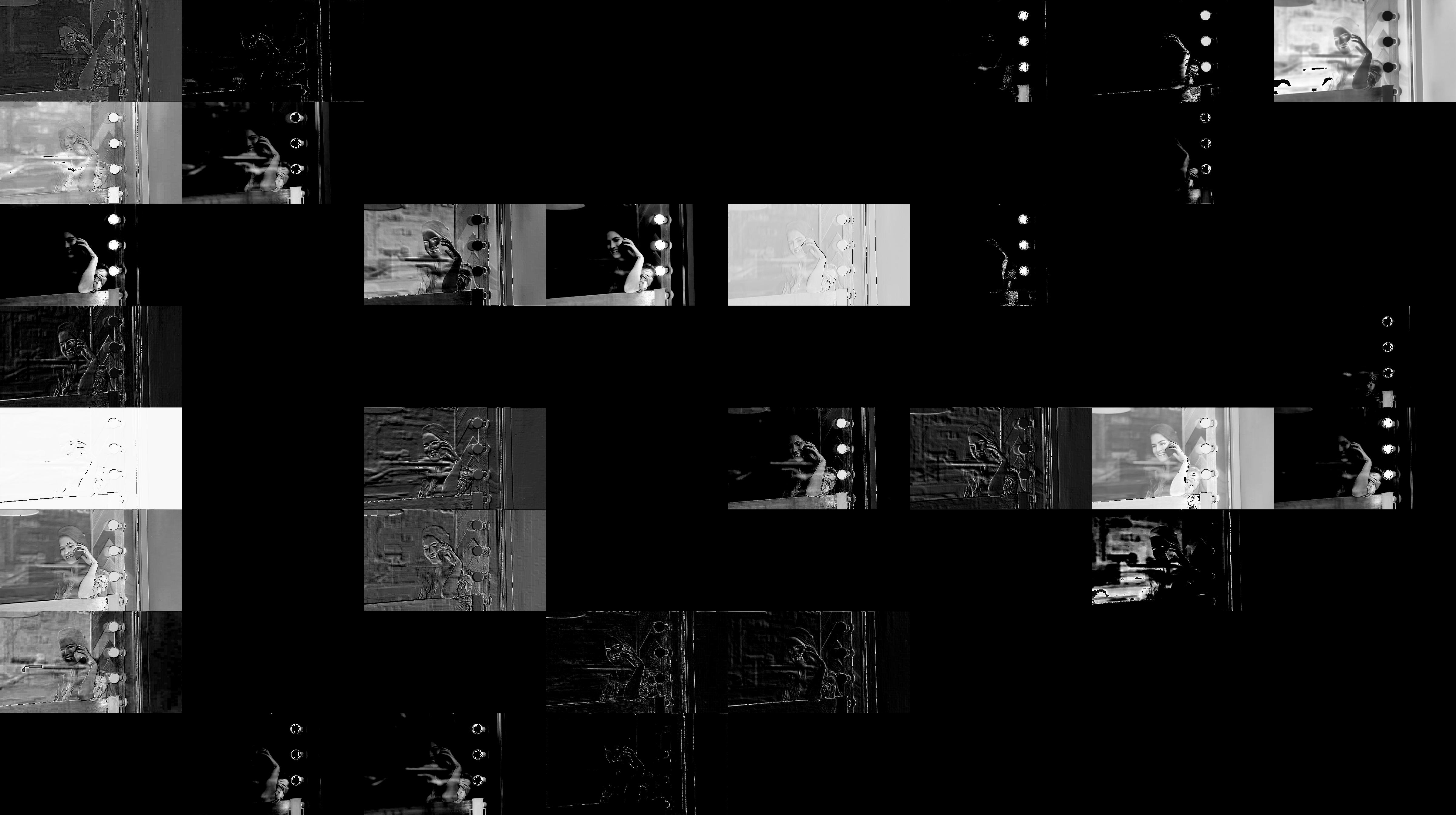}
			& \includegraphics[width=2.7cm]{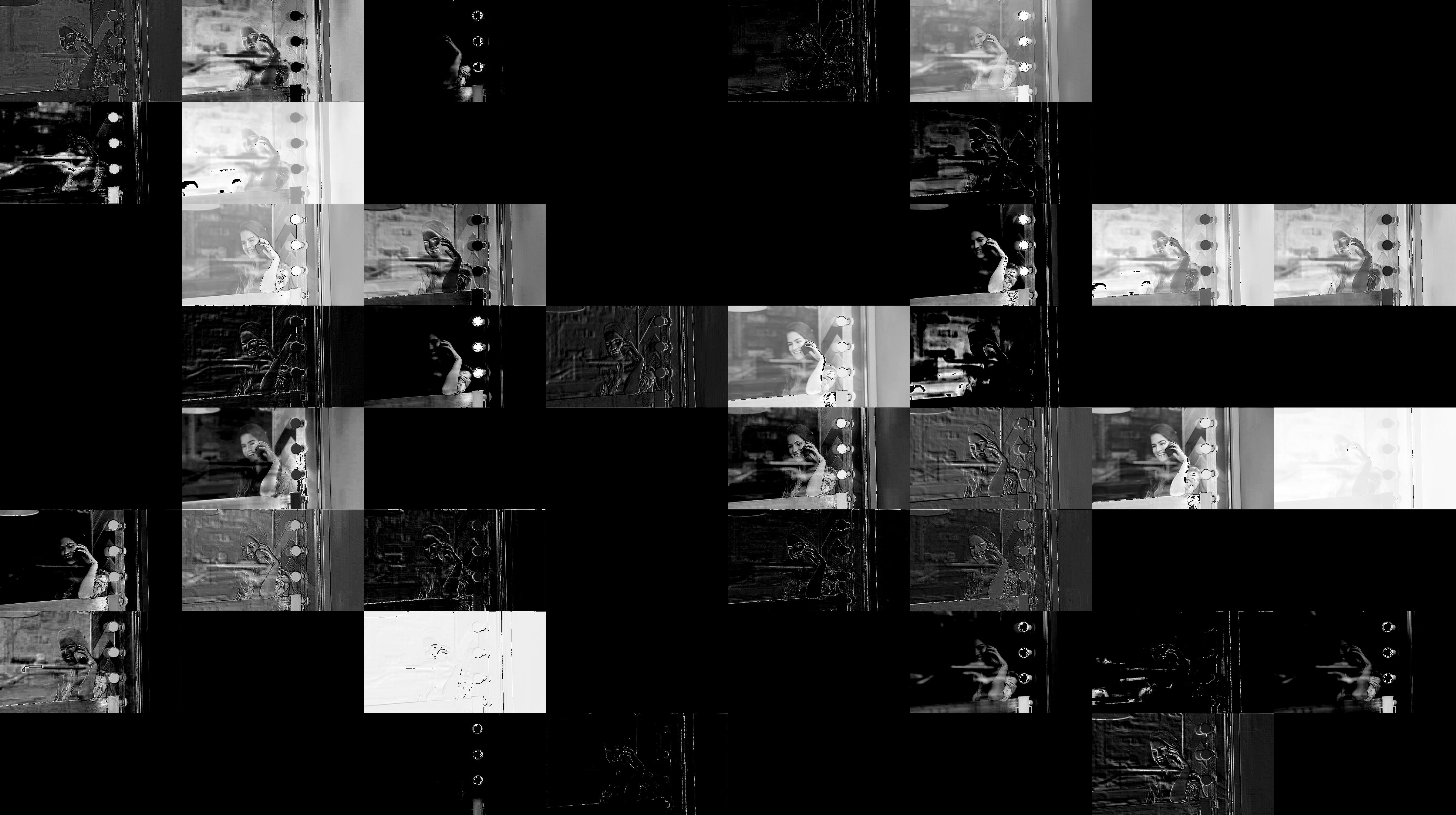}
			\\
			
			\small{Input} & \small{Background feat.} & \small{Reflection feat.}
		\end{tabular}
	
	\end{center}
	\caption{Visualization of background and reflection features after the first instance normalization layer where the latent code is embedded. 64 feature maps are allocated into an 8$\times$8 block for visualization.}
	\label{figure:feature}
\end{figure}

To further justify the above observation, we visualize the normalized feature maps of two randomly picked real images in Figure \ref{figure:feature}. As shown in this figure, many feature maps are totally blank while only part of them contain color intensities. And the active background and reflection feature maps are mostly different from each other. A similar phenomenon is consistently observed in many more examples. 

\setlength{\tabcolsep}{1pt}
\renewcommand{\arraystretch}{1}

\revise{
\subsection{Evaluations on the Separated Reflection Layers.}

The quantitative and qualitative comparisons of the separated reflection layers are shown in Table \ref{tab:refl_layer}, Figure \ref{figure:fanICCV} and \ref{figure:wanICCV}. %
As shown in the visual comparisons, our method learns to remove more reflections and generate cleaner backgrounds even with strong reflection artifacts. For the numerical evaluations of separated reflection layers, our method yields good performance, especially on the SSIM$_r$ metric. 

\begin{table}[htbp]
  \centering
  \footnotesize
  \setlength{\tabcolsep}{10pt}
    \begin{tabular}{lrrr}
    \toprule
     Data source & \multicolumn{1}{l}{PSNR} & \multicolumn{1}{l}{SSIM} & \multicolumn{1}{l}{SSIM$_r$} \\
    \midrule
    Input                   & 28.33 & 0.546 & 0.410 \\
    \midrule
    \cite{wan2018crrn}      & 23.30 & 0.377 & 0.428 \\
    \cite{li2020single}     & 24.33 & \textcolor{red}{0.534} & 0.454 \\
    \cite{wen2019single}    & 23.68 & 0.484 & 0.407 \\
    \cite{yang2018seeing}   & 24.65 & \textcolor{blue}{0.523} & \textcolor{red}{0.495} \\
    \cite{li2014single}     & 19.65 & 0.291 & 0.337 \\
    \cite{fan2017generic}   & 23.33 & 0.421 & 0.358 \\
    \cite{zhang2018single}  & 22.12 & 0.379 & 0.404 \\
    \cite{wei2019single}    & \textcolor{red}{25.70} & 0.515 & 0.463 \\
    \midrule
    Ours*                   & 23.83 & 0.459 & 0.427 \\
    Ours                    & \textcolor{blue}{24.97} & 0.503 & \textcolor{blue}{0.479} \\
    \bottomrule
    \end{tabular}%
    \vspace{3mm}
    \caption{\revise{Comparison of the separated reflection layers on wild scene images of SIR$^2$ dataset.}}
  \label{tab:refl_layer}%
\end{table}%

}

\revise{
\subsection{More Inputs to the Network?}
For sake of simplicity, we take two input reflection images to train our framework in the main paper. To further explore its performance, we experiment with our algorithm with three and four input images. As shown in Table \ref{table:ablation}, feeding more input images leads to similar results to a simple two-image case, which demonstrates that our pipeline learns sufficiently good reflection removal effects with only two input images. 
}

\subsection{Robustness of Hyper-parameters?}

In order to test the robustness of hyper-parameters in the objective function, \revise{we test our algorithm on SIR$^2$ dataset with variations of $\lambda_1$ and $\lambda_2$ separately and report the results in Table \ref{tab:hyper}. The numerical comparisons show the robustness of our algorithm to the change of hyper parameters. 
}

\begin{table}[htbp]
\footnotesize
\centering
\centering
\setlength{\tabcolsep}{10pt}
    \begin{tabular}{lccc}
        \toprule
        $\lambda_1$ & PSNR & SSIM & SSIM$_r$ \\
        \midrule
        70    & 23.23 & 0.858 & 0.812 \\
        80    & {23.24} & \textbf{0.870}  & \textbf{0.833} \\
        90    & \textbf{23.63} & 0.866 & 0.824 \\
        \bottomrule
    \end{tabular}%
\vspace{3mm}
\setlength{\tabcolsep}{10pt}
\centering
    \begin{tabular}{lccc}
    \toprule
    $\lambda_2$ & PSNR & SSIM & SSIM$_r$ \\
    \midrule
    40    & 23.23 & 0.849 & 0.816 \\
    50    & \textbf{23.24} & \textbf{0.870}  & \textbf{0.833} \\
    60    & 23.12 & 0.856 & 0.825 \\
    \bottomrule
    \end{tabular}%
\vspace{3mm}
\caption{\revise{Comparison with variant hyper parameters $\lambda_1$ and $\lambda_2$ on wild scene images of SIR$^2$ dataset.}}
\label{tab:hyper}%
\end{table}%

\section{Conclusion}

In this work, we propose to solve the single image reflection removal problem by taking advantage of the multi-image prior in the training phase in an optimization manner, and transferring the learned knowledge into a novel reflection removal network backbone that learns to remove reflections from a single input. Our proposed algorithm generalizes better in real image domain, achieves state-of-the-art performance, and runs in real-time.

\bibliographystyle{IEEEtran}
\bibliography{bare_jrnl}

\begin{IEEEbiography}[{\includegraphics[width=1in,height=1.25in,clip,keepaspectratio]{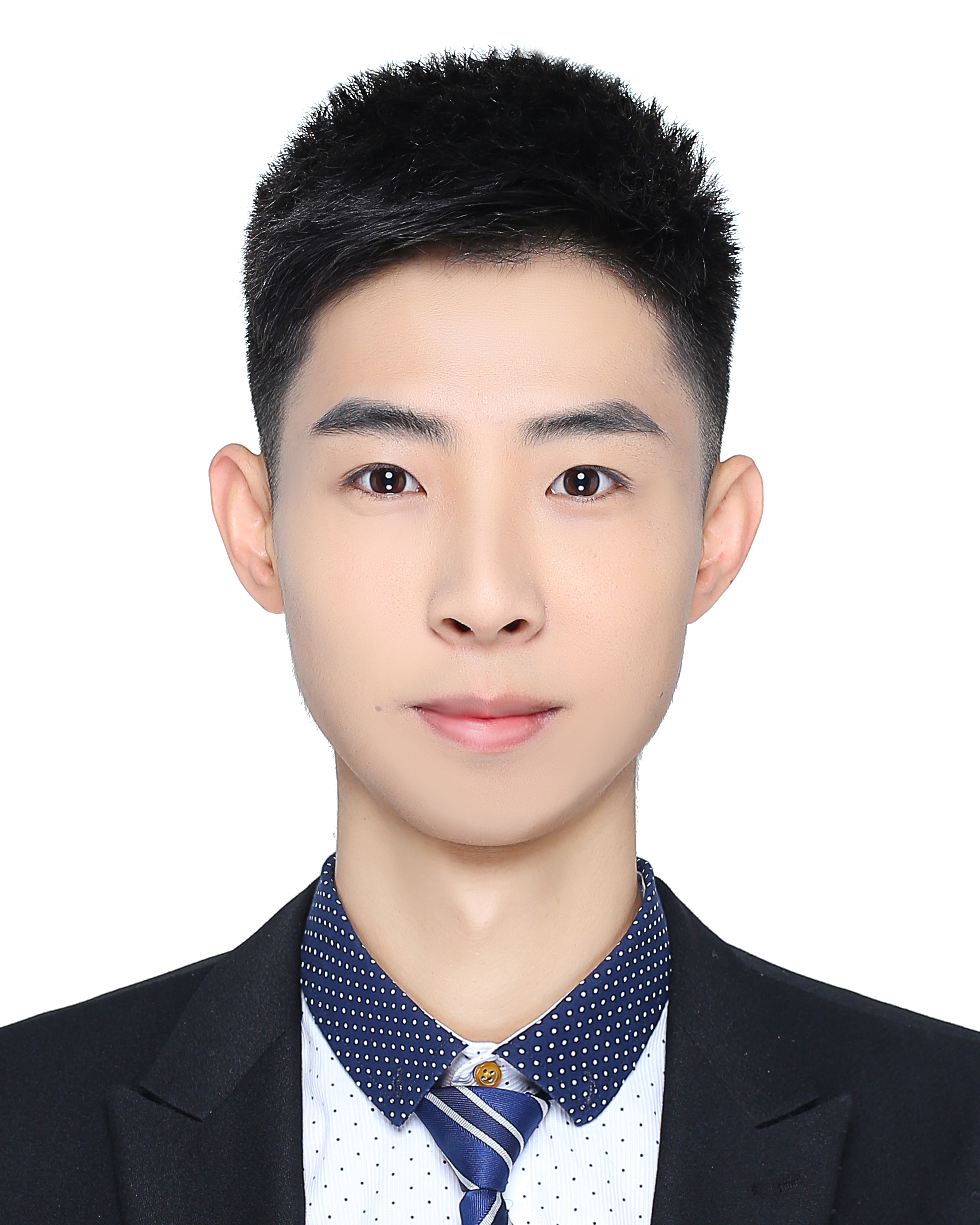}}]{Yingda Yin}
is a Ph.D. student at Center on Frontiers of Computing Studies (CFCS), Peking University. His research interests mainly include image processing and 3D vision.
\end{IEEEbiography}

\begin{IEEEbiography}[{\includegraphics[width=1in,height=1.25in,clip,keepaspectratio]{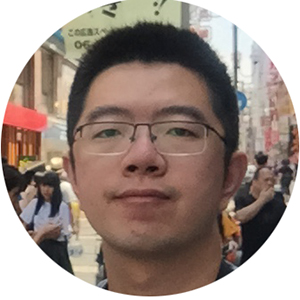}}]{Qingnan Fan}
is a Senior Researcher in the Visual Computing Center of Tencent AI Lab. He received his PhD degree from Shandong University in 2019. His research interests mainly include image/video processing and 3D vision.
\end{IEEEbiography}

\begin{IEEEbiography}[{\includegraphics[width=1in,height=1.25in,clip,keepaspectratio]{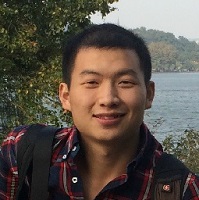}}]{Dongdong Chen}
is a senior researcher at Microsoft Cloud + AI. Before that, he obtained his PhD degree from University of Science and Technology of China (USTC) under the joint PhD program between MSRA and USTC in 2019. His research interests mainly include image generation, image restoration, low-level image processing, and high-level image recognition tasks.
\end{IEEEbiography}

\begin{IEEEbiography}[{\includegraphics[width=1in,height=1.25in,clip,keepaspectratio]{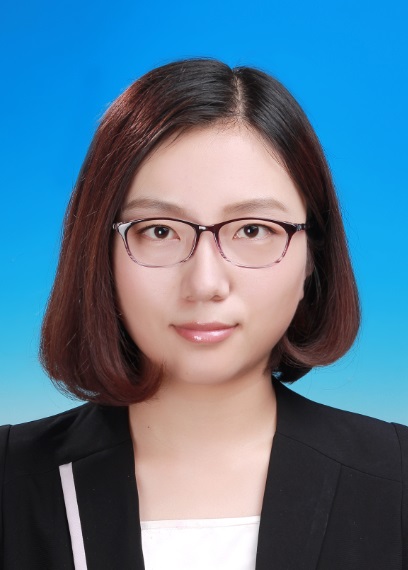}}]{Yujie Wang}
is a Ph.D. student at School of Computer Science and Technology, Shandong University. Before that, she received her MS degree from Tianjin University. Her research interests mainly include image processing and computational holography.
\end{IEEEbiography}

\begin{IEEEbiography}[{\includegraphics[width=1in,height=1.25in,clip,keepaspectratio]{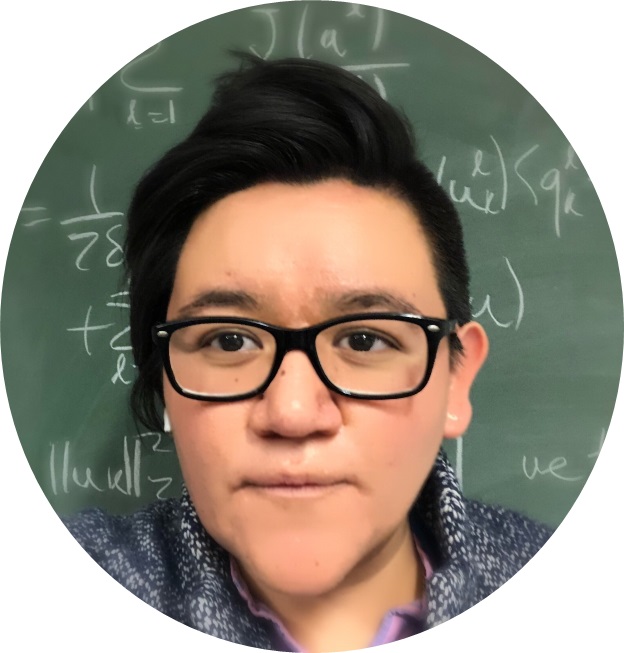}}]{Angelica I. Aviles-Rivero}
is currently a post-doctoral researcher (Research Associate) at DPMMS, University of Cambridge. She received her Ph.D. degree (in 2017) in Computer Vision and Robotics at the UPC-BarcelonaTech, Spain under the supervision of Prof. A. Casals. Her research lies at the intersection of computational mathematics and machine learning for applications to large-scale real world problems.
\end{IEEEbiography}

\begin{IEEEbiography}[{\includegraphics[width=1in,height=1.25in,clip,keepaspectratio]{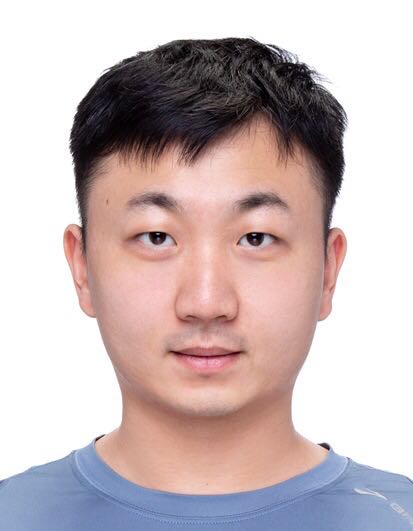}}]{Ruoteng Li}
is a Ph.D. student at Electrical \& Computer Engineering department, National University of Singapore. His research interests involves optical flow estimation, bad weather, image restoration and object detection \& recognition.
\end{IEEEbiography}

\begin{IEEEbiography}[{\includegraphics[width=1in,height=1.25in,clip,keepaspectratio]{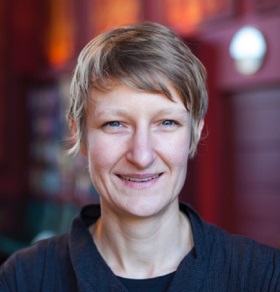}}]{Carola-Bibiane Schönlieb}
is a Professor of Applied Mathematics and head of the Cambridge Image Analysis (CIA) group at the Department of Applied Mathematics and Theoretical Physics (DAMTP), University of Cambridge. Moreover, she is the Director of the Cantab Capital Institute for the Mathematics of Information, Director of the EPSRC Centre for Mathematical and Statistical Analysis of Multimodal Clinical Imaging, a Fellow of Jesus College, Cambridge, and co-leader of the IMAGES network. Her research interests lie in the mathematics of digital image and video processing.
\end{IEEEbiography}

\begin{IEEEbiography}[{\includegraphics[width=1in,height=1.25in,clip,keepaspectratio]{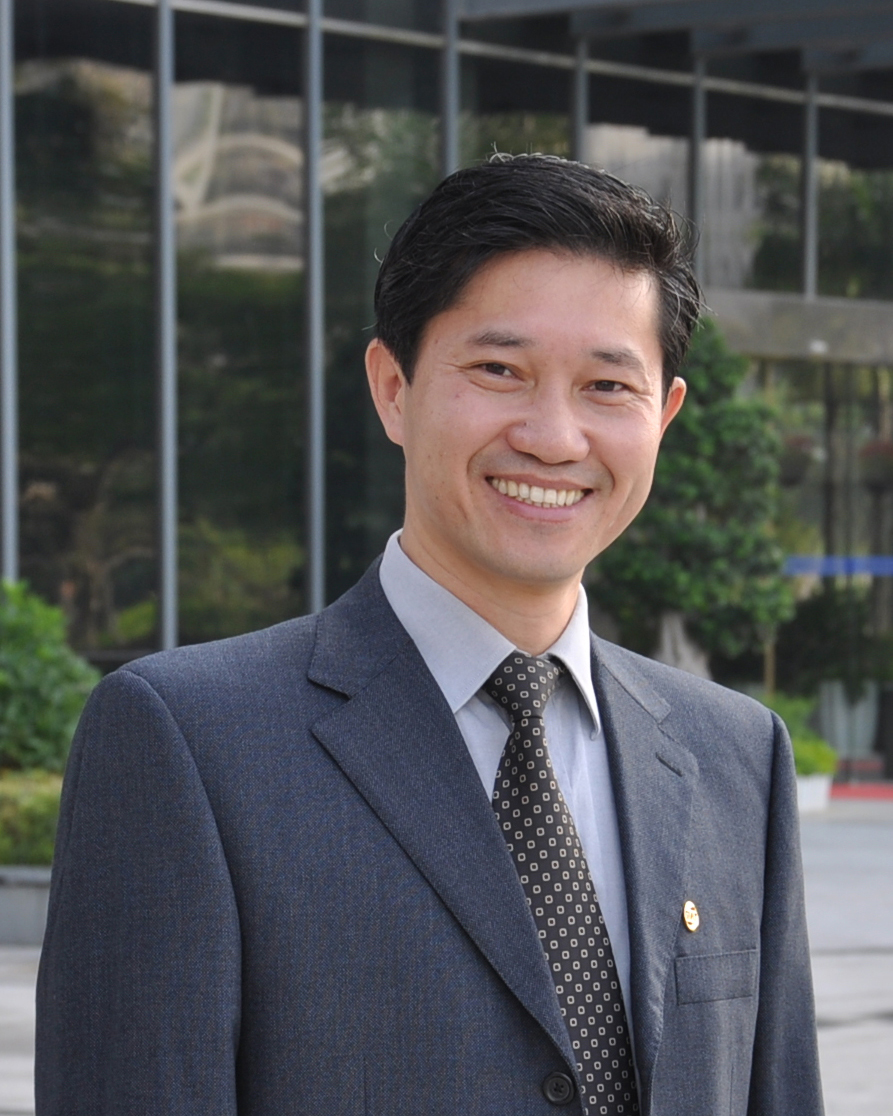}}]{Baoquan Chen}
is an Endowed Boya Professor of Peking University, where he is the Executive Director of the Center on Frontiers of Computing Studies. Prior to the current post, he was Dean, School of Computer Science and Technology, SDU. Chen received an MS in Electronic Engineering from Tsinghua University, Beijing (1994), and a second MS (1997) and then PhD (1999) in Computer Science from the State University of New York at Stony Brook. For his contribution to spatial data (modeling) and visualization, he was elected IEEE Fellow 2020. His research interests generally lie in computer graphics, visualization, and human-computer interaction, focusing specifically on large-scale city modeling, simulation, and visualization. 
\end{IEEEbiography}

\vfill
\vfill

\end{document}